\documentclass{article} 
\usepackage{arxiv, times, natbib}

\usepackage{amsmath,amsfonts,bm}

\def\eqref#1{equation~\ref{#1}}

\def\1{\bm{1}}

\DeclareMathAlphabet{\mathsfit}{\encodingdefault}{\sfdefault}{m}{sl}
\SetMathAlphabet{\mathsfit}{bold}{\encodingdefault}{\sfdefault}{bx}{n}

\newcommand{\R}{\mathbb{R}}

\usepackage{mathtools}
\usepackage{pifont}
\usepackage{xcolor}
\usepackage{url}

\usepackage{multirow}
\usepackage{graphicx, subfig}
\usepackage{booktabs}
\usepackage{amsmath, amssymb, amsthm}
\usepackage{hyperref, cleveref}
\usepackage{xkcdcolors}
\usepackage{wrapfig}
\usepackage[labelfont=bf, justification=justified]{caption}
\usepackage{dsfont}
\usepackage{floatrow}

\usepackage{booktabs}
\usepackage[export]{adjustbox}  

\newtheorem{theorem}{Theorem}

\usepackage{xfakebold,xspace}
\usepackage{ifthen}

\usepackage{tikz}
\usetikzlibrary{3d}

\newlength{\subcolumnwidth}
\newenvironment{subcolumns}[1][0.45\columnwidth]
 {\valign\bgroup\hsize=#1\setlength{\subcolumnwidth}{\hsize}\vfil##\vfil\cr}
 {\crcr\egroup}
\newcommand{\nextsubcolumn}[1][]{%
  \cr\noalign{\hfill}
  \if\relax\detokenize{#1}\relax\else\hsize=#1\setlength{\subcolumnwidth}{\hsize}\fi
}

\NewDocumentCommand{\fbseries}{ o }{%
  \IfValueTF{#1}
    {\unskip\setBold[#1]}%
    {\setBold[0.3]}
  \aftergroup\unsetBold\aftergroup
  \xspace
}

\newcommand{\cmark}{\ding{51}}%
\newcommand{\xmark}{\ding{55}}%
\newcommand{\good}{{\color{xkcdGreen}\cmark}}
\newcommand{\bad}{{\color{xkcdRed}\xmark}}
\newcommand{\fair}{{\color{xkcdGoldenrod}{\fbseries[1]$\sim$}}}

\newcommand{\modelname}{GA-Planes}
\newcommand{\decoder}[1][]{%
    \text{D}%
    \ifx\relax#1\relax\else(#1)\fi%
}

\newcommand{\eone}{\mathbf{e}_1}
\newcommand{\etwo}{\mathbf{e}_2}
\newcommand{\ethree}{\mathbf{e}_3}
\newcommand{\eonetwo}{\mathbf{e}_{12}}
\newcommand{\eonethree}{\mathbf{e}_{13}}
\newcommand{\etwothree}{\mathbf{e}_{23}}
\newcommand{\eonetwothree}{\mathbf{e}_{123}}
\newcommand{\e}[1]{\mathbf{e}_{#1}}
\newcommand{\ebar}[1]{\mathbf{\overline{e}}_{#1}}
\newcommand{\gone}{\mathbf{g}_1}
\newcommand{\gtwo}{\mathbf{g}_2}
\newcommand{\gthree}{\mathbf{g}_3}
\newcommand{\gonetwo}{\mathbf{g}_{12}}
\newcommand{\gonethree}{\mathbf{g}_{13}}
\newcommand{\gtwothree}{\mathbf{g}_{23}}
\newcommand{\gonetwothree}{\mathbf{g}_{123}}
\newcommand{\g}[1]{\mathbf{g}_{#1}}

\title{Geometric Algebra Planes: \\Convex Implicit Neural Volumes}

\author{Irmak Sivgin*, Sara Fridovich-Keil*\thanks{Incoming faculty at the School of Electrical \& Computer Engineering, Georgia Institute of Technology (\texttt{sfk@gatech.edu}).}, Gordon Wetzstein, \& Mert Pilanci  \\
Department of Electrical Engineering, Stanford University\\
\texttt{\{isivgin, sarafk, gordonwz, pilanci\}@stanford.edu} \\
* denotes equal contribution\\
}

\begin{document}

\maketitle

\begin{abstract}
Volume parameterizations abound in recent literature, from the classic voxel grid to the implicit neural representation and everything in between.
While implicit representations have shown impressive capacity and better memory efficiency compared to voxel grids, to date they require training via nonconvex optimization. 
This nonconvex training process can be slow to converge and sensitive to initialization and hyperparameter choices that affect the final converged result.
We introduce a family of models, \modelname{}, that is the first class of implicit neural volume representations that can be trained by \emph{convex} optimization. 
\modelname{} models include any combination of features stored in tensor basis elements, followed by a neural feature decoder. 
They generalize many existing representations and can be adapted for convex, semiconvex, or nonconvex training as needed for different inverse problems.
In the 2D setting, we prove that \modelname{} is equivalent to a low-rank plus low-resolution matrix factorization; we show that this approximation outperforms the classic low-rank plus sparse decomposition for fitting a natural image.
In 3D, we demonstrate \modelname{}' competitive performance in terms of expressiveness, model size, and optimizability across three volume fitting tasks: radiance field reconstruction, 3D segmentation, and video segmentation. Code is available at \url{https://github.com/sivginirmak/Geometric-Algebra-Planes}. 
\end{abstract}

\vspace{-0.2cm}
\section{Introduction}

Volumes are everywhere---from the world we live in to the videos we watch to the organs and tissues inside our bodies. 
In recent years tremendous progress has been made in modeling these volumes using measurements and computation \citep{tewarisurvey}, to make them accessible for downstream tasks in applications including manufacturing \citep{manufacturingreview, navigationandmanufacturing}, robotic navigation \citep{nerfforrobots, navigationreview1}, entertainment and culture \citep{contentgeneration, culturalheritage}, and medicine \citep{3dmedicalimagingbook, volumesforhealthcare, 3dforendoscopy, nerfforsurgery}. 
All methods that seek to model a volume face a three-way tradeoff between \emph{model size}, which determines hardware memory requirements, \emph{expressiveness}, which determines how faithfully the model can represent the underlying volume, and \emph{optimizability}, which captures how quickly and reliably the model can learn the volume from measurements. 
Certain applications place stricter requirements on model size (e.g. for deployment on mobile or edge devices), expressiveness (e.g. resolution required for medical diagnosis or safe robotic navigation), or optimizability (e.g. for interactive applications), but all stand to benefit from improvements to this three-way pareto frontier.

Many existing strategies have been successfully applied at different points along this pareto frontier; some representative examples from computer vision are summarized in \Cref{sec:appendixcontext}. 
Our goal is to maintain or surpass the existing pareto frontier of model size and expressiveness while improving optimization stability through convex optimization.

Our approach introduces \emph{convex} and \emph{semiconvex} reformulations of the volume modeling optimization process that apply to a broad class of volume models we call \emph{Geometric Algebra Planes}, or \modelname{} for short.
We adopt the term \emph{semiconvex} for Burer-Monteiro (BM) factorizations of a convex objective, as introduced in \cite{mertsemiconvex}, within the context of convex neural networks. BM factorized problems have the property that every local minimum is globally optimal \citep{mertsemiconvex}.

\modelname{} is a mixture-of-primitives model that generalizes several existing volume models including voxels and tensor factorizations. Most importantly, most models in this family can be formulated for optimization by a convex program, as long as the objective function (to fit measurements of the volume) is convex. At the same time, \emph{any} \modelname{} model can also be formulated for nonconvex optimization towards \emph{any} objective, matching the range of applicability enjoyed by common models. While only our convex and semiconvex models come with guarantees of convergence to global optimality, all the models we introduce extend the pareto frontier of model size, expressiveness, and optimizability on diverse tasks.

Concretely, we make the following contributions:
\begin{itemize}
    \item We introduce \modelname{}, a mixture-of-primitives volume parameterization inspired by geometric algebra basis elements. \modelname{} combines any subset of line, plane, and volume features at different resolutions, with an MLP decoder. This GA-Planes family of parameterizations generalizes many existing volume and radiance field models. 
    \item We derive convex and semiconvex reformulations of the \modelname{} training process for certain tasks and a large subset of the \modelname{} model family, to ensure our model optimizes globally regardless of initialization.
    \item We analyze \modelname{} in the 2D setting and show equivalence to a low-rank plus low-resolution matrix approximation whose expressiveness can be directly controlled by design choices. We demonstrate that this matrix decomposition is expressive for natural images, outperforming the classic low-rank plus sparse approximation.
    \item We demonstrate convex, semiconvex, and nonconvex \modelname{}' high performance in terms of memory, expressiveness, and optimizability across three volume-fitting tasks: 3D radiance field reconstruction, 3D segmentation, and video segmentation.
\end{itemize}

\vspace{-0.2cm}
\section{Related Work}

\paragraph{Volume parameterization.}
Many volume parameterizations have been proposed and enjoy widespread use across diverse applications. Here we give an overview of representative methods used in computer vision, focusing on methods that parameterize an entire volume (rather than e.g. a surface). These parameterizations achieve different tradeoffs between memory usage, representation quality, and ease of optimization; richer descriptions are provided in \Cref{sec:appendixcontext}. 

Coordinate MLPs like NeRF \citep{nerf} and Scene Representation Networks \citep{srn} are representative of Implicit Neural Representations (INRs), which excel at reducing model size (with decent expressiveness) but suffer from slow optimization. 
At the opposite end of the spectrum, explicit voxel grid representations like Plenoxels \citep{yu_and_fridovichkeil2021plenoxels} and Direct Voxel Grid Optimization \citep{dvgo} can optimize quickly but require large model size to achieve good expressiveness (resolution). 
Many other methods \citep{tensorf, kplanes, ingp, 3Dgaussians, merf, mvp} find their niche somewhere in between, achieving tractable model size, good expressiveness, and reasonably fast optimization time in exchange for some increased sensitivity (to initialization, randomness, and prior knowledge) in the optimization process.
\modelname{} matches or exceeds the performance of strong baselines \citep{tensorf, kplanes, mipnerf} in terms of model size and expressiveness, while introducing the option to train by convex or semiconvex optimization with guaranteed convergence to global optimality.

\paragraph{Radiance field modeling.}
Most of the works described above are designed for the task of modeling a radiance field, in which the training measurements consist of color photographs from known camera poses. The goal is then to faithfully model the optical density and view-dependent color of light inside a volume so that unseen views can by rendered accurately. This task is also referred to as \emph{novel view synthesis} \citep{nerf, srn}. Although we do demonstrate superior performance of \modelname{} in this setting, we note that the volumetric rendering formula used in radiance field modeling \citep{max, kajiya, nerf} yields a nonconvex photometric loss function, regardless of model parameterization. 

\paragraph{3D segmentation.}
We test our convex and semiconvex \modelname{} parameterizations on fully convex objectives, namely volume (xyz) segmentation with either indirect 2D tomographic supervision or direct supervision, as well as video (xyt) segmentation with direct 3D supervision.
This 3D (xyz) segmentation task has also been studied in recent work \citep{cen2023segment, scade}, though these methods require additional inputs such as a pretrained radiance field model or monocular depth estimator. Our setup is most similar to \cite{occupancynet}, which uses an implicit neural representation trained with cross-entropy loss and direct 3D supervision of the occupancy function. Instead of having direct access to this 3D training data, we infer 3D supervision labels via Space Carving \citep{spacecarving} from 2D image masks obtained by image segmentation (via \cite{sam1}).

\paragraph{Convex neural networks.}
Recent work has exposed an equivalence between training a shallow \citep{convexnn2layer} or deep \citep{deepconvex} neural network and solving a convex program whose structure is defined by the architecture and parameter dimensions of the corresponding neural network. 
The key idea behind this convexification procedure is to enumerate (or randomly sample from) the possible activation paths through the neural network, and then treat these paths as a fixed dictionary whose coefficients may be optimized according to a convex program.
Given a data matrix $X \in \mathbb{R}^{n\times d} $ and labels $y \in \mathbb{R}^{n}$, a 2-layer nonconvex ReLU MLP approximates $y$ as
\begin{equation}
\label{eq:nonconvNN}
    y \approx \sum_{j=1}^m {(X U_j)_{+} \alpha_j},
\end{equation}
where $m$ is the number of hidden neurons and $U$ and $\alpha$ are the first and second linear layer weights, respectively. \cite{convexnn2layer} proposed to instead approximate $y$ as 
\begin{equation}
\label{eq:convNN}
    y \approx \sum_{i=1}^P {D_i X (v_i - w_i)},
\end{equation}
subject to $(2D_i-I_n)Xv_i \geq 0$ and $(2D_i-I_n)Xw_i \geq 0$ for all $i$.
The parameters $v$ and $w$ in \cref{eq:convNN} replace the first and second layer weights $U$ and $\alpha$ from the nonconvex formulation in \cref{eq:nonconvNN} (optimal values of $U$ and $\alpha$ can be recovered from optimal values of $v$ and $w$).
Here $D_i$ represent different possible activation patterns of the hidden neurons as $\{D_i\}_{i=1}^P := \{\text{Diag}(\mathds{1}[X u \geq 0]):u\in \mathbb{R}^d\}$, which is the finite set of hyperplane arrangement patterns obtained for all possible $u \in \mathbb{R}^{d}$. 
We can sample different $u$'s to find all distinct activation patterns $\{D_i\}_{i=1}^P$, where $P$ is the number of regions in the partitioned input space.
Enumerating all such patterns would yield an exact equivalence with the global minimizer of the nonconvex ReLU MLP in \cref{eq:nonconvNN}, but may be complicated or intractable due to memory limitations. Subsampling $\Tilde{P}$ patterns results in a convex program with tractable size, whose solution is one of the stationary points of the original non-convex problem \citep{convexnn2layer}.
We apply this idea to create convex and semiconvex \modelname{} models by convexifying the feature decoder MLP according to this procedure.

\paragraph{Geometric (Clifford) algebra.}
Geometric algebra (GA) is a powerful framework for modeling geometric primitives and interactions between them \citep{geometricalgebrabook}. The fundamental entity in GA is the multivector, which is a sum of vectors, bivectors, trivectors, etc. In 3D GA, an example is the multivector \(\eone\etwo + \eone\etwo\ethree\), representing the sum of a bivector (a plane) and a trivector (a volume). The geometric product in GA allows us to derive a volume element by multiplying a plane and a line, e.g. \((\eone\etwo)\ethree = \eone\etwo\ethree\). We use the shorthand $\eonetwothree=\eone\etwo\ethree$, and similarly for other multivector components. Inspired by this framework, we define the \modelname{} model family to include any volume parameterization that combines any subset (including the complete subset and the empty subset) of the linear geometric primitives $\{ \eone, \etwo, \ethree \}$, planar geometric primitives $\{ \eonetwo, \eonethree, \etwothree \}$, and/or volumetric primitive $\{ \eonetwothree \}$ with a (potentially convexified) MLP feature decoder. We leverage geometric algebra to combine these primitives into a desired trivector (volume). To our knowledge, this work is the first to use geometric algebra in neural volume models.

\vspace{-0.2cm}
\section{Model}
\begin{figure}
\raggedright
    \includegraphics[height=0.1\linewidth]{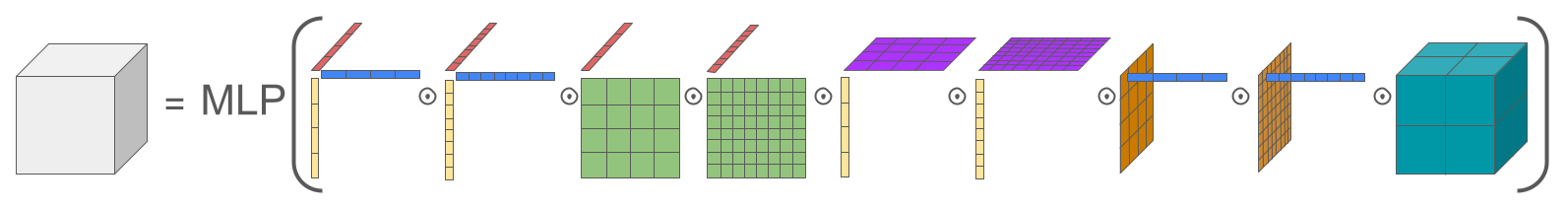}
    \includegraphics[height=0.1\linewidth]{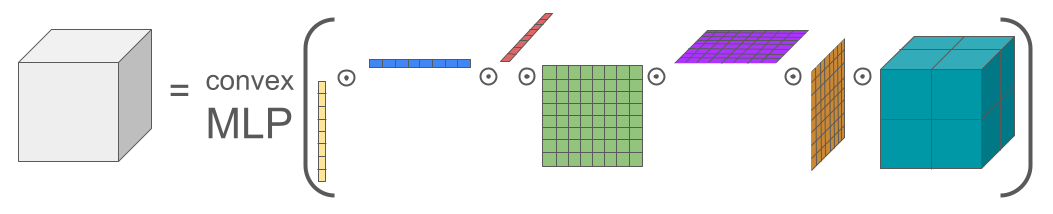}
    \includegraphics[height=0.1\linewidth]{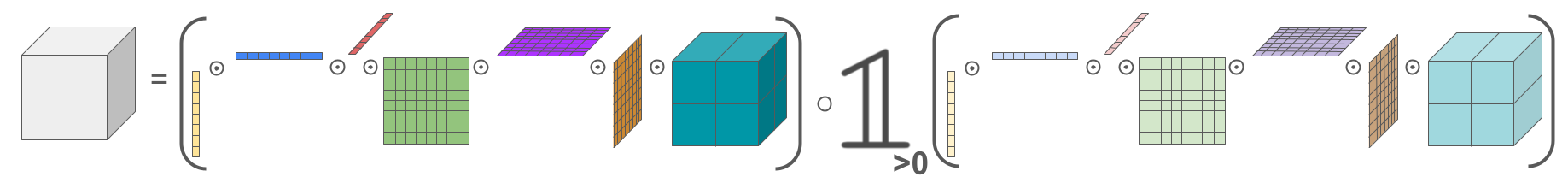}
    \caption{Overview of the \modelname{} models we use in our experiments. Our nonconvex model (top) uses a standard MLP decoder and multiplication of features when the result yields a volume under geometric algebra; it also concatenates features across mult-resolution grids. Our semiconvex (middle) and convex (bottom) models use a single resolution for each feature grid, and avoid multiplication of features since that would induce nonconvexity. The pastel-colored grids inside the indicator function of the convex model are frozen at initialization and used as fixed ReLU gating patterns. $\odot$ denotes concatenation and $\circ$ denotes elementwise multiplication.}
    \label{fig:method}
\end{figure}

\subsection{The \modelname{} Model Family}
\label{sec:gaplane-family}
A \modelname{} model represents a volume using a combination of geometric algebra features $\e{c}$ derived by interpolating the following parameter grids:
\begin{itemize}
    \item Line (1-dimensional) feature grids $\{ \gone, \gtwo, \gthree \}$, where each grid has shape $[r_{1}, d_{1}]$ with spatial resolution $r_{1}$ and feature dimension $d_{1}$.
    \item Plane (2-dimensional) feature grids $\{ \gonetwo, \gonethree, \gtwothree \}$, where each grid has shape $[r_{2}, r_2, d_{2}]$ with spatial resolution $r_{2}$ and feature dimension $d_{2}$. 
    \item A single volume feature grid $\{\gonetwothree\}$ with shape $[r_3, r_3, r_3, d_3]$. 
\end{itemize}
A \modelname{} model may include multiple copies of a given basis element with different resolution and feature dimensions, to effectively capture multi-resolution signal content. The x, y, and z spatial resolutions of each grid may differ in practice; for simplicity of notation we use isotropic resolutions. 

Let $q = (x, y, z)$ be a coordinate of interest in $\R^3$. We first extract features corresponding to $q$ from each of our line, plane, and volume feature grids $\g{c}$ by linear, bilinear, and trilinear interpolation, respectively:
\begin{equation}\label{eq:feature_extraction}
    \e{c} := \psi\big(\g{c}, \pi_c(q)\big),
\end{equation}
where $\pi_c$ projects $q$ onto the coordinates of the $c$'th feature grid $\g{c}$ and $\psi$ denotes (bi/tri)linear interpolation of a point into a regularly spaced grid. 
The resulting feature $\e{c}$ is a vector of length $d_1$ if $c\in \{1, 2, 3\}$, length $d_2$ if $c\in \{12, 13, 23\}$ or length $d_3$ if $c = 123$.
We repeat this projection and interpolation procedure over each of our line, plane, and volume feature grids, and combine the resulting feature vectors by any combination of elementwise multiplication ($\circ$), addition ($+$), and concatenation ($\odot$) along the feature dimension. Note that multiplication and addition require $d_1 = d_2$.
Finally, the spatially-localized combined feature vector is decoded using an MLP-based decoder $\decoder{}$. The decoder can take as input both the feature vector arising from the feature grids as well as possible auxiliary inputs, such as (positionally encoded) viewing direction or 3D coordinates, depending on the task.

We consider any model that fits the above description to fall into the \modelname{} family. The specific models we use for nonconvex, semiconvex, and convex optimization are illustrated in \Cref{fig:method}. Any \modelname{} model can be trained by nonconvex optimization; any \modelname{} model that avoids elementwise multiplication of features can be made semiconvex or fully convex as long as its training objective is also convex. 
For example, minimizing mean squared error for a linear inverse problem (like MRI, CT, additive denoising, super-resolution, inpainting, or segmentation) is a convex objective. However, the photometric loss used in radiance field modeling cannot be readily convexified because of the nonlinear accumulation of light along rays due to occlusion---so for our radiance field experiments we use a nonconvex \modelname{} model.

Our experiments focus primarily on two specific \modelname{} models that exemplify some of the strongest convex and nonconvex representations in the \modelname{} family.
For our experiments including convex optimization, namely 3D segmentation with 2D or 3D supervision, and video segmentation, we use the following \modelname{} model (illustrated in the second and third rows of \Cref{fig:method}) which can be trained by either convex, semi-convex, or nonconvex optimization as described in the following subsections:
\begin{equation}
\label{eq:withconcatentation}
    \decoder[\eone \odot \etwo \odot \ethree \odot \eonetwo \odot \eonethree \odot \etwothree \odot \eonetwothree].
\end{equation}
Here we use $\odot$ to denote concatenation of features.
For our radiance field experiments, we use the following nonconvex member of the \modelname{} family (illustrated with multiresolution feature grids in the first row of \Cref{fig:method}):
\begin{equation}
\label{eq:withmultiplication}
    \decoder[(\eone\circ\etwo\circ\ethree) \odot (\eone\circ\etwothree) \odot (\etwo\circ\eonethree) \odot (\ethree\circ\eonetwo) \odot \eonetwothree],
\end{equation}
which leverages geometric algebra to multiply ($\circ$) lower-dimensional (vector and bivector) features together into 3D volume (trivector) features, but cannot be convexified because of this multiplication.
We use multi-resolution copies of the line and plane feature grids $\gone, \gtwo, \gthree, \gonetwo, \gonethree, \gtwothree$, but only a single resolution for the volume grid $\gonetwothree$ since it is already at lower resolution.
Note that the precise architecture of the decoder $\decoder{}$ may vary depending on the specific modeling task, such as whether the quantity of interest is view-dependent. We use the notation $\decoder{}$ to denote any feature decoder that uses a combination of linear layers and ReLU nonlinearities. For our nonconvex experiments the decoder is a standard fully-connected ReLU neural network; decoder details for our semiconvex and convex models are presented in the following subsections.

\subsection{Semiconvex \modelname{}}

For our segmentation experiments (with volumes and videos), we use the \modelname{} architecture in \cref{eq:withconcatentation}, with concatenation instead of multiplication of features; we denote this concatenated feature vector as $f(q)$, the input to the decoder. 
Our semiconvex formulation of this model uses a convex MLP as the decoder:
\begin{gather}
    \tilde{y}(q) = \sum_{i=1}^{h} {({{{W}}_i}^\top f(q))}
    \mathds{1}[{\overline{{W}}_i}^\top f(q)  \geq 0].
\end{gather}
Here ${{{W}}}$ denotes the trainable hidden layer MLP weights, and ${\overline{{W}}}$ denotes the same weights frozen at initialization inside the indicator function. The indicator function, denoted as $\mathds{1} (*)$, returns 1 if the argument is true, and 0 otherwise.
Although this MLP decoder is fully convex, we refer to this model as semiconvex (in particular biconvex; see \citep{mertsemiconvex}) because the combined grid features $f(q)$ are multiplied by the trainable MLP hidden layer weights ${{{W}}}$, though the objective is separately convex in each of these parameters.

\subsection{Convex \modelname{}}

For our segmentation experiments (with volumes and videos), we also present a fully convex \modelname{} model that is similar to the semiconvex model described above, except that we fuse the learnable weights of the MLP decoder with the weights of the feature mapping, to remove the product of parameters (which is semiconvex but not convex). 
Our convex model is:
\begin{align}
\tilde{y}(q) = \sum_{c\in\{1,2,3,12,13,23,123\}} \mathds{1}_{d(c)}^\top (\e{c} \circ \mathds{1}\left[ \ebar{c} \geq 0\right]) ,
\end{align}
where the features $\e{c}$ are interpolated from optimizable parameter grids with feature dimension $d(c) \in \{d_1, d_2, d_3\}$, whereas the gating variables $\ebar{c}$ inside the indicator function are derived from the same grids frozen at their initialization values to preserve convexity. Here $\circ$ denotes elementwise (Hadamard) product of vectors. These indicator functions take the same role as the ReLU in a nonconvex MLP, using a sampling of random activation patterns based on the grid values at initialization.

\vspace{-0.2cm}
\section{Theory}

\subsection{Equivalence to Matrix Completion in 2D}

In three dimensions, the complete set of geometric algebra feature grids are those that we include in the \modelname{} family: $\{\gone, \gtwo, \gthree, \gonetwo, \gonethree, \gtwothree, \gonetwothree\}$.
In two dimensions, the complete set of geometric algebra feature grids is: $\{ \gone, \gtwo, \gonetwo \}$. 
In this two-dimensional setting, we can analyze different members of our \modelname{} family, and show equivalence to various formulations of the classic matrix completion problem. 

\paragraph{Notation.} 
As usual, we use $\circ$ to denote elementwise multiplication and $\odot$ to denote concatenation. We use $\mathds{1}_{a \times b}$ to denote the all-ones matrix of size $a \times b$ and $\mathds{1}[\cdot]$ to denote the indicator function, which evaluates to 1 when its argument is positive and 0 otherwise. 
Our theorem statements consider equivalence to a matrix completion problem in which $M \in \R^{m\times n}$ is the target matrix and $U \in \R^{m\times k}, V \in \R^{n \times k}$ are the low-rank components to be learned. 
We include theoretical results for 2D \modelname{} models combine features by addition ($+$), multiplication ($\circ$), or concatenation ($\odot$) and decode features using a linear decoder (as a warmup), a convex MLP, or a nonconvex MLP. The most illuminating results are presented in the theorem statements that follow; the rest (and all proofs) are deferred to \Cref{sec:appendix-general-proof}.

\paragraph{Assumptions.}
Our theorem statements assume that the line feature grids have the same spatial resolution as the target matrix, and thus do not specify the type of interpolation. 
However, the results hold exactly even if the dimensions do not match, and nearest neighbor interpolation is used; the empirical performance is similar or even slightly improved in practice by using (bi)linear interpolation of features (see \Cref{sec:interpolationcomparison} for a comparison and \Cref{sec:appendix-general-proof} for a generalization to other interpolation methods). 
The theorems assume that the optimization objective is to minimize the Frobenius norm of the error relative to the target matrix; this is equivalent to minimizing mean squared error measured directly in the representation space. In particular, this objective function is the one we use for our convex experiments (video and volume segmentation fitting), where we have access to direct supervision; our radiance field experiments instead use indirect measurements (along rays) that are not exactly equivalent to the setting of the theorems. 

\begin{theorem}
\label{thm:triplane}
    The two-dimensional representation $\decoder{(\eone + \etwo)}$ with linear decoder $\decoder{(f(q))} = \alpha^T f(q)$ is equivalent to a low-rank matrix completion model with the following structure:
    \begin{align}
    \label{eq:suboptimallowrank}
    \min_{U,V} \| M - (U\mathds{1}_{k\times n} + \mathds{1}_{m\times k} V^T) \|_F^2 .
    \end{align}
    These two models are equivalent in the sense that $U^* = \gone^* \text{diag}(\alpha^*)$ and $V^* = \gtwo^* \text{diag}(\alpha^*)$ where $U^*, V^*$ is the optimal solution to the low-rank matrix completion problem in \cref{eq:suboptimallowrank} and $\gone^*, \gtwo^*, \alpha^*$ are the optimal grid features and linear decoder for the  $\decoder{(\eone + \etwo)}$ model.

    The two-dimensional representation $\decoder{(\eone \circ \etwo)}$ with the same linear decoder is equivalent to the standard low-rank matrix completion model:
    \begin{align}
    \label{eq:lin-mult}
    \min_{U,V} \| M - UV^T \|_F^2 .
    \end{align}
    These two models are equivalent in the same sense as above, except that $V^* = \gtwo^*$.
\end{theorem}

\paragraph{Remark.}
Using a linear decoder reveals a dramatic difference in representation capacity between feature addition (or concatenation) and multiplication. Using addition, the maximum rank of the matrix approximation is $2$ regardless of the feature dimension $k$. Using multiplication, the maximum rank of the approximation is $k$.
With feature multiplication, the optimal values of the feature grids are identical to the rank-thresholded singular value decomposition (SVD) of $M$, where the feature grids $\gone$ and $\gtwo$ recover the left and right singular vectors and the decoder $\alpha$ learns the singular values of $M$. This is the optimal rank $k$ approximation of a matrix $M$.

\begin{theorem}
\label{thm:gaplane}
    The two-dimensional representation $\decoder{(\eone + \etwo + \eonetwo)}$ with linear decoder $\decoder{(f(q))} = \alpha^T f(q)$ is equivalent to a low-rank plus low-resolution matrix completion model with the following structure:
    \begin{align}
    \label{eq:suboptimallowranklowres}
    \min_{U,V,L} \| M - (U\mathds{1}_{k\times n} + \mathds{1}_{m\times k} V^T + \varphi(L)) \|_F^2 ,
    \end{align}
    where $L \in \R^{m_l \times n_l}$ is the low-resolution component to be learned, with upsampling (interpolation) function $\varphi$.
    These two models are equivalent in the sense that $U^* = \gone^* \text{diag}(\alpha^*)$, $V^* = \gtwo^* \text{diag}(\alpha^*)$, and $L^* = \gonetwo^* \alpha^*$,
    where $U^*, V^*, L^*$ is the optimal solution to the low-rank plus low-resolution matrix completion problem in \cref{eq:suboptimallowranklowres} and $\gone^*, \gtwo^*, \gonetwo^*, \alpha^*$ are the optimal grid features and linear decoder for the  $\decoder{(\eone + \etwo + \eonetwo)}$ model.

    The two-dimensional representation $\decoder{(\eone \circ \etwo + \eonetwo)}$ with the same linear decoder is equivalent to a low-rank plus low-resolution matrix completion model:
    \begin{align}
    \label{eq:optimallowranklowres}
    \min_{U,V,L} \| M - (UV^T + \varphi(L)) \|_F^2 .
    \end{align}
    These two models are equivalent in the same sense as above, except that $V^* = \gtwo^*$.
\end{theorem}

\paragraph{Remark.} \Cref{thm:gaplane} describes the behavior of a 2D, linear-decoder version of our \modelname{} model, both the version with addition/concatenation of features (\cref{eq:withconcatentation}) and the version with multiplication of features (\cref{eq:withmultiplication}).
Extending the same idea to 3D, we can interpret \modelname{} as a low-rank plus low-resolution approximation of a 3D tensor (volume). 
We can understand this model as first fitting a low-resolution volume and then finding a low-rank approximation to the high-frequency residual volume. 
When we use multiplication of features, the low-rank residual approximation is optimal and analogous to the rank-thresholded SVD.




In \Cref{thm:triplane-convex-mlp} and \Cref{thm:triplane-mlp} we extend the analysis of this two-dimensional \modelname{} model with line features to the version with a convex and nonconvex MLP decoder, respectively. We again derive analogies to matrix completion, but find that the introduction of an MLP decoder, whether convex or nonconvex, fundamentally alters the rank constraints faced by the model.

\begin{theorem}
\label{thm:triplane-convex-mlp}
    The two-dimensional representation $\decoder{(\eone \circ \etwo)}$ with a two-layer convex MLP decoder $\decoder{(f(q))} = \sum_{i=1}^{h} {({{{W}}_i}^\top f(q))}
    \mathds{1}[{\overline{{W}}_i}^\top f(q)  \geq 0]$ is equivalent to a masked low-rank matrix completion model:
    \begin{align}
    \label{eq:masked_svd}
    \min_{U,V,W} \Big\| M -  \sum_{i,j}{ {W_{i,j} {U}_{j} {V}_{j}^\top } \circ B_i} \Big\|_F^2 ,
    \end{align}
    where $W \in \R^{h \times k}$ contains the trainable weights of the convex MLP decoder, with indices $j=1,\dots,k$ for the input dimension and $i=1,\dots,h$ for the hidden layer dimension. $B_i \in \R^{m\times n}$ denotes the binary masking matrix formed by random, fixed gates of the convex MLP decoder; $B_i = \mathds{1}[\sum_{j} \overline{W}_{i,j} {U}_{j} {V}_{j}^\top \geq 0]$, where $\overline{W}$ denotes the weight matrix $W$ with values fixed at random initialization. 
    
    This matrix completion model and our \modelname{} model $\decoder{(\eone\circ\etwo)}$ with convex MLP decoder are equivalent in the sense that $U^* = \gone^*$, $V^* = \gtwo^*$, and $W^* = W^*$, where $U^*, V^*, W^*$ is the optimal solution to the masked low-rank matrix completion problem \cref{eq:masked_svd} and $\gone^*, \gtwo^*, W^*$ are the optimal grid features and convex MLP decoder weights for the $\decoder{(\eone \circ \etwo)}$ model. The optimal mask matrices $B^*_i$ are defined by the fixed random weight initialization $\overline{W}$ and the optimal singular vector matrices $U^*, V^*$. 


\end{theorem}

\paragraph{Remark.} We can interpret the matrix completion model of \Cref{eq:masked_svd} as a sum of $h$ different low-rank approximations, where the matrices within each of the $h$ groups are constrained to share the same singular vectors $U_j, V_j$.
The binary masks $B_i$ effectively allow each of these $h$ low-rank approximations to attend to (or complete) a different part of the matrix $M$ before being linearly combined through the weights (singular values) $W_{i,j}$. The upper limit of the rank of this matrix approximation is thus $\min(n,m)$, because the mask matrices can arbitrarily increase the rank beyond the constraint faced by models with a linear decoder.
Note that if the feature grids $\gone$ and $\gtwo$ have spatial resolution $r_1$ less than $\min(n,m)$, the maximum rank will be $r_1$.

\begin{theorem}
\label{thm:triplane-mlp}
    The two-dimensional representation $\decoder{(\eone \circ \etwo)}$ with a standard two-layer MLP decoder $\decoder{(f(q))} = \alpha^T(W f(q))_+$ is equivalent to a low-rank matrix completion model with the following structure:
    \begin{align}
    \label{eq:optimallowrank}
    \min_{U,V,W,\alpha} \Big\| M -  \sum_{i=1}^h{\alpha_i  \Big( \sum_{j=1}^{k} {W_{i,j} {U}_{j} {V}_{j}^\top } \Big)_+} \Big\|_F^2 ,
    \end{align}
    where $W \in \R^{h\times k}$ is the weight matrix for the MLP decoder's hidden layer (with width $h$) and $\alpha \in \R^h$ is the weight vector of the MLP decoder's output layer. 

    This matrix completion model and our \modelname{} model $\decoder{(\eone\circ\etwo)}$ with nonconvex MLP decoder are equivalent in the sense that $U^* = \gone^*$, $V^* = \gtwo^*$, $W^* = W^*$, and $\alpha^* = \alpha^*$, where $U^*, V^*, W^*, \alpha^*$ is the optimal solution to the masked low-rank matrix completion problem \cref{eq:optimallowrank} and $\gone^*, \gtwo^*, W^*, \alpha^*$ are the optimal grid features and MLP decoder for the  $\decoder{(\eone \circ \etwo)}$ model. 
    
\end{theorem}

\paragraph{Remark.} 
The upper limit of the rank of this matrix approximation is $\min(n,m)$, the same as with a convex MLP decoder. Again, note that if the feature grids $\gone$ and $\gtwo$ have spatial resolution $r_1$ less than $\min(n,m)$, the maximum rank will be $r_1$.

 We summarize the maximum attainable ranks of different 2D models in \Cref{tab:ranks} (see \Cref{sec:prooftriplaneconvexmlp} and \Cref{sec:prooftriplanemlp} for matrix representations of $\decoder{(\eone+\etwo)}$ and $\decoder{(\eone \odot \etwo)}$ with convex and nonconvex MLP decoders). 
Experimental validation of these theoretical results on the task of 2D image fitting (compression) is provided in \Cref{fig:2dvalidation}, with a comparison of interpolation schemes in \Cref{fig:interpolationcomparison} in the appendix.
 As expected, we find that a linear decoder model with multiplication dramatically outperforms its additive counterpart, which does not improve with increasing model size. We also find that 2D \modelname{} models with MLP decoders can match or exceed the compression performance of the optimal low-rank representation found by singular value decomposition (SVD), especially when using a nonconvex MLP. This is a testament to the capacity of an MLP decoder to increase representation rank using fewer parameters than a traditional low-rank decomposition, as well as to the resolution compressibility of natural images.
 These experimental results also provide complementary information the one-sided bounds on representation rank and fitting error in our theoretical analysis.

\begin{table}[h]
  \centering
  \begin{tabular}{|c|c|c|c|}
    \hline
    Model & Linear decoder & convex MLP decoder & MLP decoder \\ \hline
    $\decoder{(\eone + \etwo)}$    & $2$   & $r_1$ &  $r_1$ \\ \hline
    $\decoder{(\eone \odot \etwo)}$   & $2$   & $r_1$ &  $r_1$ \\ \hline
    $\decoder{(\eone \circ \etwo)}$   & $k$   & $r_1$ & $r_1$  \\ \hline
  \end{tabular}
  \caption{Maximum attainable ranks of different 2D \modelname{} models, using only line features. Here $k$ is the feature dimension and $r_1$ is the spatial dimension of the features, which need never exceed $\min(m,n)$. Replacing a linear decoder with a convex or nonconvex MLP can dramatically increase the rank of the representation.}
  \label{tab:ranks}
\end{table}

\begin{figure}
\floatbox[{\capbeside\thisfloatsetup{capbesideposition={right,top},capbesidewidth=0.3\linewidth}}]{figure}[\linewidth]
{\caption{2D image fitting experiments with the \emph{astronaut} image from SciPy, validating matrix completion analysis summarized in \Cref{tab:ranks}. We compare 2D \modelname{} models of the form  $\decoder{(\eone \circ \etwo)}$ (solid colorful lines) and $\decoder{(\eone + \etwo)}$ (dotted colorful lines) with the optimal low-rank approximation provided by singular value decomposition (solid black line).}
\label{fig:2dvalidation}}
{\includegraphics[width=\linewidth]{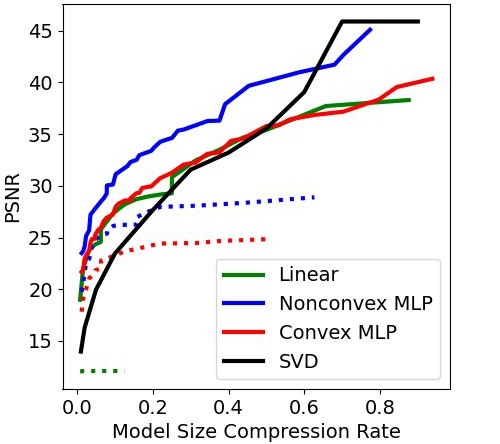}}
\end{figure}

\subsection{Lower Bounds}


Based on the matrix completion theorems and their summary in \Cref{tab:ranks}, we present lower bounds on the Frobenius norm errors of each 2D \modelname{} model. We denote the optimal fitting error of the linear and MLP decoder models by $E_{linear}(\decoder{(f(q))})$ and $E_{MLP}(\decoder{(f(q))})$ for different feature combinations $f(q)$.
For models with a linear decoder,
\begin{align}
E_{linear}(\decoder{(\eone + \etwo)}) &\geq \sigma_2 (M) \\
E_{linear}(\decoder{(\eone \circ \etwo)}) &\geq \sigma_k (M) \\
E_{linear}(\decoder{(\eone \circ \etwo + \eonetwo)}) &\geq \sigma_k (M - \varphi(L^*)) ,
\end{align}
where $L^*$ is a downsampled version of the target $M$, at the same resolution as the feature grid $\gonetwo$.

For models with convex or nonconvex MLP decoders,
\begin{align}
E_{MLP}(\decoder{(\eone + \etwo)}) &\geq \sigma_{r_1}(M) \\
E_{MLP}(\decoder{(\eone \circ \etwo)}) &\geq \sigma_{r_1}(M) \\
E_{MLP}(\decoder{(\eone \circ \etwo + \eonetwo)}) &\geq \sigma_{r_1} (M - \varphi(L^*)).
\end{align}
We can see from these bounds that the approximation error of a model can be reduced dramatically by the introduction of a convex or nonconvex MLP decoder, depending on the singular value decay of the target image $M$.

\subsection{Interpretation: Low Rank + Low Resolution}

Combining multiple parameterization strategies with complementary representation capacities is a time-honored strategy in signal processing. A classic example is the combination of sparse and low-rank models used to represent matrices in the compressive sensing literature \citep{sparselowrank}. This decomposition tends to work well because the residual error of a low-rank matrix approximation is often itself well-approximated by a sparse set of entries.

As shown in \Cref{thm:gaplane}, we can view the \modelname{} family as following a similar strategy with a combination of low-rank and low-resolution approximations. In this framing, the line and plane features combine to form a low-rank but high-resolution approximation, while the volume grid is a full-rank but low-resolution approximation. This parameterization is generally easier to train because sparse models must either store large numbers of empty values (high memory) or store the locations of nonzero entries and suffer from poorly-conditioned spatial gradients (difficult optimization). Indeed there are volume parameterizations that utilize sparsity, such as point clouds, surface meshes, surfels, and Gaussian splats, but these tend to be more challenging to optimize (e.g. requiring high memory \citep{yu_and_fridovichkeil2021plenoxels} or heuristic updates and good initialization \citep{3Dgaussians}).

\begin{figure}[h]
\centering
\begin{minipage}{\textwidth}
\begin{subcolumns}[\textwidth]
\centering
\nextsubcolumn[0.29\textwidth]
  \subfloat[Pareto Frontiers \\ for Image Compression]{\includegraphics[width=\subcolumnwidth]{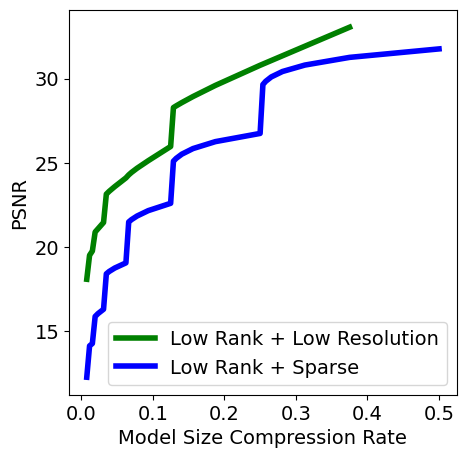}}
\nextsubcolumn[0.29\textwidth]
  \subfloat[Low Rank + Low Res\\PSNR 29.60]{\includegraphics[width=\subcolumnwidth]{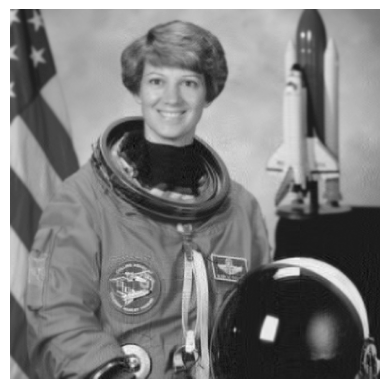}}
\nextsubcolumn[0.29\textwidth]
  \subfloat[Low Rank + Sparse\\PSNR 26.26]{\includegraphics[width=\subcolumnwidth]{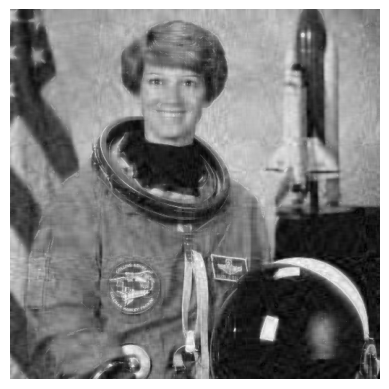}}
\end{subcolumns}
\end{minipage}
\caption{For a natural image, approximation as a sum of low rank and low resolution components (green points and subfigure b) achieves higher fidelity compared to the classic matrix decomposition as a sum of low rank and sparse components (blue points and subfigure c), with the same parameter budget (18.75\% of the original image size, for subfigures b and c). The \modelname{} model family generalizes the idea of a low rank plus low resolution approximation to three dimensions.}
\label{fig:imagepsnr}
\end{figure}

We illustrate this low-rank plus low-resolution interpretation in \Cref{fig:imagepsnr} with a simple experiment, in which we approximate a grayscale image (the \emph{astronaut} image from SciPy) using either a sum of low-rank and low-resolution components (similar to the structure of \modelname{}) or the classic sum of low-rank and sparse components. In this experiment we compute the optimal low-rank components of each model type using the SVD, which corresponds to multiplication of features.

\vspace{-0.2cm}
\section{Experiments}
\subsection{Radiance Field Modeling}

\begin{figure}[h]
  \centering
  \begin{minipage}[b]{ 0.3\textwidth}
    \includegraphics[width=\textwidth]{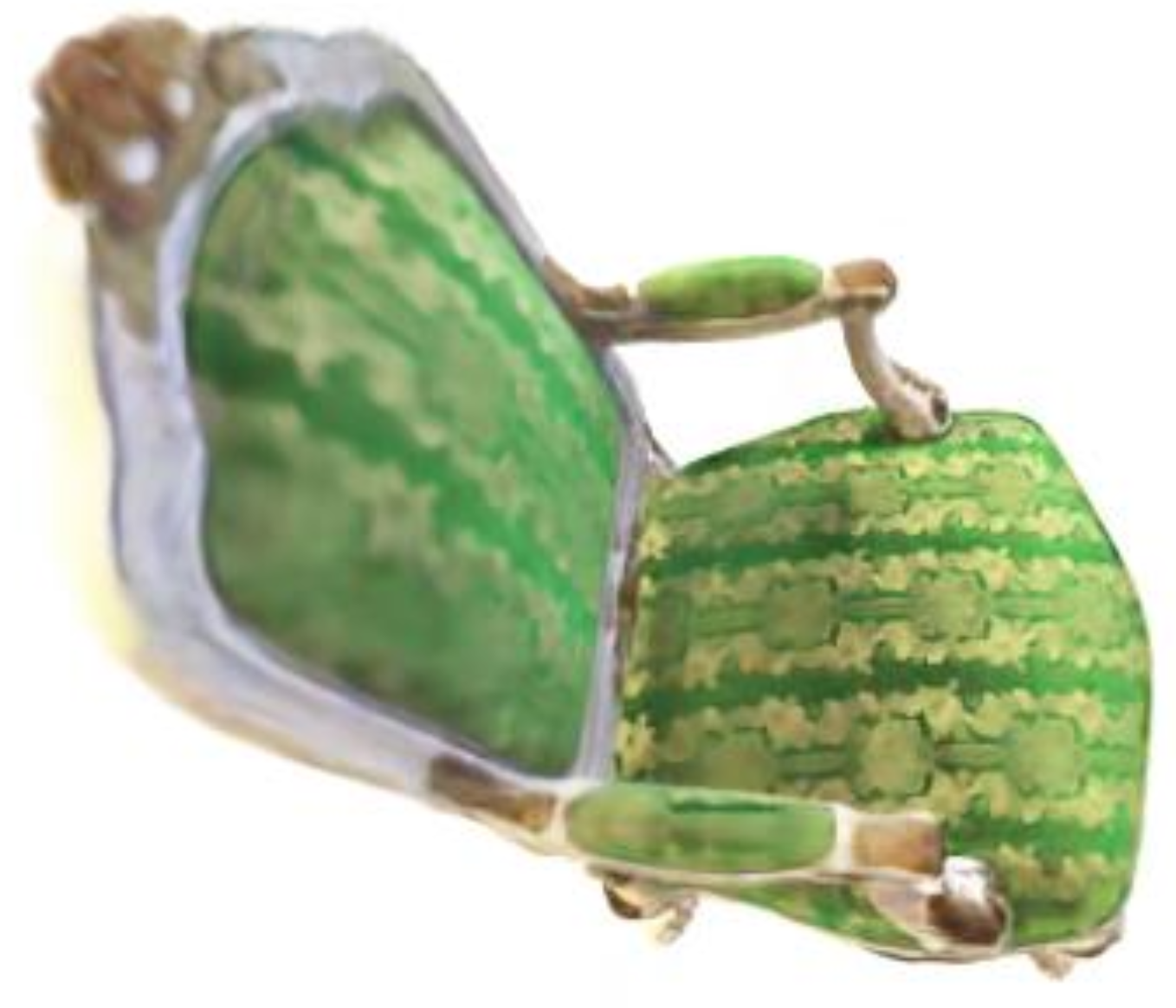}
  \end{minipage}
  \hfill
  \begin{minipage}[b]{ 0.3\textwidth}
    \includegraphics[width=\textwidth]{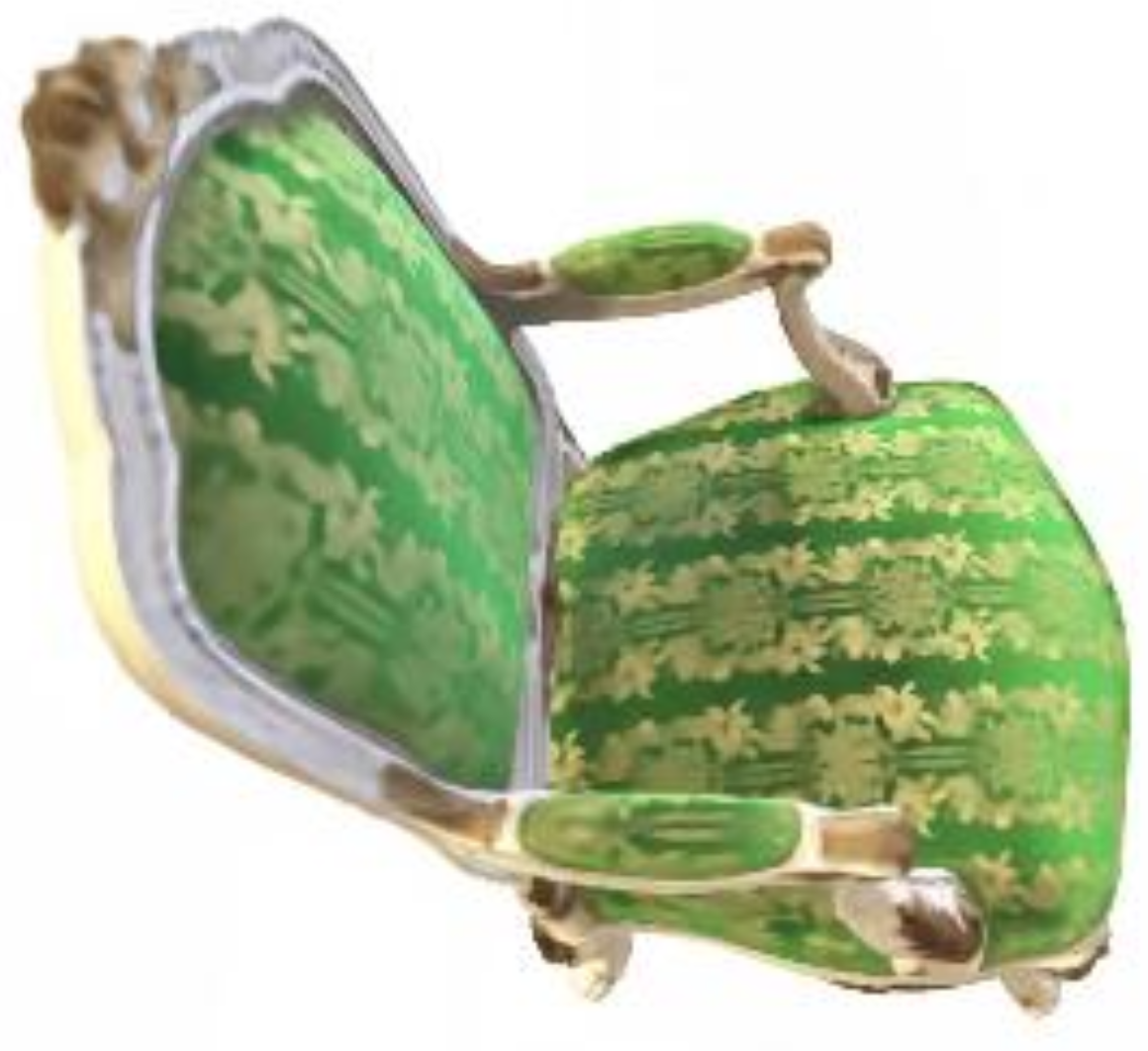}
  \end{minipage}
  \hfill
  \begin{minipage}[b]{ 0.3\textwidth}
    \includegraphics[width=\textwidth]{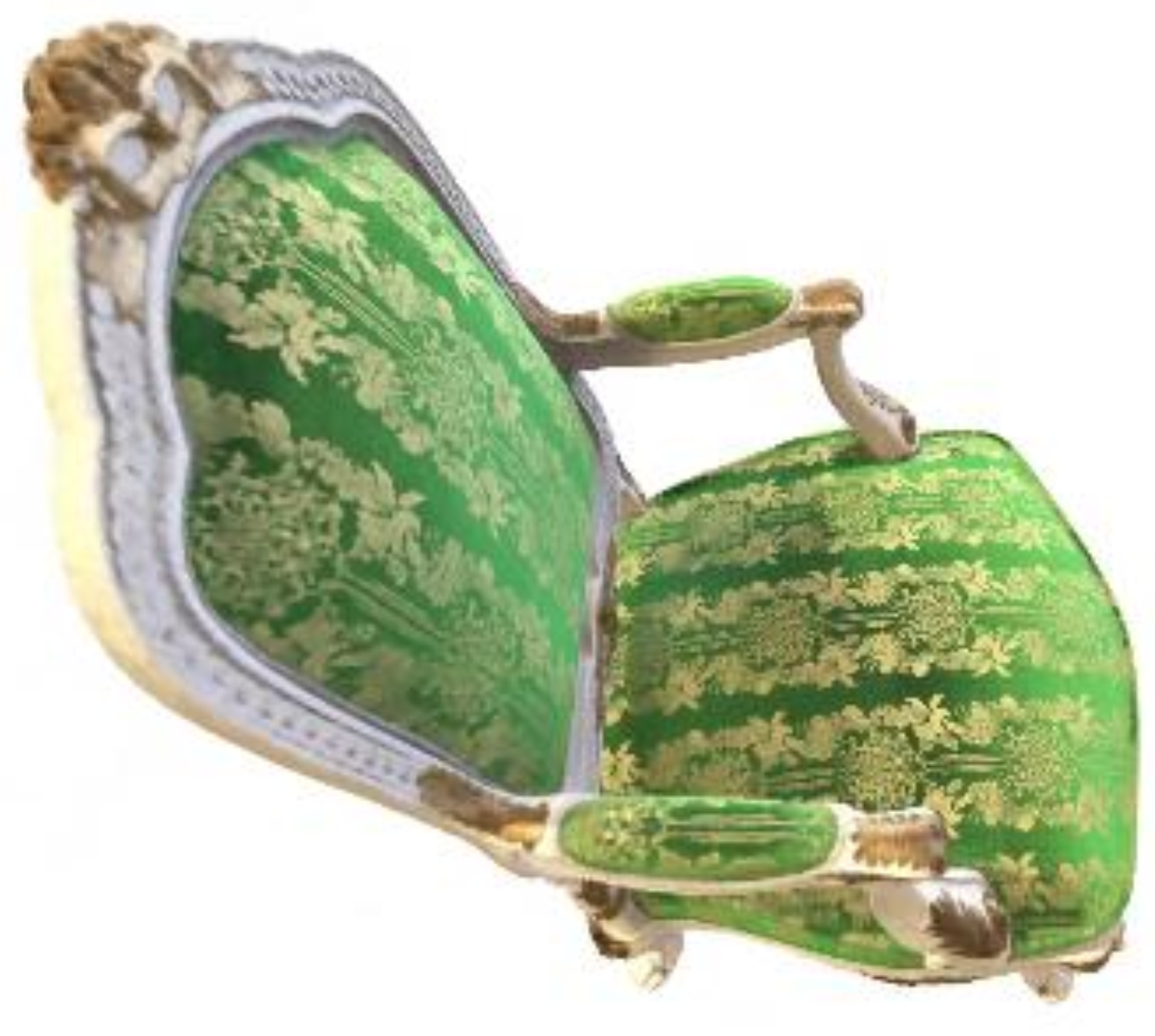}
  \end{minipage}
  \vfill
  \begin{minipage}[b]{ 0.3\textwidth}
    \includegraphics[width=\textwidth]{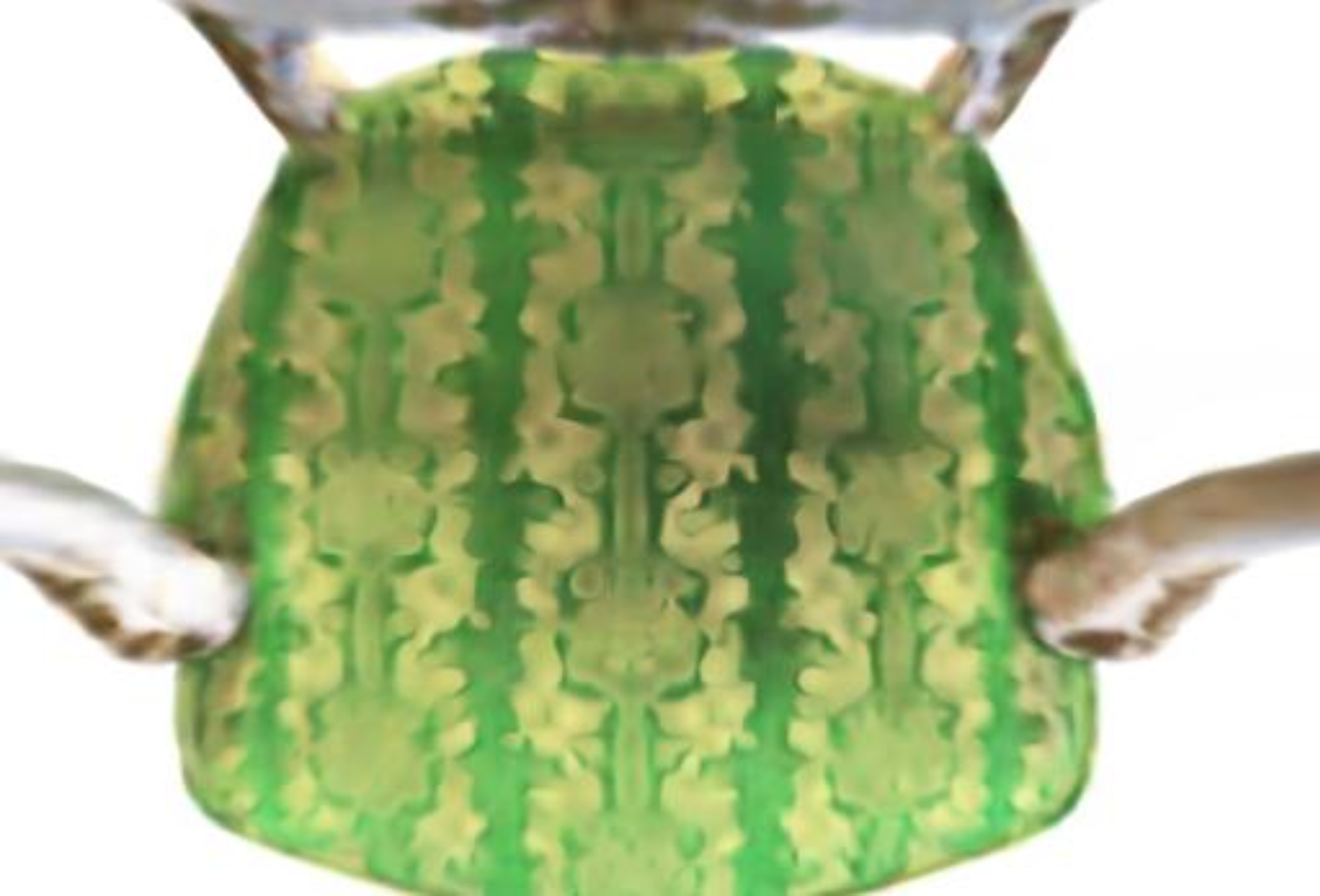}
  \end{minipage}
  \hfill
  \begin{minipage}[b]{ 0.3\textwidth}
    \includegraphics[width=\textwidth]{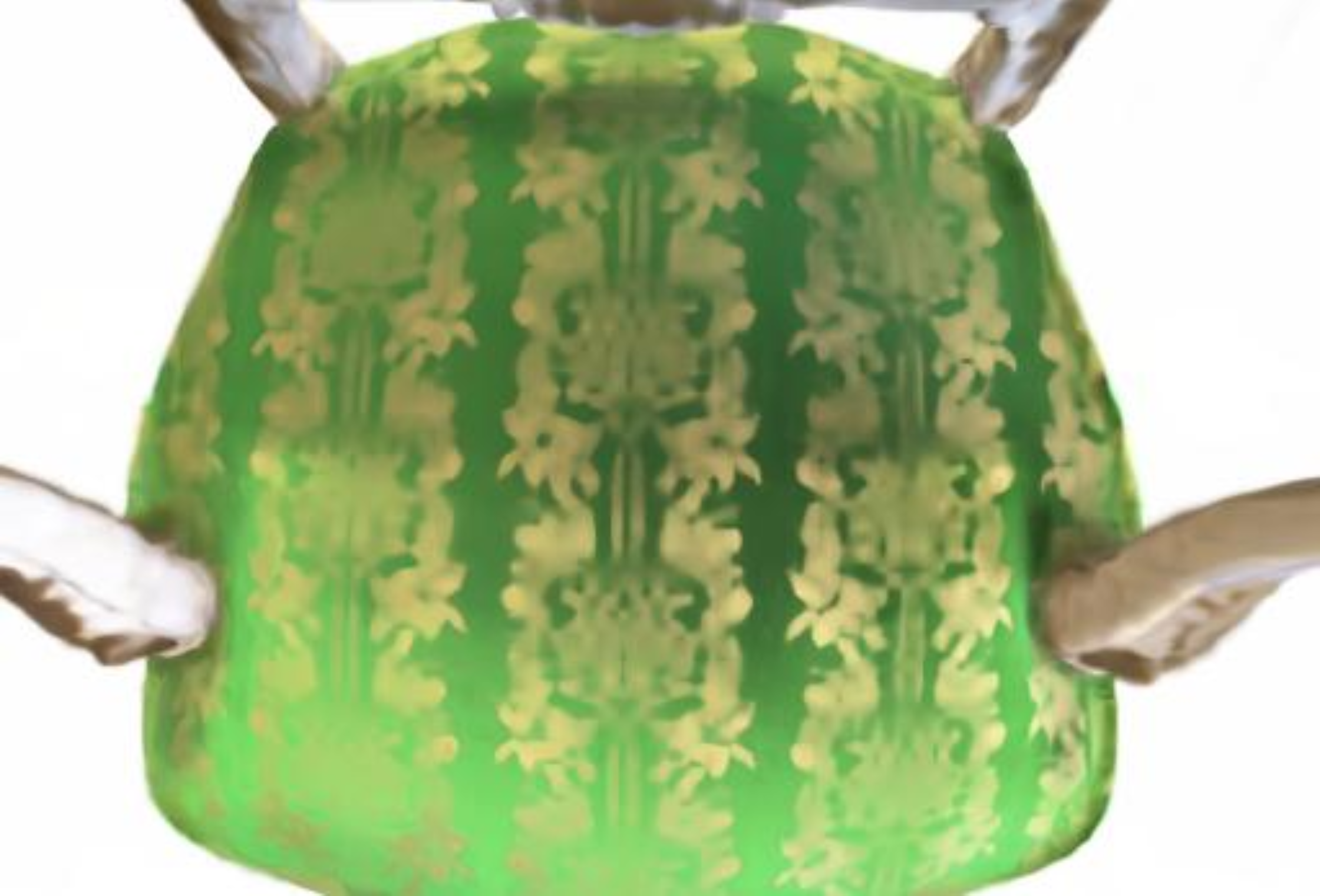}
  \end{minipage}
  \hfill
  \begin{minipage}[b]{ 0.3\textwidth}
    \includegraphics[width=\textwidth]{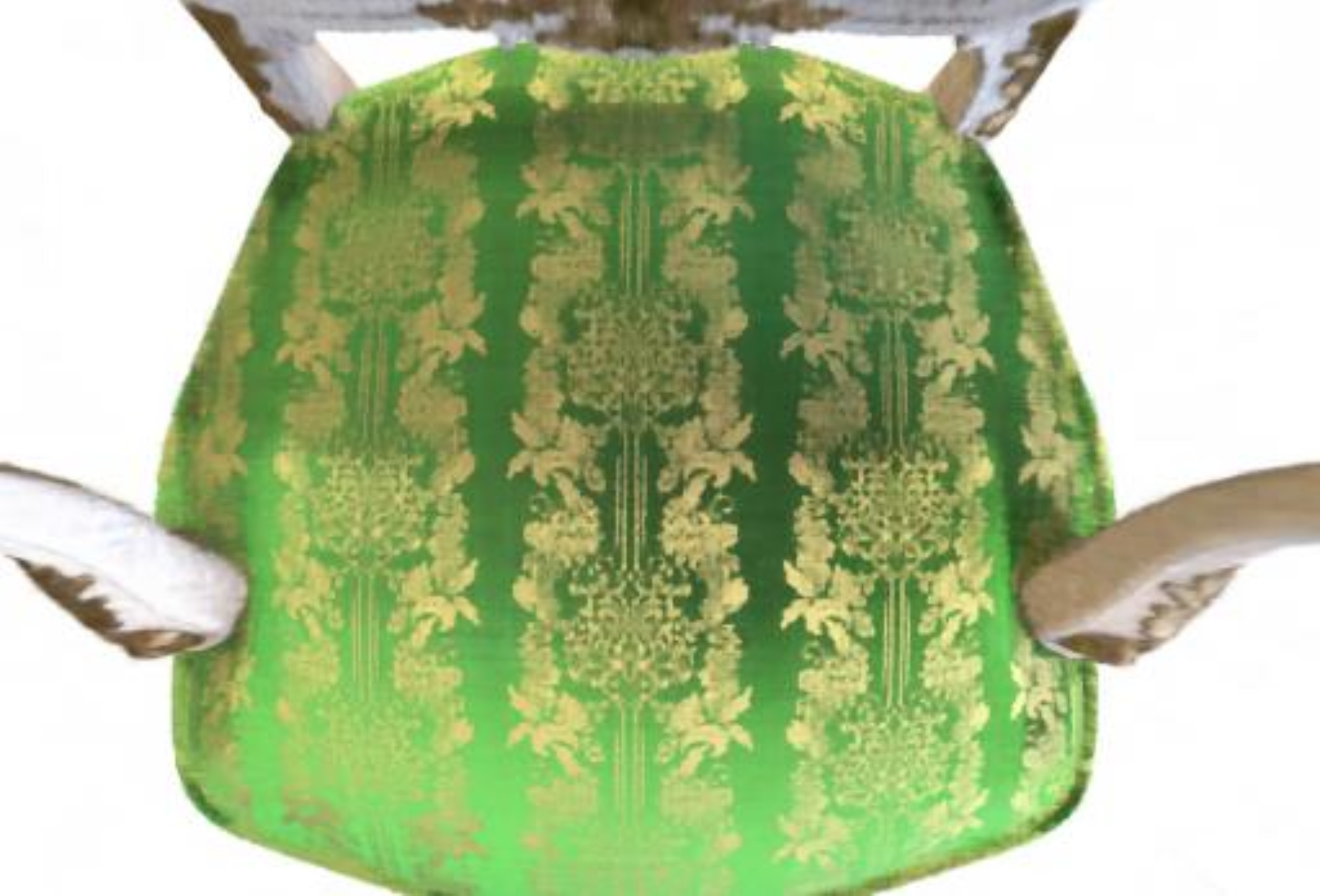}
  \end{minipage}
\caption{Rendering comparison for the \textit{chair} scene: TensoRF on the left (0.32 M parameters), K-Planes in the middle (0.39 M parameters), \modelname{} on the right (0.25 M parameters).}
\label{fig:chair}
\end{figure}

\begin{figure}[h]
  \centering
  \begin{minipage}[b]{ 0.3\textwidth}
    \includegraphics[width=\textwidth]{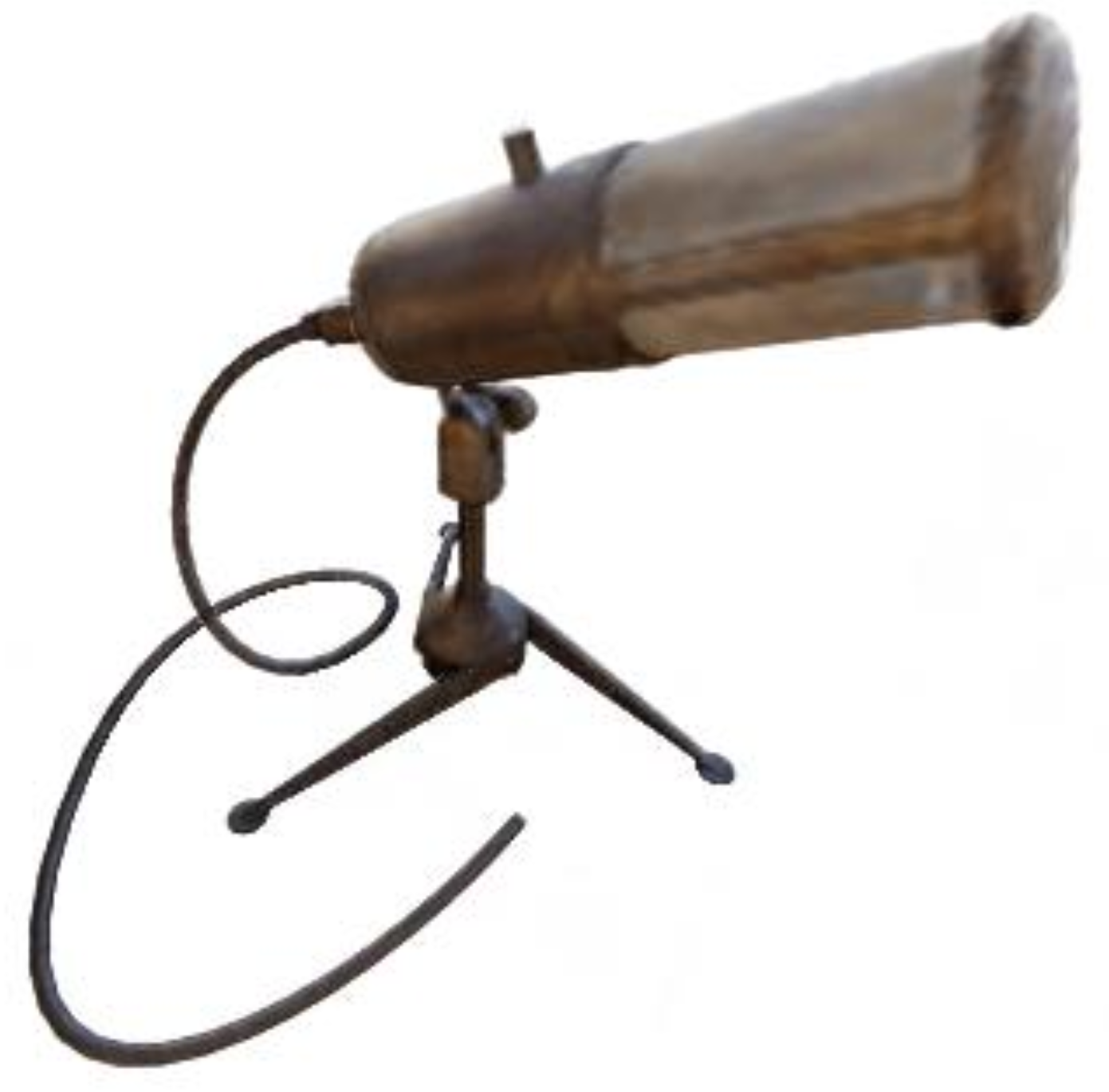}
  \end{minipage}
  \hfill
  \begin{minipage}[b]{ 0.3\textwidth}
    \includegraphics[width=\textwidth]{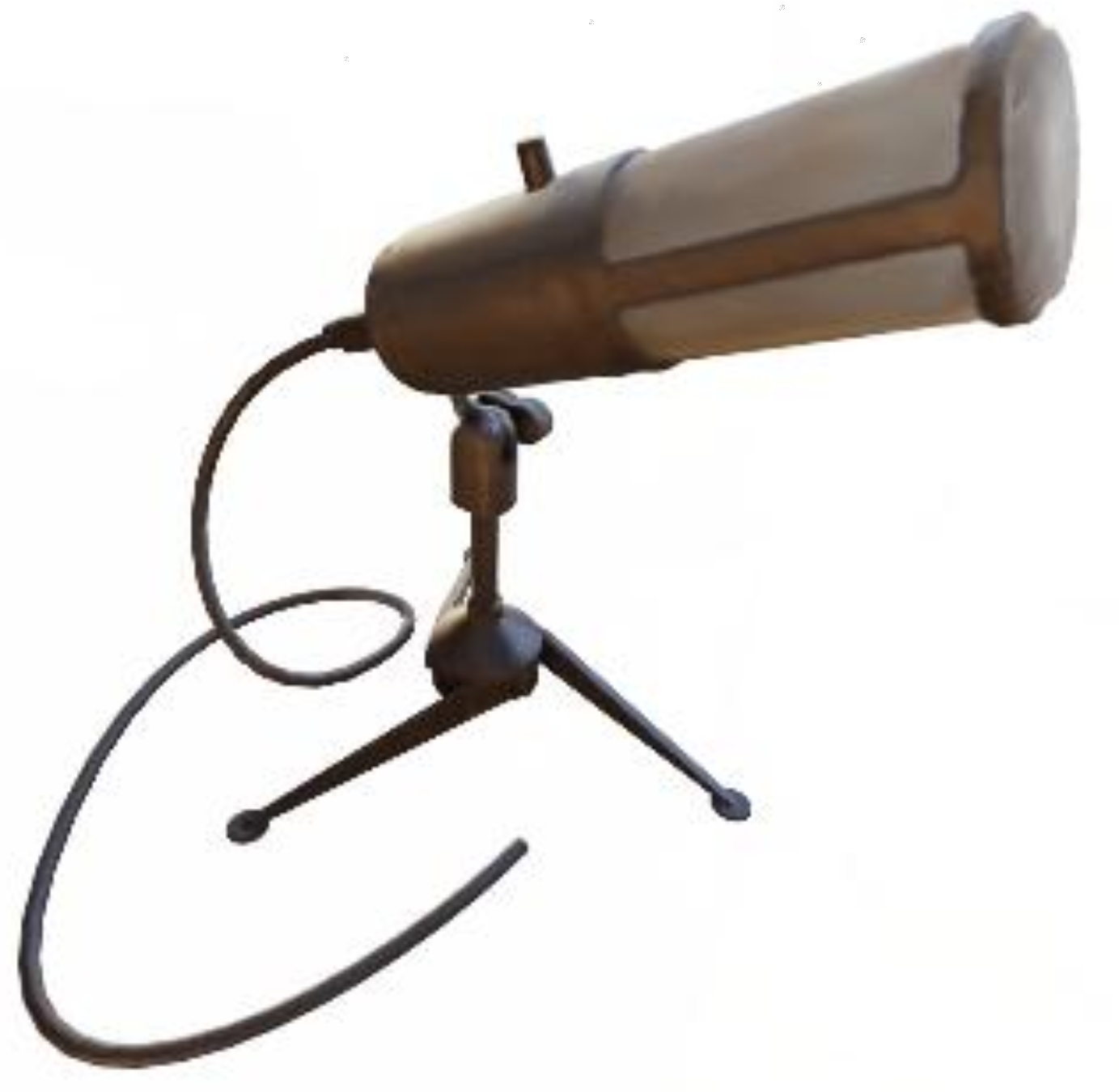}
  \end{minipage}
  \hfill
  \begin{minipage}[b]{ 0.3\textwidth}
    \includegraphics[width=\textwidth]{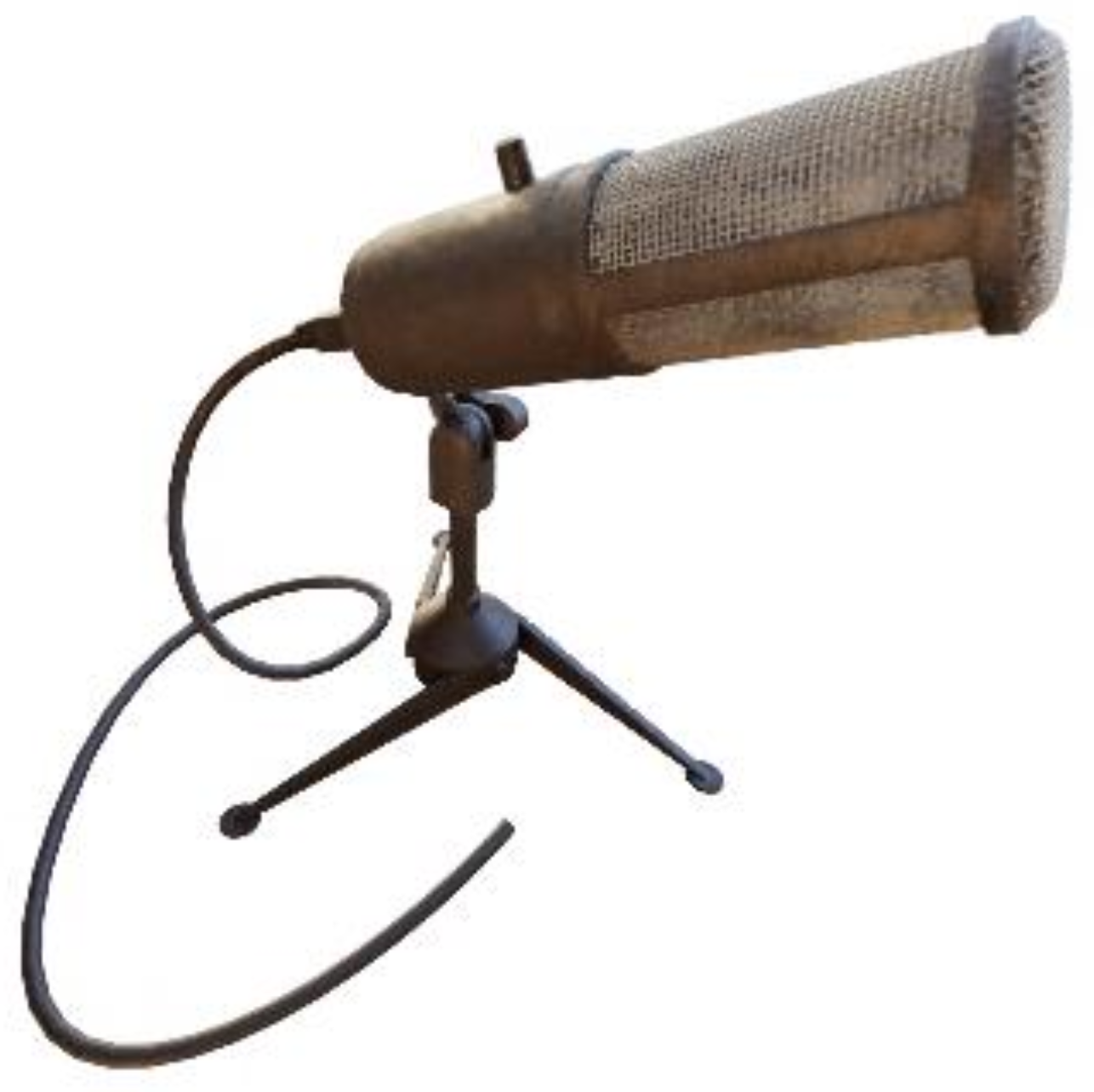}
  \end{minipage}
  \vfill
  \begin{minipage}[b]{ 0.3\textwidth}
    \includegraphics[width=\textwidth]{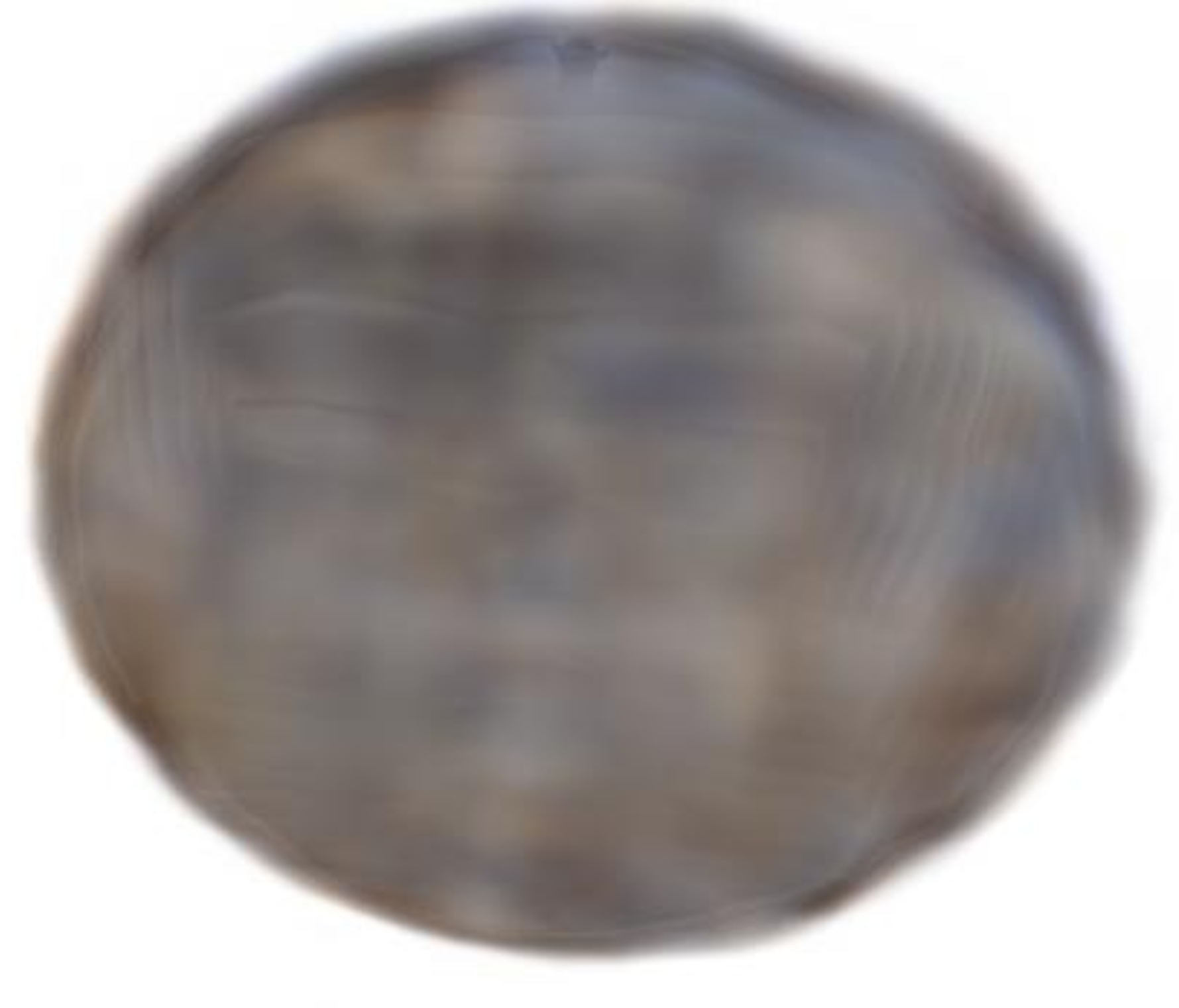}
  \end{minipage}
  \hfill
  \begin{minipage}[b]{ 0.3\textwidth}
    \includegraphics[width=\textwidth]{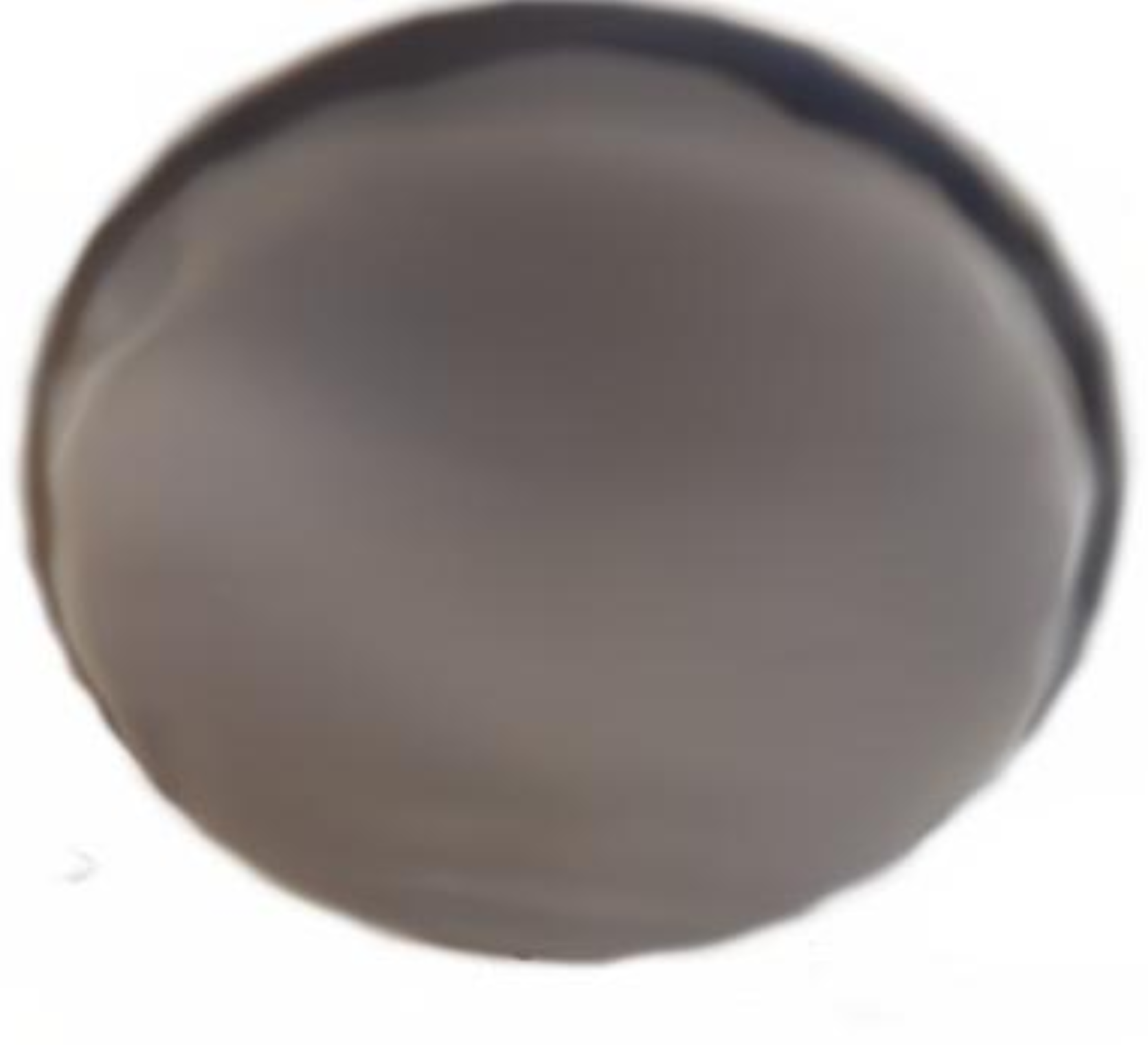}
  \end{minipage}
  \hfill
  \begin{minipage}[b]{ 0.3\textwidth}
    \includegraphics[width=\textwidth]{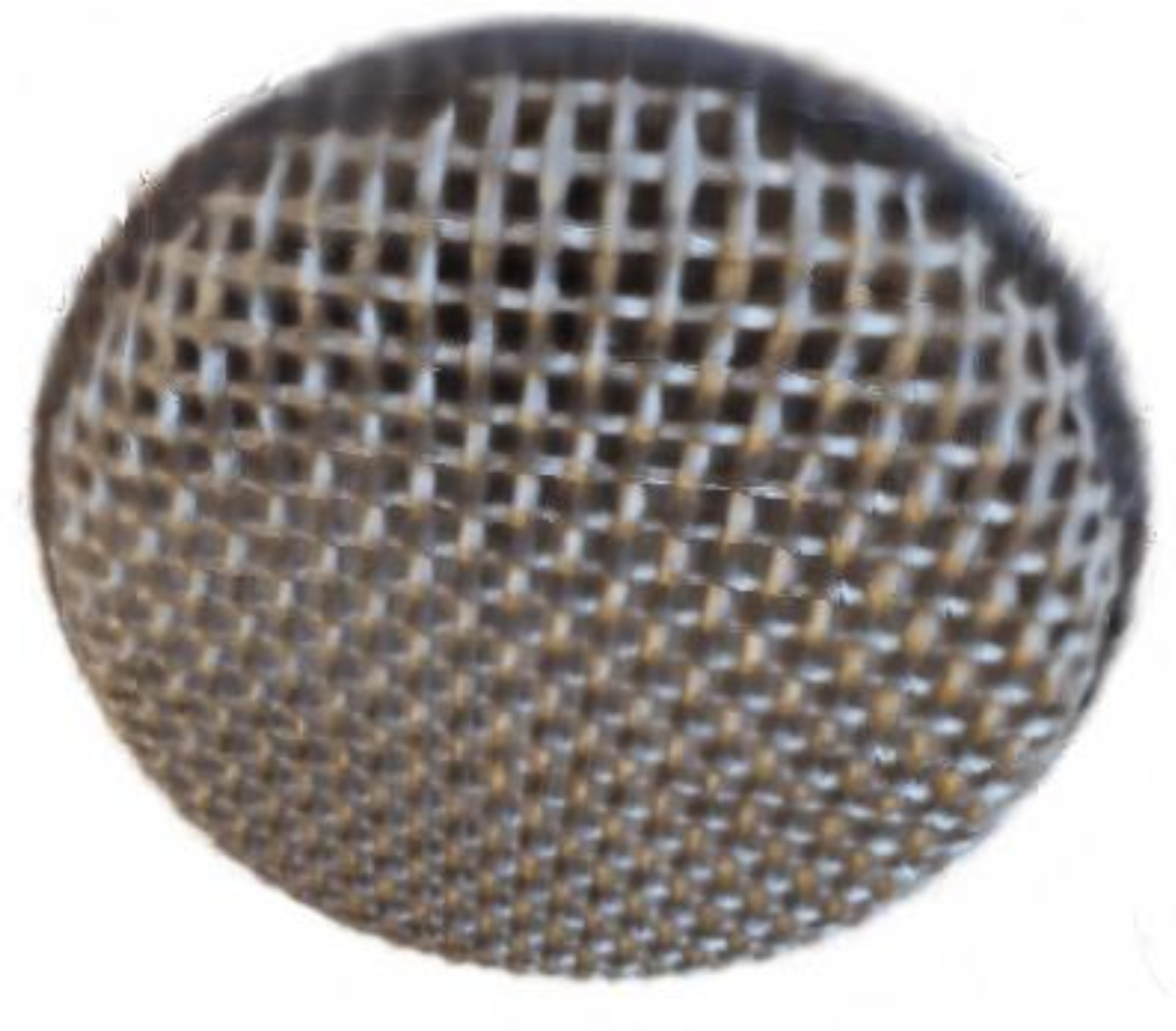}
  \end{minipage}
\caption{Rendering comparison for the \textit{mic} scene: TensoRF on the left (0.32 M parameters), K-Planes in the middle (0.39 M parameters), \modelname{} on the right (0.25 M parameters).}
\label{fig:mic}
\end{figure}

Our experiments for the radiance field reconstruction task are built on the NeRFStudio framework \citep{nerfstudio} and use all scenes from NeRF-Blender \citep{nerf}. For this task, we train each volume representation based on the photometric loss that is standard in the NeRF literature \citep{max, kajiya, nerf}. This loss is the mean squared error at the pixel color level, but is inherently nonconvex because the forward model for volume rendering is nonlinear. Our results are summarized in \Cref{fig:nerfstudio}, which reports PSNR, SSIM \citep{ssim}, and LPIPS \citep{lpips} for each model as a function of its size. We provide example renderings in \Cref{fig:chair} and \Cref{fig:mic}; per-scene pareto-optimal curves and additional renderings are in \Cref{sec:appendixnerf} and \Cref{sec:appendix-render}.

\newcommand{\plotcrop}[1]{%
  \adjincludegraphics[trim={{0.02\width} {0.0\height} {0.1\width} {0.1\height}}, clip, width=0.4\linewidth]{#1}%
}
\newcommand{\plotcroplegend}[1]{%
  \adjincludegraphics[trim={{0.35\width} {0.2\height} {-0.15\width} {0.04\height}}, clip, width=0.4\linewidth]{#1}%
}

\begin{figure}[h]
    \centering
    \plotcrop{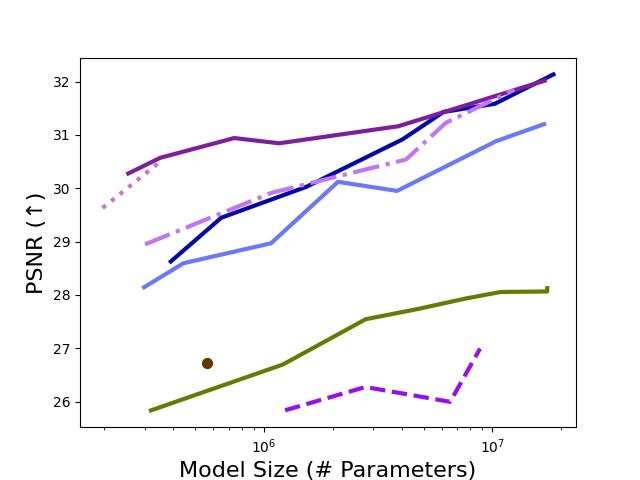}
    \plotcrop{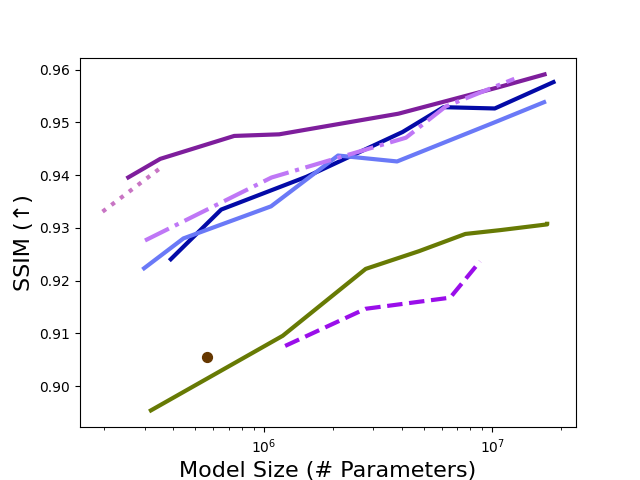}
    \plotcrop{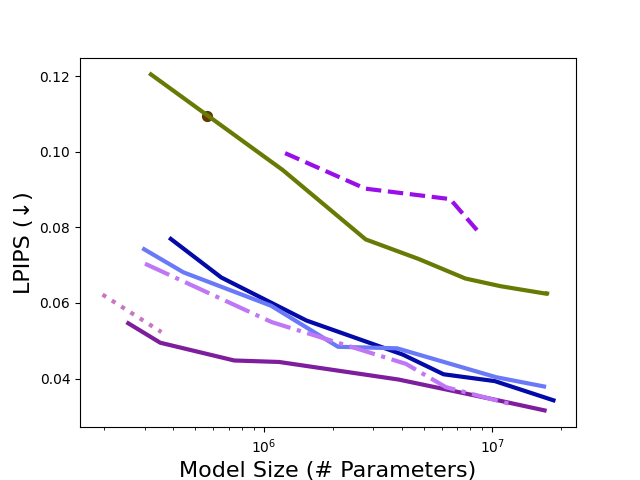}  
    \plotcroplegend{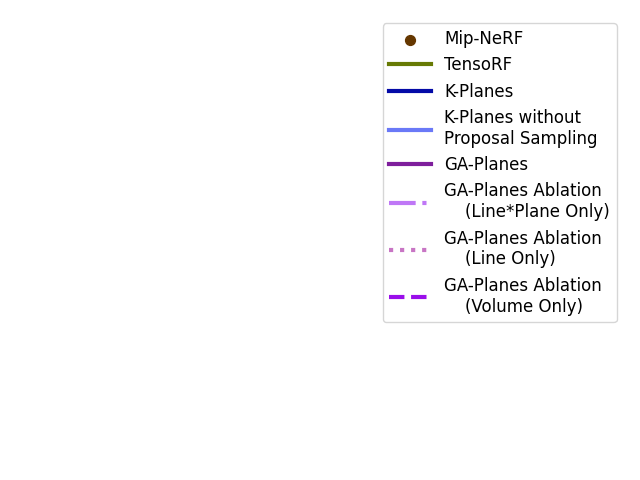}
    \caption{Results on radiance field reconstruction. Nonconvex \modelname{} (with feature multiplication) offers the most efficient representation: when the model is large it performs comparably to the state of the art models, but when model size is reduced it retains higher performance than other models. Here all models are trained for the same number of epochs on all 8 scenes from the Blender dataset, and the average results are shown.}
    \label{fig:nerfstudio}
\end{figure}

Because the loss function is inherently nonconvex for this task, we focus only on nonconvex models and use our most expressive \modelname{} parameterization as defined in \cref{eq:withmultiplication}. We compare this \modelname{} model with several popular models as implemented in NeRFStudio: Mip-NeRF \citep{mipnerf}, TensoRF \citep{tensorf}, and K-Planes \citep{kplanes}. 
For K-Planes, we include versions with and without proposal sampling, a strategy for efficiently allocating ray samples during training and rendering. Proposal sampling is the default for K-Planes, but we include a version without proposal sampling because (1) all other models in this experiment do not use proposal sampling, and (2) we find that proposal sampling boosts the PSNR for K-Planes but hurts its SSIM and LPIPS.
We also include several ablations of our \modelname{} model: one with only the volume features (similar to a multiresolution version DVGO \citep{dvgo} or Plenoxels \citep{yu_and_fridovichkeil2021plenoxels}), one with only the line features (similar to a multiresolution TensoRF-CP), and one with only the line-plane products (similar to a multiresolution TensoRF-VM).
At large model sizes ($\sim$10 million parameters) most models including \modelname{} perform comparably well.
However, as model size shrinks we find that only \modelname{} and its line-only ablation reach comparable metrics as the larger models.

Indeed, we find that because \modelname{} contains features with different dimensionalities (line, plane, and volume), it experiences little loss in quality over a wide range of model sizes. For small model sizes, we allocate most of the model memory to the line features, since their spatial resolution grows linearly with parameter count. 
As the parameter budget grows, we allocate more parameters to the plane features, whereas the performance of the line-only model stagnates with increasing size.
Similarly, as model size grows even further we allocate more parameters to the volume features, whose memory footprint grows cubically with spatial resolution.

\vspace{-0.2cm}
\subsection{3D Segmentation}

Our experiments for the 3D segmentation task use the opacity masks from the NeRF-Blender \emph{lego} scene \citep{nerf}. The task is to ``lift'' 2D segmentation masks to 3D, rather than generating 3D segmentations directly from raw photographs.
We compare the \modelname{} architecture with concatenation of features (described by \cref{eq:withconcatentation}) and the simpler Tri-Plane representation proposed in \cite{eg3d}, in which the three plane features are added together and decoded without use of line or volume features ($\decoder[\eonetwo + \eonethree + \etwothree]$). We train three versions of each model: convex, semiconvex, and nonconvex, to validate that the \modelname{} architecture in \cref{eq:withconcatentation} is more robust than the Tri-Plane model across these different formulations. We also compare \modelname{} with multiplication of features, in \cref{eq:withmultiplication}, to the Tri-Plane model using multiplication of plane features, like K-Planes ($\decoder[\eonetwo \circ \eonethree \circ \etwothree]$). These only have nonconvex formulations.

\paragraph{2D Supervision.}
One set of 3D segmentation experiments relies on 2D supervision, in which we minimize the mean squared error between the ground truth object segmentation masks and the average ray density at each viewpoint. This form of 2D supervision is essentially tomography in the presence of occlusion. 
After training, we render the projections of our reconstructed volume model at each of the test angles, and threshold it to produce a binary object mask. We then compute the intersection over union (IOU) metric to compare our predicted segmentation masks with the ground truth masks. Results are reported in \Cref{tab:2dsupervision}.

\begin{table}[h]
    \centering
    \begin{tabular}{l|c|c|c}
     & Convex & Semiconvex & Nonconvex \\
    \midrule
        \modelname{} (with $\circ$) & - & - & 0.877 \\
        \modelname{} (with $\odot$) & 0.875 & 0.883 & 0.880 \\
        Tri-Planes (planes with $\circ$, like K-Planes) & - & - & 0.877 \\
        Tri-Planes (planes with $+$, like \citep{eg3d})& 0.681 & 0.868 & 0.863
    \end{tabular}
    \caption{Intersection over union (IOU) for recovering novel view object segmentation masks from segmentation mask training with 2D tomographic supervision.}
    \label{tab:2dsupervision}
\end{table}

We find that the \modelname{} architecture with concatenation outperforms the Tri-Planes architecture using addition when trained with equal total number of parameters (see \Cref{sec:configs} for specific feature resolution and dimensions), regardless of whether training is convex, semiconvex, or nonconvex. Further, we see that \modelname{} retains similar performance regardless of optimization strategy, whereas the Tri-Plane model learns poorly via convex optimization. When the features are combined through multiplication, the models achieve the same IOU score.

\paragraph{3D Supervision.}

Our second set of 3D segmentation experiments leverages direct 3D supervision via Space Carving \citep{spacecarving}. 
Space Carving supervision operates on the principle that if any ray passing through a given 3D coordinate is transparent, the density at that 3D coordinate must be zero. This method recovers the visual hull of the object by ``carving out'' empty space around it.
Our results with 3D supervision, in \Cref{tab:3dsupervision}, parallel those with 2D supervision: \modelname{} performs well regardless of whether it is optimized in a convex, semiconvex, or nonconvex formulation, whereas the Tri-Plane model performs decently under nonconvex optimization but much worse with convex or semiconvex formulation.

\begin{table}[h]
    \centering
    \begin{tabular}{l|c|c|c}
     & Convex & Semiconvex & Nonconvex \\
    \midrule
        \modelname{} (with $\circ$) & - & - & 0.926 \\
        \modelname{} (with $\odot$) & 0.932 & 0.957 & 0.964 \\
        Tri-Planes (planes with $\circ$, like K-Planes) & - & - & 0.881 \\
        Tri-Planes (planes with $+$, like \citep{eg3d}) & 0.642 & 0.636 & 0.941
    \end{tabular}
    \caption{Intersection over union (IOU) for recovering novel view object segmentation masks from segmentation mask training with 3D Space Carving supervision.}
    \label{tab:3dsupervision}
\end{table}

\vspace{-0.2cm}
\subsection{Video Segmentation}

Our experiments for the video segmentation task use a variety of nonconvex, semiconvex, and convex models from the \modelname{} family. We treat the video segmentation task as similar to the volume segmentation task with 3D supervision described above: now the volume dimensions are $x, y, t$ rather than $x, y, z$, and the supervision is performed directly in 3D using segmentation masks for a subset of the video frames (every third frame is held out for testing).  
Note that although we refer to this task as video segmentation, the models are essentially tasked with temporal superresolution of object masks in a video rather than predicting masks from a raw video of a moving object. 
Our dataset preparation pipeline uses the skateboarding video and preprocessing steps described at \cite{videoprocessing}, which involves first extracting a bounding box with YOLOv8 \citep{yolo} and then segmenting the skateboarder with SAM \citep{sam1}.

Our results are summarized in \Cref{fig:video} and \Cref{tab:video}.
Similar to the volume segmentation setting, here we again find that \modelname{} performs well across convex, semiconvex, and nonconvex training, though its performance is slightly reduced under fully convex training, perhaps because the convex model size is slightly reduced due to fusing the decoder parameters into the feature grids. In contrast, the simpler Tri-Plane models perform poorly on this task regardless of training strategy: they fail to learn the temporal sequence of the video, producing masks that focus only on the skateboarder's less-mobile core.
In this experiment, we also compare nonconvex models of each type (Tri-Plane and full \modelname{}) in which the features are combined in a linear way (by concatenation or addition) versus a nonlinear way (by multiplication). 
The similar performance between addition/concatenation and multiplication of features observed here is in line with results reported in a similar ablation study in K-Planes \citep{kplanes}, in which the benefits of feature multiplication were only evident when using a linear decoder rather than a nonlinear MLP decoder as is used here.

\newcommand{\plotcropvideo}[1]{%
  \adjincludegraphics[trim={{0.32\width} {0.1\height} {0.28\width} {0.2\height}}, clip, width=0.19\linewidth]{#1}%
}

\begin{figure}[h]
    \raggedright
    \rotatebox{90}{~~~~~~~~~~~\modelname{}}
    \plotcropvideo{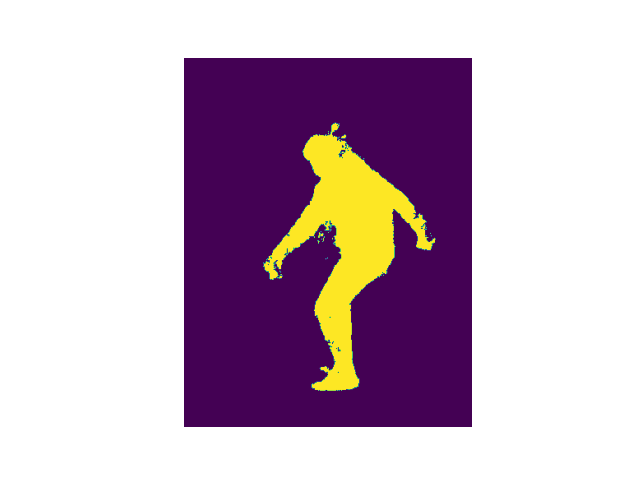}
    \plotcropvideo{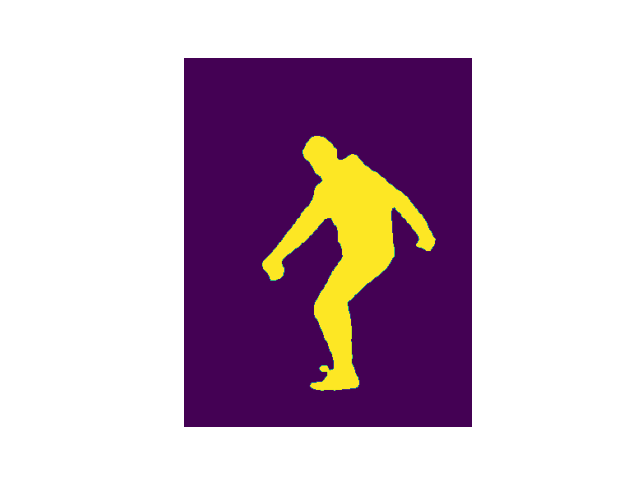}
    \plotcropvideo{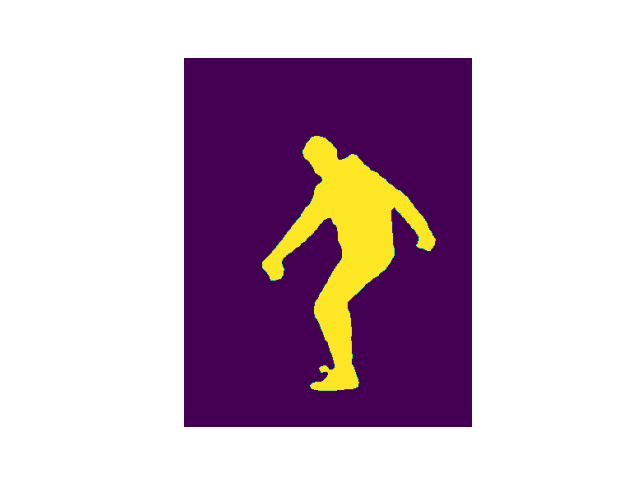}
    \plotcropvideo{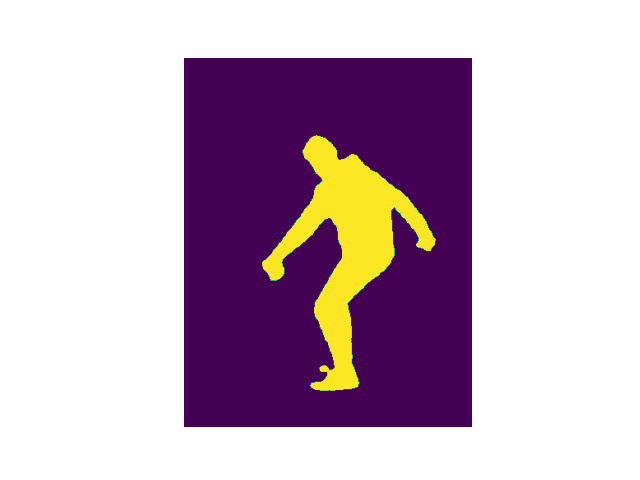}
    \hfill \\
    \rotatebox{90}{~~~~~~~~~~~Tri-Planes}
    \plotcropvideo{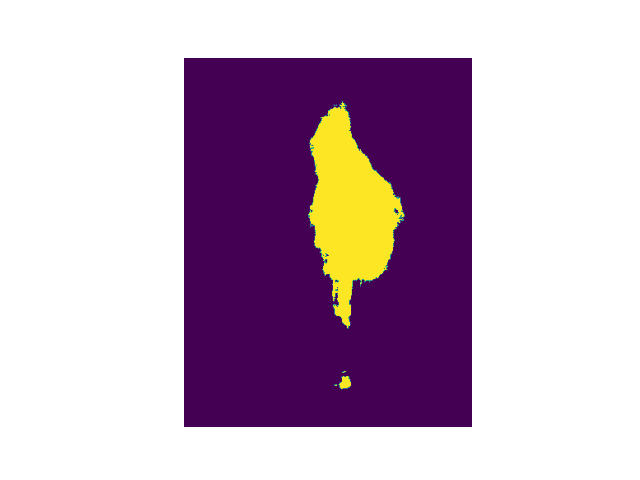}
    \plotcropvideo{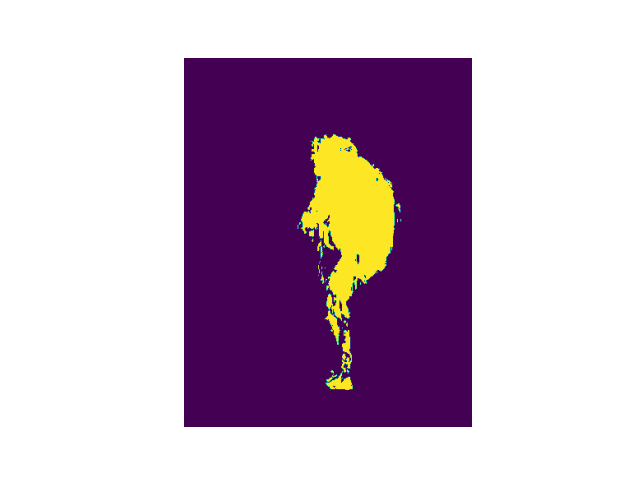}
    \plotcropvideo{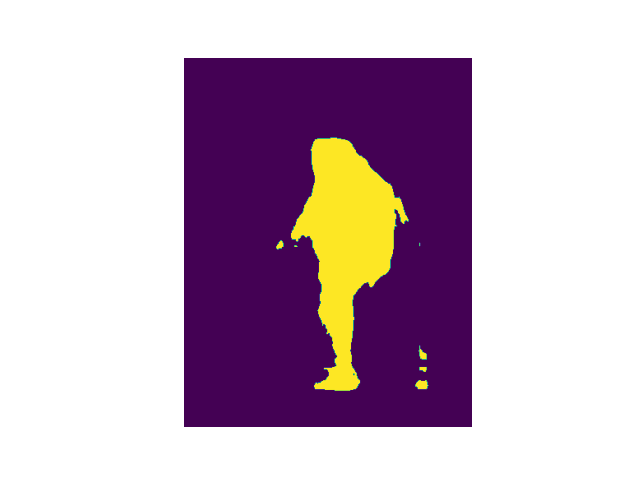}
    \plotcropvideo{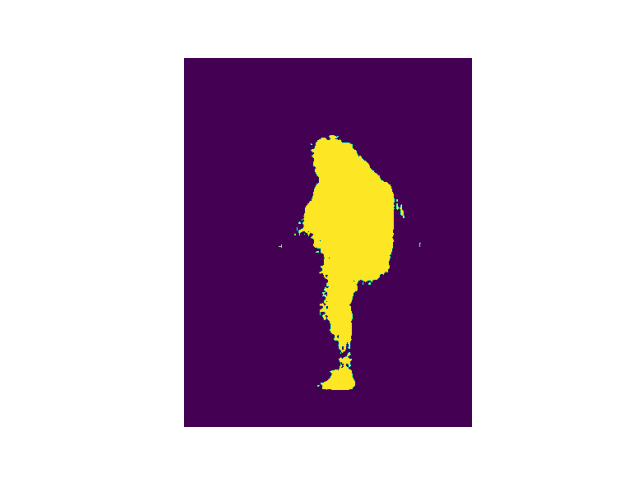}
    \plotcropvideo{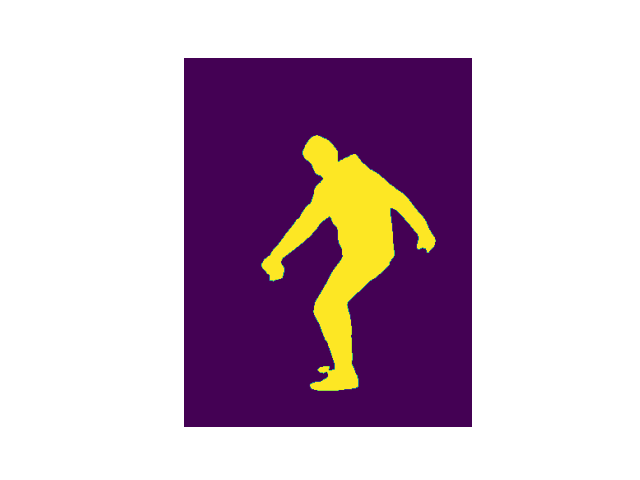}
    \centering
        \begin{tabular}{ccccc}
            ~~~~~~~~~~Convex~~~~~~~~  & ~~Semiconvex~ & Nonconvex ($\odot$ or $+$) & Nonconvex ($\circ$)~ & ~Ground Truth \\
        \end{tabular}

    \caption{Intersection over union (IOU) for predicting segmentation masks for unseen frames within a video of a segmented skateboarder.}
    \label{fig:video}
\end{figure}

\begin{table}[h]
    \centering
    \begin{tabular}{l|c|c|c}
     & Convex & Semiconvex & Nonconvex \\
    \midrule
        \modelname{} (with $\circ$) & - & - & 0.974 \\
        \modelname{} (with $\odot$) & 0.913 & 0.975 & 0.981 \\
        Tri-Planes (planes with $\circ$, like K-Planes) & - & - & 0.727 \\
        Tri-Planes (planes with $+$, like \cite{eg3d}) & 0.557 & 0.647 & 0.732 \\
    \end{tabular}
    \caption{Intersection over union (IOU) for temporal superresolution of segmentation masks in a video, computed on held-out test frames. Models that involve multiplication of features can only be trained by nonconvex optimization.}
    \label{tab:video}
\end{table}
\vspace{-0.4cm}

\section{Discussion}

In this work we introduce \modelname{}, a family of volume parameterizations that generalizes many existing representations (see \Cref{sec:appendixcontext}). 
We specifically focus on two members of the \modelname{} family (with concatenation versus multiplication of features), and offer both theoretical interpretation and empirical evaluation of these models.

Theoretically, we show that these models form a low-rank plus low-resolution tensor factorization. In two dimensions, the \modelname{} model with feature multiplication yields the optimal low-rank plus low-resolution matrix approximation, whereas the model with feature concatenation is likewise low-rank plus low-resolution but may not be optimally parameter-efficient. In other words, 2D \modelname{} with feature multiplication is equivalent to first taking a low-resolution matrix approximation (by filtering and downsampling) and then finding the optimal low-rank approximation to the residual (by thresholding the singular values in the SVD).

For the \modelname{} model with feature concatenation (or addition, but not multiplication), we derive semiconvex and fully convex formulations based on convexifying the MLP decoder and fusing weights as needed.
We empirically demonstrate convex, semiconvex, and nonconvex \modelname{} models are effective on three tasks: radiance field reconstruction, 3D object segmentation (with 2D or 3D supervision), and video segmentation (temporal superresolution of object masks) (see \Cref{sec:configs} for additional details concerning model configurations for each task). 
We find that \modelname{} matches or exceeds the performance of strong baselines across a range of model sizes, and that \modelname{} without multiplication performs similarly well regardless of whether its formulation (with decoder) is nonconvex, semiconvex, or fully convex.

\vspace{-0.2cm}
\paragraph{Limitations.}
In this work we focus on 3D (or smaller) representations, rather than higher dimensions (e.g. dynamic volumes), and we demonstrate \modelname{} for reconstruction tasks rather than also generation tasks. Both of these extensions are promising avenues for extending \modelname{}. In our experiments, we use the same first-order optimization algorithm for all models.  However, our convex GA-Planes formulation is designed to be compatible with any convex solver (e.g. cvxpy), and we expect its performance may improve by leveraging these efficient convex optimization algorithms. We demonstrate preliminary benefits of convexity and semiconvexity in terms of training stability in \Cref{sec:benefitsofconvexity}.

\section*{Acknowledgments}
We are grateful to Axel Levy for helpful discussions on space carving, and to Abhishek Shetty for helpful discussions on matrix completion.
This work was supported in part by the NSF Mathematical Sciences Postdoctoral Research Fellowship under award number 2303178, in part by the National Science Foundation (NSF) under Grant DMS-2134248, in part by the NSF CAREER Award under Grant CCF-2236829, in part by the U.S. Army Research Office Early Career Award under Grant W911NF-21-1-0242, and in part by the Office of Naval Research under Grant N00014-24-1-2164.

\bibliography{iclr2025_conference}

\begin{thebibliography}{47}
\providecommand{\natexlab}[1]{#1}
\providecommand{\url}[1]{\texttt{#1}}
\expandafter\ifx\csname urlstyle\endcsname\relax
  \providecommand{\doi}[1]{doi: #1}\else
  \providecommand{\doi}{doi: \begingroup \urlstyle{rm}\Url}\fi

\bibitem[Barron et~al.(2021)Barron, Mildenhall, Tancik, Hedman, Martin{-}Brualla, and Srinivasan]{mipnerf}
Jonathan~T. Barron, Ben Mildenhall, Matthew Tancik, Peter Hedman, Ricardo Martin{-}Brualla, and Pratul~P. Srinivasan.
\newblock Mip-nerf: {A} multiscale representation for anti-aliasing neural radiance fields.
\newblock In \emph{ICCV}, pp.\  5835--5844. {IEEE}, 2021.
\newblock \doi{10.1109/ICCV48922.2021.00580}.
\newblock URL \url{https://doi.org/10.1109/ICCV48922.2021.00580}.

\bibitem[Cen et~al.(2023)Cen, Zhou, Fang, Yang, Shen, Xie, Jiang, Zhang, and Tian]{cen2023segment}
Jiazhong Cen, Zanwei Zhou, Jiemin Fang, Chen Yang, Wei Shen, Lingxi Xie, Dongsheng Jiang, Xiaopeng Zhang, and Qi~Tian.
\newblock Segment anything in 3d with nerfs.
\newblock In \emph{NeurIPS}, 2023.

\bibitem[Chan et~al.(2022)Chan, Lin, Chan, Nagano, Pan, Mello, Gallo, Guibas, Tremblay, Khamis, Karras, and Wetzstein]{eg3d}
Eric~R. Chan, Connor~Z. Lin, Matthew~A. Chan, Koki Nagano, Boxiao Pan, Shalini~De Mello, Orazio Gallo, Leonidas Guibas, Jonathan Tremblay, Sameh Khamis, Tero Karras, and Gordon Wetzstein.
\newblock Efficient geometry-aware {3D} generative adversarial networks.
\newblock In \emph{CVPR}, 2022.

\bibitem[Chandrasekaran et~al.(2009)Chandrasekaran, Sanghavi, Parrilo, and Willsky]{sparselowrank}
Venkat Chandrasekaran, Sujay Sanghavi, Pablo~A Parrilo, and Alan~S Willsky.
\newblock Sparse and low-rank matrix decompositions.
\newblock \emph{IFAC Proceedings Volumes}, 42\penalty0 (10):\penalty0 1493--1498, 2009.

\bibitem[Chen et~al.(2022)Chen, Xu, Geiger, Yu, and Su]{tensorf}
Anpei Chen, Zexiang Xu, Andreas Geiger, Jingyi Yu, and Hao Su.
\newblock Tensorf: Tensorial radiance fields.
\newblock In \emph{European Conference on Computer Vision (ECCV)}, 2022.

\bibitem[Croce et~al.(2023)Croce, Caroti, De~Luca, Piemonte, and V{\'e}ron]{culturalheritage}
V~Croce, G~Caroti, L~De~Luca, A~Piemonte, and P~V{\'e}ron.
\newblock Neural radiance fields (nerf): Review and potential applications to digital cultural heritage.
\newblock \emph{The International Archives of the Photogrammetry, Remote Sensing and Spatial Information Sciences}, 48:\penalty0 453--460, 2023.

\bibitem[Dorst et~al.(2009)Dorst, Fontijne, and Mann]{geometricalgebrabook}
Leo Dorst, Daniel Fontijne, and Stephen Mann.
\newblock \emph{Geometric algebra for computer science (revised edition): An object-oriented approach to geometry}.
\newblock Morgan Kaufmann, 2009.

\bibitem[Ergen \& Pilanci(2024)Ergen and Pilanci]{deepconvex}
Tolga Ergen and Mert Pilanci.
\newblock Path regularization: A convexity and sparsity inducing regularization for parallel relu networks.
\newblock \emph{Advances in Neural Information Processing Systems}, 36, 2024.

\bibitem[Fridovich-Keil et~al.(2023)Fridovich-Keil, Meanti, Warburg, Recht, and Kanazawa]{kplanes}
Sara Fridovich-Keil, Giacomo Meanti, Frederik~Rahbæk Warburg, Benjamin Recht, and Angjoo Kanazawa.
\newblock K-planes: Explicit radiance fields in space, time, and appearance.
\newblock In \emph{CVPR}, 2023.

\bibitem[Intwala \& Magikar(2016)Intwala and Magikar]{manufacturingreview}
Aditya~M Intwala and Atul Magikar.
\newblock A review on process of 3d model reconstruction.
\newblock In \emph{2016 International Conference on Electrical, Electronics, and Optimization Techniques (ICEEOT)}, pp.\  2851--2855. IEEE, 2016.

\bibitem[Jocher et~al.(2023)Jocher, Qiu, and Chaurasia]{yolo}
Glenn Jocher, Jing Qiu, and Ayush Chaurasia.
\newblock {Ultralytics YOLO}, January 2023.
\newblock URL \url{https://github.com/ultralytics/ultralytics}.

\bibitem[Kajiya(1986)]{kajiya}
James~T Kajiya.
\newblock The rendering equation.
\newblock In \emph{Proceedings of the 13th annual conference on Computer graphics and interactive techniques}, pp.\  143--150, 1986.

\bibitem[Kerbl et~al.(2023)Kerbl, Kopanas, Leimk{\"u}hler, and Drettakis]{3Dgaussians}
Bernhard Kerbl, Georgios Kopanas, Thomas Leimk{\"u}hler, and George Drettakis.
\newblock 3d gaussian splatting for real-time radiance field rendering.
\newblock \emph{ACM Transactions on Graphics}, 42\penalty0 (4), July 2023.
\newblock URL \url{https://repo-sam.inria.fr/fungraph/3d-gaussian-splatting/}.

\bibitem[Kirillov et~al.(2023)Kirillov, Mintun, Ravi, Mao, Rolland, Gustafson, Xiao, Whitehead, Berg, Lo, Dollár, and Girshick]{sam1}
Alexander Kirillov, Eric Mintun, Nikhila Ravi, Hanzi Mao, Chloe Rolland, Laura Gustafson, Tete Xiao, Spencer Whitehead, Alexander~C. Berg, Wan-Yen Lo, Piotr Dollár, and Ross Girshick.
\newblock Segment anything, 2023.
\newblock URL \url{https://arxiv.org/abs/2304.02643}.

\bibitem[Kutulakos \& Seitz(1999)Kutulakos and Seitz]{spacecarving}
K.N. Kutulakos and S.M. Seitz.
\newblock A theory of shape by space carving.
\newblock In \emph{Proceedings of the Seventh IEEE International Conference on Computer Vision}, volume~1, pp.\  307--314 vol.1, 1999.
\newblock \doi{10.1109/ICCV.1999.791235}.

\bibitem[Labelbox.com()]{videoprocessing}
Labelbox.com.
\newblock Using meta's segment anything (sam) model on video with labelbox's model-assisted labeling.
\newblock \url{https://labelbox.com/guides/using-metas-segment-anything-sam-model-on-video-with-labelbox-model-assisted-labeling/}.
\newblock Accessed: 2024-10-01.

\bibitem[Liu et~al.(2024)Liu, Huang, Huang, Chen, Hou, Tang, Liu, Ouyang, Zuo, Jiang, et~al.]{contentgeneration}
Jian Liu, Xiaoshui Huang, Tianyu Huang, Lu~Chen, Yuenan Hou, Shixiang Tang, Ziwei Liu, Wanli Ouyang, Wangmeng Zuo, Junjun Jiang, et~al.
\newblock A comprehensive survey on 3d content generation.
\newblock \emph{arXiv preprint arXiv:2402.01166}, 2024.

\bibitem[Lombardi et~al.(2021)Lombardi, Simon, Schwartz, Zollhoefer, Sheikh, and Saragih]{mvp}
Stephen Lombardi, Tomas Simon, Gabriel Schwartz, Michael Zollhoefer, Yaser Sheikh, and Jason Saragih.
\newblock Mixture of volumetric primitives for efficient neural rendering.
\newblock \emph{ACM Trans. Graph.}, 40\penalty0 (4), jul 2021.
\newblock ISSN 0730-0301.
\newblock \doi{10.1145/3450626.3459863}.
\newblock URL \url{https://doi.org/10.1145/3450626.3459863}.

\bibitem[Masero et~al.(2002)Masero, LE{\'O}N-ROJAS, and Moreno]{volumesforhealthcare}
Valent{\'\i}n Masero, JUAN~M LE{\'O}N-ROJAS, and Jos{\'e} Moreno.
\newblock Volume reconstruction for health care: a survey of computational methods.
\newblock \emph{Annals of the New York Academy of Sciences}, 980\penalty0 (1):\penalty0 198--211, 2002.

\bibitem[Max(1995)]{max}
Nelson Max.
\newblock Optical models for direct volume rendering.
\newblock \emph{IEEE Transactions on Visualization and Computer Graphics}, 1\penalty0 (2):\penalty0 99--108, 1995.

\bibitem[Mescheder et~al.(2019)Mescheder, Oechsle, Niemeyer, Nowozin, and Geiger]{occupancynet}
Lars Mescheder, Michael Oechsle, Michael Niemeyer, Sebastian Nowozin, and Andreas Geiger.
\newblock Occupancy networks: Learning 3d reconstruction in function space.
\newblock In \emph{Proceedings of the IEEE/CVF conference on computer vision and pattern recognition}, pp.\  4460--4470, 2019.

\bibitem[Mildenhall et~al.(2020)Mildenhall, Srinivasan, Tancik, Barron, Ramamoorthi, and Ng]{nerf}
Ben Mildenhall, Pratul~P. Srinivasan, Matthew Tancik, Jonathan~T. Barron, Ravi Ramamoorthi, and Ren Ng.
\newblock Nerf: Representing scenes as neural radiance fields for view synthesis.
\newblock In \emph{ECCV}, 2020.

\bibitem[Ming et~al.(2024)Ming, Yang, Wang, Chen, Feng, Xing, and Zhang]{nerfforrobots}
Yuhang Ming, Xingrui Yang, Weihan Wang, Zheng Chen, Jinglun Feng, Yifan Xing, and Guofeng Zhang.
\newblock Benchmarking neural radiance fields for autonomous robots: An overview.
\newblock \emph{arXiv preprint arXiv:2405.05526}, 2024.

\bibitem[M\"uller et~al.(2022)M\"uller, Evans, Schied, and Keller]{ingp}
Thomas M\"uller, Alex Evans, Christoph Schied, and Alexander Keller.
\newblock Instant neural graphics primitives with a multiresolution hash encoding.
\newblock \emph{ACM Trans. Graph.}, 41\penalty0 (4):\penalty0 102:1--102:15, July 2022.
\newblock \doi{10.1145/3528223.3530127}.
\newblock URL \url{https://doi.org/10.1145/3528223.3530127}.

\bibitem[Oppenheim(1999)]{oppenheim1999discrete}
Alan~V Oppenheim.
\newblock \emph{Discrete-time signal processing}.
\newblock Pearson Education India, 1999.

\bibitem[Pilanci \& Ergen(2020)Pilanci and Ergen]{convexnn2layer}
Mert Pilanci and Tolga Ergen.
\newblock Neural networks are convex regularizers: Exact polynomial-time convex optimization formulations for two-layer networks.
\newblock In Hal~Daumé III and Aarti Singh (eds.), \emph{Proceedings of the 37th International Conference on Machine Learning}, volume 119 of \emph{Proceedings of Machine Learning Research}, pp.\  7695--7705. PMLR, 13--18 Jul 2020.
\newblock URL \url{https://proceedings.mlr.press/v119/pilanci20a.html}.

\bibitem[Reiser et~al.(2023)Reiser, Szeliski, Verbin, Srinivasan, Mildenhall, Geiger, Barron, and Hedman]{merf}
Christian Reiser, Richard Szeliski, Dor Verbin, Pratul~P. Srinivasan, Ben Mildenhall, Andreas Geiger, Jonathan~T. Barron, and Peter Hedman.
\newblock Merf: Memory-efficient radiance fields for real-time view synthesis in unbounded scenes.
\newblock \emph{SIGGRAPH}, 2023.

\bibitem[Richter et~al.(2024)Richter, Steinmann, Rosenthal, and Rupitsch]{3dforendoscopy}
Alexander Richter, Till Steinmann, Jean-Claude Rosenthal, and Stefan~J Rupitsch.
\newblock Advances in real-time 3d reconstruction for medical endoscopy.
\newblock \emph{Journal of Imaging}, 10\penalty0 (5):\penalty0 120, 2024.

\bibitem[Sahiner et~al.(2024)Sahiner, Ergen, Ozturkler, Pauly, Mardani, and Pilanci]{mertsemiconvex}
Arda Sahiner, Tolga Ergen, Batu Ozturkler, John~M Pauly, Morteza Mardani, and Mert Pilanci.
\newblock Scaling convex neural networks with burer-monteiro factorization.
\newblock In \emph{ICLR}, 2024.

\bibitem[{Sara Fridovich-Keil and Alex Yu} et~al.(2022){Sara Fridovich-Keil and Alex Yu}, Tancik, Chen, Recht, and Kanazawa]{yu_and_fridovichkeil2021plenoxels}
{Sara Fridovich-Keil and Alex Yu}, Matthew Tancik, Qinhong Chen, Benjamin Recht, and Angjoo Kanazawa.
\newblock Plenoxels: Radiance fields without neural networks.
\newblock In \emph{CVPR}, 2022.

\bibitem[Saragadam et~al.(2023)Saragadam, LeJeune, Tan, Balakrishnan, Veeraraghavan, and Baraniuk]{wire}
Vishwanath Saragadam, Daniel LeJeune, Jasper Tan, Guha Balakrishnan, Ashok Veeraraghavan, and Richard~G Baraniuk.
\newblock Wire: Wavelet implicit neural representations.
\newblock In \emph{Conf. Computer Vision and Pattern Recognition}, 2023.

\bibitem[Sch\"{o}nberger \& Frahm(2016)Sch\"{o}nberger and Frahm]{schoenberger2016sfm}
Johannes~Lutz Sch\"{o}nberger and Jan-Michael Frahm.
\newblock Structure-from-motion revisited.
\newblock In \emph{Conference on Computer Vision and Pattern Recognition (CVPR)}, 2016.

\bibitem[Sch\"{o}nberger et~al.(2016)Sch\"{o}nberger, Zheng, Pollefeys, and Frahm]{schoenberger2016mvs}
Johannes~Lutz Sch\"{o}nberger, Enliang Zheng, Marc Pollefeys, and Jan-Michael Frahm.
\newblock Pixelwise view selection for unstructured multi-view stereo.
\newblock In \emph{European Conference on Computer Vision (ECCV)}, 2016.

\bibitem[Sitzmann et~al.(2019{\natexlab{a}})Sitzmann, Thies, Heide, Nie{\ss}ner, Wetzstein, and Zollh{\"o}fer]{deepvoxels}
Vincent Sitzmann, Justus Thies, Felix Heide, Matthias Nie{\ss}ner, Gordon Wetzstein, and Michael Zollh{\"o}fer.
\newblock Deepvoxels: Learning persistent 3d feature embeddings.
\newblock In \emph{Proc. Computer Vision and Pattern Recognition (CVPR), IEEE}, 2019{\natexlab{a}}.

\bibitem[Sitzmann et~al.(2019{\natexlab{b}})Sitzmann, Zollh{\"o}fer, and Wetzstein]{srn}
Vincent Sitzmann, Michael Zollh{\"o}fer, and Gordon Wetzstein.
\newblock Scene representation networks: Continuous 3d-structure-aware neural scene representations.
\newblock In \emph{Advances in Neural Information Processing Systems}, 2019{\natexlab{b}}.

\bibitem[Sitzmann et~al.(2020)Sitzmann, Martel, Bergman, Lindell, and Wetzstein]{siren}
Vincent Sitzmann, Julien~N.P. Martel, Alexander~W. Bergman, David~B. Lindell, and Gordon Wetzstein.
\newblock Implicit neural representations with periodic activation functions.
\newblock In \emph{Proc. NeurIPS}, 2020.

\bibitem[{\v{S}}lapak et~al.(2024){\v{S}}lapak, Pardo, Dopiriak, Maksymyuk, and Gazda]{navigationandmanufacturing}
Eugen {\v{S}}lapak, Enric Pardo, Mat{\'u}{\v{s}} Dopiriak, Taras Maksymyuk, and Juraj Gazda.
\newblock Neural radiance fields in the industrial and robotics domain: applications, research opportunities and use cases.
\newblock \emph{Robotics and Computer-Integrated Manufacturing}, 90:\penalty0 102810, 2024.

\bibitem[Sun et~al.(2022)Sun, Sun, and Chen]{dvgo}
Cheng Sun, Min Sun, and Hwann{-}Tzong Chen.
\newblock Direct voxel grid optimization: Super-fast convergence for radiance fields reconstruction.
\newblock In \emph{CVPR}, 2022.

\bibitem[Tancik et~al.(2020)Tancik, Srinivasan, Mildenhall, Fridovich-Keil, Raghavan, Singhal, Ramamoorthi, Barron, and Ng]{fourfeat}
Matthew Tancik, Pratul~P. Srinivasan, Ben Mildenhall, Sara Fridovich-Keil, Nithin Raghavan, Utkarsh Singhal, Ravi Ramamoorthi, Jonathan~T. Barron, and Ren Ng.
\newblock Fourier features let networks learn high frequency functions in low dimensional domains.
\newblock \emph{NeurIPS}, 2020.

\bibitem[Tancik et~al.(2023)Tancik, Weber, Ng, Li, Yi, Kerr, Wang, Kristoffersen, Austin, Salahi, Ahuja, McAllister, and Kanazawa]{nerfstudio}
Matthew Tancik, Ethan Weber, Evonne Ng, Ruilong Li, Brent Yi, Justin Kerr, Terrance Wang, Alexander Kristoffersen, Jake Austin, Kamyar Salahi, Abhik Ahuja, David McAllister, and Angjoo Kanazawa.
\newblock Nerfstudio: A modular framework for neural radiance field development.
\newblock In \emph{ACM SIGGRAPH 2023 Conference Proceedings}, SIGGRAPH '23, 2023.

\bibitem[Tewari et~al.(2022)Tewari, Thies, Mildenhall, Srinivasan, Tretschk, Yifan, Lassner, Sitzmann, Martin-Brualla, Lombardi, et~al.]{tewarisurvey}
Ayush Tewari, Justus Thies, Ben Mildenhall, Pratul Srinivasan, Edgar Tretschk, Wang Yifan, Christoph Lassner, Vincent Sitzmann, Ricardo Martin-Brualla, Stephen Lombardi, et~al.
\newblock Advances in neural rendering.
\newblock In \emph{Computer Graphics Forum}, volume~41, pp.\  703--735. Wiley Online Library, 2022.

\bibitem[Udupa \& Herman(1999)Udupa and Herman]{3dmedicalimagingbook}
Jayaram~K Udupa and Gabor~T Herman.
\newblock \emph{3D imaging in medicine}.
\newblock CRC press, 1999.

\bibitem[Uy et~al.(2023)Uy, Martin-Brualla, Guibas, and Li]{scade}
Mikaela~Angelina Uy, Ricardo Martin-Brualla, Leonidas Guibas, and Ke~Li.
\newblock Scade: Nerfs from space carving with ambiguity-aware depth estimates.
\newblock In \emph{Conference on Computer Vision and Pattern Recognition (CVPR)}, 2023.

\bibitem[Wang et~al.(2004)Wang, Bovik, Sheikh, and Simoncelli]{ssim}
Zhou Wang, Alan~C Bovik, Hamid~R Sheikh, and Eero~P Simoncelli.
\newblock Image quality assessment: from error visibility to structural similarity.
\newblock \emph{IEEE transactions on image processing}, 13\penalty0 (4):\penalty0 600--612, 2004.

\bibitem[Wijayathunga et~al.(2023)Wijayathunga, Rassau, and Chai]{navigationreview1}
Liyana Wijayathunga, Alexander Rassau, and Douglas Chai.
\newblock Challenges and solutions for autonomous ground robot scene understanding and navigation in unstructured outdoor environments: A review.
\newblock \emph{Applied Sciences}, 13\penalty0 (17):\penalty0 9877, 2023.

\bibitem[Xu et~al.(2024)Xu, Guo, Wang, Bai, and Ren]{nerfforsurgery}
Mengya Xu, Ziqi Guo, An~Wang, Long Bai, and Hongliang Ren.
\newblock A review of 3d reconstruction techniques for deformable tissues in robotic surgery.
\newblock \emph{arXiv preprint arXiv:2408.04426}, 2024.

\bibitem[Zhang et~al.(2018)Zhang, Isola, Efros, Shechtman, and Wang]{lpips}
Richard Zhang, Phillip Isola, Alexei~A Efros, Eli Shechtman, and Oliver Wang.
\newblock The unreasonable effectiveness of deep features as a perceptual metric.
\newblock In \emph{CVPR}, 2018.

\end{thebibliography}
\bibliographystyle{iclr2025_conference}

\clearpage
\appendix
\section{Appendix}
\label{sec:appendix}

\subsection{Context for \modelname{}}
\label{sec:appendixcontext}

\begin{wraptable}{r}{0.5\linewidth}
    \centering
    \resizebox{\linewidth}{!}{
    \begin{tabular}{l|ccc}
         &  \rotatebox[origin=l]{90}{Model Size} & \rotatebox[origin=l]{90}{Expressiveness} & \rotatebox[origin=l]{90}{Optimizability}  \\ \midrule
         Coordinate MLP {\tiny (NeRF, SRN)} & \good & \fair & \bad \\
         Voxels {\tiny (Space Carving, Plenoxels, DVGO)} & \bad & \good & \good \\
         Tensor Factorization {\tiny (TensoRF, K-Planes)} & \fair & \fair & \fair \\
         Hash Embedding {\tiny (Instant-NGP)} & \fair & \good & \fair \\
         Point Cloud / Splat {\tiny (3D Gaussian Splatting)} & \fair & \good & \fair \\
         Mixture of Primitives {\tiny (MVP, MERF)} & \fair & \good & \fair \\
         \midrule
         \modelname{} (Nonconvex) & \fair & \good & \fair \\
         \modelname{} (Convex) & \fair & \good & \good \\
         \modelname{} (Semi-Convex) & \fair & \good & \good \\ \bottomrule
    \end{tabular}}
    \vspace{-0.1cm}
    \caption{\textbf{Context.} All volume models face a tradeoff between memory efficiency, expressiveness, and optimizability. The qualitative categorizations here are based on the tradeoffs achieved by representative example methods listed in each category. \emph{Model Size} denotes memory usage during training; other methods exist to compress trained models, e.g. for rendering on mobile hardware. \emph{Optimizability} denotes both speed and stability of optimization/training. For example, Coordinate MLPs tend to train slowly, while Splats train quickly but are sensitive to initialization.}
    \label{tab:relatedwork}
\end{wraptable}

\Cref{tab:relatedwork} summarizes some representative volume models popular in computer vision, and how they relate to \modelname{} along the three-way pareto frontier of model size, expressiveness, and optimizability.

\emph{Implicit Neural Representations} (INRs) or Coordinate Neural Networks \citep{nerf, srn, fourfeat, siren, wire} parameterize the volume implicitly through the weights of a neural network, typically a multilayer perceptron (MLP) with some modification to overcome spectral bias and represent high frequency content. These models tend to provide decent expressiveness with very small model size; their main drawback is slow optimization.

\emph{Voxel grids} \citep{spacecarving, yu_and_fridovichkeil2021plenoxels, dvgo, deepvoxels} are perhaps the most traditional parameterization of a volume, where each parameter denotes the function value (density, color, a latent feature, etc.) at a specific grid cell location within the volume. These voxel values can then be combined into a continuous function over the 3D space by some form of interpolation, following the standard Nyquist sampling and reconstruction paradigm of digital signal processing \citep{oppenheim1999discrete}. Voxels offer direct control over expressivity (via resolution) and are easily optimized; their main drawback is memory usage because the number of parameters grows cubically with the spatial resolution. 

\emph{Tensor factorizations} \citep{tensorf, eg3d, kplanes} parameterize a 3D volume as a combination of lower-dimension objects, namely vectors and matrices (lines and planes). Tensor factorizations tend to balance the three attributes somewhat evenly, offering decent expressiveness and optimizability while using more memory than an INR but less than a high resolution voxel grid. 

\emph{Hash embeddings} \citep{ingp, nerfstudio} are similar to voxels, but replace the explicit voxel grid in 3D with a multiresolution 3D hash function followed by a small MLP decoder to disambiguate hash collisions. They can optimize very quickly and with better memory efficiency compared to voxels; quality is mixed with good high-resolution details but also some high-frequency noise likely arising from unresolved hash collisions or sensitivity to random initialization. 

\emph{Point clouds / splats} \citep{3Dgaussians, schoenberger2016sfm, schoenberger2016mvs} represent a volume as a collection of 3D points or blobs, where the points need not be arranged on a regular grid. They are highly expressive and less memory-intensive than voxels (but still more so than some other methods). They can optimize very quickly but often require heuristic or discrete optimization strategies that result in sensitivity to initialization. 

\emph{Mixture of primitives} \citep{merf, mvp} models combine multiple of the above representation strategies to balance their strengths and weaknesses. For example, combining low resolution voxels with a high resolution tensor factorization is an effective strategy to improve on the expressiveness of tensor factorizations without resorting to the cubic memory requirement of a high resolution voxel grid; this strategy underlies both MERF \citep{merf} and \modelname{}.

We emphasize that all of these existing methods (except perhaps voxels) require nonconvex optimization, often for a feature decoder MLP, and thus risk getting stuck in suboptimal local minima depending on the randomness of initialization and the trajectory of stochastic gradients. In practice, as described above, some of the prior methods exhibit greater optimization stability than others, though none (except voxels in limited settings) come with guarantees of convergence to global optimality. In contrast, both the convex and semiconvex \modelname{} formulations come with guarantees that all local optima are also global \citep{mertsemiconvex}.

\paragraph{Relation to Prior Models.}
Without any convexity restrictions, the \modelname{} family includes many previously proposed models as special cases:
\begin{itemize}
    \item NeRF \citep{nerf}: $\decoder{}$
    \item Plenoxels \citep{yu_and_fridovichkeil2021plenoxels}, DVGO \citep{dvgo}: $\decoder[\eonetwothree]$
    \item TensoRF \citep{tensorf}: $\decoder[(\eone\circ\etwothree) \odot (\etwo\circ\eonethree) \odot (\ethree\circ\eonetwo)]$
    \item Tri-Planes \citep{eg3d}: $\decoder[\eonetwo + \eonethree + \etwothree]$
    \item K-Planes \citep{kplanes}: $\decoder[\eonetwo \circ \eonethree \circ \etwothree]$
    \item MERF \citep{merf}: $\decoder[\eonetwo + \eonethree + \etwothree + \eonetwothree]$
\end{itemize}

Of these, all except for TensoRF and K-Planes are compatible with convex optimization towards any convex objective. Note that different models may use different decoder architectures for $\decoder{}$, including both linear and MLP decoders and additional decoder inputs such as encoded viewing direction and/or positionally-encoded coordinates.

\subsection{Proof of Theorems}
\label{sec:appendix-general-proof}
\paragraph{A general note on proofs.} 
In order to represent a matrix $M \in \mathbb{R}^{m\times n}$ with an implicit model, we compute $\decoder{(f(q))}$ for $q = (k,l) ,\,\ \forall k \in \{1,\dots, m\} ,\,\ \forall l \in \{1,\dots, n\}$. Considering line feature grids with resolutions matching $m$, $n$; the features will become $\eone = {(\gone)}_k$, $\etwo = {(\gtwo)}_l$ for $q = (k,l)$ otherwise they will be $\eone = {\varphi(\gone)}_k$, $\etwo = {\varphi(\gtwo)}_l$ where $\varphi(\gone)$, $\varphi(\gtwo)$ now have resolutions $m$, $n$ after interpolation through $\varphi$. 
Here $\varphi$ can be any interpolation scheme with linear weighting of inputs, e.g. nearest neighbor, (bi)linear, (bi)cubic, spline, Gaussian, sinc, etc.
For the simplicity of notation, we omit $\varphi$ in line feature grids, and only apply it to the plane feature grid $\gonetwo$, which has lower resolution by design. The proofs consider $\gone, \gtwo \in \mathbb{R}^{r_1 \times d_1}$ and $\gonetwo \in \mathbb{R}^{r_2 \times r_2 \times d_1}$ (equal feature dimensions, different resolutions), resulting in the matrix representation $\hat{M} \in \mathbb{R}^{r_1 \times r_1}$ (the case where $m=n=r_1$). Note that if the interpolation is done by a method other than nearest neighbor, this may allow a (convex or nonconvex) MLP decoder to increase the rank beyond $r_1$. We derive expressions for $\hat{M}$ implied by different \modelname{} variations in the parts that follow. The coordinate-wise optimization objective (in the case of a directly supervised mean-square-error loss) corresponds to minimizing the Frobenius norm of the ground truth matrix $M$ and its approximation $\hat{M}$.

\subsubsection{Proof of \cref{thm:triplane}}
\label{sec:prooftriplane}

The forward mapping of the model $\decoder{(\eone + \etwo)}$ is: 
\begin{align}
    \tilde{y}(q) = \decoder{(\eone + \etwo)} = \alpha^\top (\eone + \etwo) = \alpha^\top ({(\gone)}_k + {(\gtwo)}_l) = \sum_{i=1}^{d_1} {\alpha_i ((\gone)_{k,i} +(\gtwo)_{l,i})} ,
\end{align}
where ${(\gone)}_k, {(\gone)}_l \in \mathbb{R}^{d_1 \times 1}$. 
 
In matrix form,
\begin{align}
\label{eq:proof-triplane-add}
    \hat{M} = \sum_{i=1}^{d_1} {\alpha_i ((\gone)_{i} \mathds{1}^\top +\mathds{1}   (\gtwo)_{i}^\top)}
    = \sum_{i=1}^{d_1} {\alpha_i (\gone)_{i} \mathds{1}^\top} +\sum_{i=1}^{d_1} {\alpha_i \mathds{1}   (\gtwo)_{i}^\top} .
\end{align}
Defining $U := \gone \mathrm{diag}(\alpha) , U \in \mathbb{R}^{r_1\times d_1}$ and $V := \gtwo \mathrm{diag}(\alpha)   , V \in \mathbb{R}^{r_1\times d_1}$, this can be expressed as 
\begin{equation}
    \hat{M} = U \mathds{1}_{r_1 \times d_1}^\top + {\mathds{1}_{r_1 \times d_1}} V^\top .
\end{equation}
Note that the resulting matrix $\hat{M} \in \mathbb{R}^{r_1 \times r_1}$ has rank at most $2$---very limited expressivity---regardless of the resolution $r_1$. This is because the all-ones matrix is rank 1, and a product of matrices cannot have higher rank than either of its factors.

Similarly for the multiplicative representation $\decoder{(\eone \circ \etwo)}$, the mapping is
\begin{align}
    \tilde{y}(q) = \decoder{(\eone \circ \etwo)} = \alpha^\top (\eone \circ \etwo) = \alpha^\top ({(\gone)}_k \circ {(\gtwo)}_l) = \sum_{i=1}^{d_1} {\alpha_i (\gone)_{k,i} (\gtwo)_{l,i}} ,
\end{align}
where ${(\gone)}_k, {(\gone)}_l \in \mathbb{R}^{d_1\times1}$. In matrix form,
\begin{align}
\label{eq:proof-triplane-mult}
    \hat{M} = \sum_{i=1}^{d_1} {\alpha_i (\gone)_{i} (\gtwo)_{i}^\top}
    = \gone \mathrm{diag}(\alpha) \gtwo^\top .
\end{align}
Defining $U := \gone \mathrm{diag}(\alpha) , U \in \mathbb{R}^{r_1\times d_1}$ and $V := \gtwo, V \in \mathbb{R}^{r_1\times d_1}$, this can be expressed as 
\begin{equation}
    \hat{M} = U V^\top ,
\end{equation}
which is the optimal rank-$d_1$ decomposition.

\subsubsection{Proof of \cref{thm:gaplane}}
\label{sec:proofgaplane}

The forward mapping of the model $\decoder{(\eone + \etwo + \eonetwo)}$ becomes: 
\begin{gather}
    \tilde{y}(q) = \decoder{(\eone + \etwo + \eonetwo)} = \alpha^\top (\eone + \etwo + \eonetwo) = \alpha^\top ({(\gone)}_k + {(\gtwo)}_l + {\varphi(\gonetwo)}_{k,l})\\ = \sum_{i=1}^{d_1} {\alpha_i ((\gone)_{k,i} +(\gtwo)_{l,i})} + \sum_{i=1}^{d_1} {\alpha_i {\varphi(\gonetwo)}_{k,l,i}} ,
\end{gather}
where ${(\gone)}_k, {(\gone)}_l, {\varphi(\gonetwo)}_{k,l} \in \mathbb{R}^{d_1 \times 1}$. 
In matrix form,
\begin{align}
    \hat{M} = \sum_{i=1}^{d_1} {\alpha_i ((\gone)_{i} \mathds{1}^\top +\mathds{1}   (\gtwo)_{i}^\top)} + \sum_{i=1}^{d_1} {\alpha_i{\varphi(\gonetwo)}_{i}} .
\end{align}
Noting that the first term is the same as in \cref{eq:proof-triplane-add} and defining $L := \gonetwo \alpha, L \in \mathbb{R}^{r_2 \times r_2}$, we reach the expression 
\begin{equation}
    \hat{M} = U \mathds{1}_{d_1 \times r_1} + {\mathds{1}_{r_1 \times d_1}} V^\top + \varphi(L) ,
\end{equation}
since $\sum_{i=1}^{d_1} {\alpha_i{\varphi(\gonetwo)}_{i}} = \varphi(\sum_{i=1}^{d_1} \alpha_i{(\gonetwo)}_{i}) = \varphi(\gonetwo \alpha)$, following the linearity of the upsampling function $\varphi$.
Note that in the definition of $L$ there is a tensor-vector product that effectively takes a dot product along the last (feature) dimension.

Similarly for the multiplicative representation $\decoder{(\eone \circ \etwo + \eonetwo)}$, the mapping is
\begin{gather}
    \tilde{y}(q) = \decoder{(\eone \circ \etwo + \eonetwo)} = \alpha^\top (\eone \circ \etwo + \eonetwo)\\ = \alpha^\top ({(\gone)}_k \circ {(\gtwo)}_l + {\varphi(\gonetwo)}_{k,l}) = \sum_{i=1}^{d_1} {\alpha_i (\gone)_{k,i} (\gtwo)_{l,i}} + \sum_{i=1}^{d_1} {\alpha_i {\varphi(\gonetwo)}_{k,l,i}} .
\end{gather}

In matrix notation, we have
\begin{align}
    \hat{M} = \sum_{i=1}^{d_1} {\alpha_i ((\gone)_{i} (\gtwo)_{i}^\top)} + \sum_{i=1}^{d_1} {\alpha_i{\varphi(\gonetwo)}_{i}} .
\end{align}
Following \cref{eq:proof-triplane-mult} and using the same definition of $L$, the final expression becomes
\begin{align}
    \hat{M} = U V^\top + \varphi(L) .
\end{align}

\subsubsection{Proof of \cref{thm:triplane-convex-mlp}}
\label{sec:prooftriplaneconvexmlp}

For a 2-layer convex MLP with hidden size $h$, denote the trainable first layer weights as $W \in \mathbb{R}^{h \times d_1}$ and the gating weights as $\overline{W} \in \mathbb{R}^{h \times d_1}$ (which are fixed at random initialization). 
We will handle three different cases for merging the interpolated features: multiplication ($\circ$), addition ($+$), and concatenation ($\odot$).

The forward mapping of the multiplicative model using a convex MLP, $\decoder{(\eone \circ \etwo)}$ at $q = (k,l)$ is
\begin{align}
\tilde{y}(q) &= \mathds{1}_h^\top \left( \left(W ({(\gone)}_k \circ {(\gtwo)}_l) \right) \circ \mathds{1} \left[\overline{W} ({(\gone)}_k \circ {(\gtwo)}_l) \geq 0 \right] \right) \\
     &= \sum_{i=1}^{h} { \left( \sum_{j=1}^{d_1} {W_{i,j} {(\gone)}_{k,j} {(\gtwo)}_{l,j}} \right)  \mathds{1} \left[ \sum_{j=1}^{d_1} {\overline{W}_{i,j} {(\gone)}_{k,j} {(\gtwo)}_{l,j}} \geq 0 \right]} ,
\end{align}
where $\circ$ denotes elementwise multiplication (Hadamard product).
The resulting matrix decomposition can then be written as
\begin{align}
    \hat{M} &= \sum_{i=1}^{h} { \left( \sum_{j=1}^{d_1} {W_{i,j} {(\gone)}_{j} {(\gtwo)}_{j}^\top } \right)} \circ
      \mathds{1} \left[ \sum_{j=1}^{d_1} {\overline{W}_{i,j} {(\gone)}_{j} {(\gtwo)}_{j}^\top }  \geq 0 \right] .
\end{align}

Now, we define the masking matrix $ B_i = \mathds{1} \left[ \sum_{j=1}^{d_1} {\overline{W}_{i,j} {(\gone)}_{j} {(\gtwo)}_{j}^\top }  \geq 0 \right]$ and the eigenvectors $U_j = {(\gone)}_{j}$, $V_j = {(\gtwo)}_{j}$ to reach the expression from the theorem statement:
\begin{align}
    \hat{M} =  \sum_{i,j} {W_{i,j} {U}_{j} {V}_{j}^\top } \circ B_i .
\end{align}

When the model uses additive features as in $\decoder{(\eone + \etwo)}$, and $\decoder{}$ is a convex MLP, the prediction is
\begin{align}
    \tilde{y}(q) &= \mathds{1}_h^\top \left( \left(W ({(\gone)}_k + {(\gtwo)}_l) \right) \circ \mathds{1} \left[\overline{W} ({(\gone)}_k + {(\gtwo)}_l) \geq 0 \right] \right) \\
    &= \sum_{i=1}^{h} { \left( \sum_{j=1}^{d_1} {W_{i,j} ({(\gone)}_{k,j} +{(\gtwo)}_{l,j})} \right)  \mathds{1} \left[ \sum_{j=1}^{d_1} {\overline{W}_{i,j} ({(\gone)}_{k,j} + {(\gtwo)}_{l,j})} \geq 0 \right]} .
\end{align}
The resulting matrix decomposition can then be written as
\begin{align}
    \hat{M} &= \sum_{i=1}^{h} { \left( \sum_{j=1}^{d_1} {W_{i,j} ({(\gone)}_{j}  \mathds{1}_{r_1}^\top + \mathds{1}_{r_1}  {(\gtwo)}_{j}^\top )} \right)} 
      \mathds{1} \left[ \sum_{j=1}^{d_1} {\overline{W}_{i,j} ({(\gone)}_{j} \mathds{1}_{r_1}^\top + \mathds{1}_{r_1}  {(\gtwo)}_{j}^\top )}  \geq 0 \right] .
\end{align}

Defining $ B_i = \mathds{1} \left[ \sum_{j=1}^{d_1} {\overline{W}_{i,j} ({(\gone)}_{j}  \mathds{1}_{r_1}^\top + \mathds{1}_{r_1}  {(\gtwo)}_{j}^\top )}  \geq 0 \right]$, $U_j = {(\gone)}_{j}$, $V_j = {(\gtwo)}_{j}$, we reach the final expression:

\begin{align}
\label{eq:add-conv-mlp}
\hat{M} = \sum_{i,j} {{W_{i,j} \big({U}_{j}  \mathds{1}_{r_1}^\top + \mathds{1}_{r_1}  {V}_{j}^\top \big)}  \circ B_i} .
\end{align}

Finally, we show that concatenation of features results in a very similar expression to \cref{eq:add-conv-mlp}.

When the model uses concatenated features as in $\decoder{(\eone \odot \etwo)}$, and $\decoder{}$ is a convex MLP (with trainable weights $W \in \mathbb{R}^{h \times 2 d_1}$ and fixed gates $\overline{W} \in \mathbb{R}^{h \times 2 d_1}$), the prediction at a point $q$ is
\begin{align}
    \tilde{y}(q) &= \mathds{1}_h^\top \left( \left(W ({(\gone)}_k \odot {(\gtwo)}_l) \right) \circ \mathds{1} \left[\overline{W} ({(\gone)}_k \odot {(\gtwo)}_l) \geq 0 \right] \right) .
\end{align}
Denoting the weights and gates each as a concatenation of 2 matrices, $W = (W_1 \odot W_2)$, $\overline{W} = (\overline{W}_1 \odot \overline{W}_2)$, where $W_1, W_2, \overline{W}_1, \overline{W}_2 \in \mathbb{R}^{h \times d_1}$, we have the following expression:
\begin{align}
    \tilde{y}(q) &= \sum_{i=1}^{h} { \left( \sum_{j=1}^{d_1} { {W_1}_{i,j} {(\gone)}_{k,j} +{{W_2}_{i,j}(\gtwo)}_{l,j}} \right)  \mathds{1} \left[ \sum_{j=1}^{d_1} {  {\overline{W}_1}_{i,j}{(\gone)}_{k,j} +  {\overline{W}_2}_{i,j} {(\gtwo)}_{l,j}} \geq 0 \right]} .
\end{align}

Following similar steps as for the additive case, we express the matrix decomposition as
\begin{align}
    \hat{M} &= \sum_{i,j} {   ({{W_1}_{i,j} U_j \mathds{1}_{r_1}^\top + {W_2}_{i,j} \mathds{1}_{r_1}  V_j^\top }) } \circ
      B_i ,
\end{align}
where $B_i = \mathds{1} \left[ \sum_{j=1}^{d_1} {{\overline{W}_1}_{i,j} {(\gone)}_{j}  \mathds{1}_{r_1}^\top + {\overline{W}_2}_{i,j} \mathds{1}_{r_1}   {(\gtwo)}_{j}^\top }  \geq 0 \right]$, $U_j = {(\gone)}_{j}$, $V_j = {(\gtwo)}_{j}$.

In all these representations, a low-rank matrix is multiplied elementwise with a binary mask $B_i$, which makes the maximum attainable rank $r_1$. Thus, with a convex MLP decoder, rank of $\hat{M}$ is limited by the resolution of the feature grids.

\subsubsection{Proof of \cref{thm:triplane-mlp}}
\label{sec:prooftriplanemlp}

For a standard 2-layer ReLU MLP with hidden size $h$, denote the trainable first and second layer weights as $W \in \mathbb{R}^{h \times d_1}$, $\alpha \in \mathbb{R}^{h \times 1}$.
We will handle three different cases for merging the interpolated features: multiplication ($\circ$), addition ($+$), and concatenation ($\odot$). 

The forward mapping of the multiplicative model using a standard nonconvex MLP, $\decoder{(\eone \circ \etwo)}$ is: 
\begin{align}
\tilde{y}(q) &= \alpha^\top \left[W ({(\gone)}_k  \circ {(\gtwo)}_l) \right]_{+}\\
     &= \sum_{i=1}^{h} {\alpha_i \left[ \sum_{j=1}^{d_1} {W_{i,j} {(\gone)}_{k,j}  {(\gtwo)}_{l,j}} \right]_{+}} .
\end{align}
The resulting matrix decomposition can then be written as
\begin{align}
    \hat{M} = \sum_{i=1}^{h} \alpha_i { \left( \sum_{j=1}^{d_1} {W_{i,j} {(\gone)}_{j} {(\gtwo)}_{j}^\top } \right)_+} 
    = \sum_{i=1}^{h} \alpha_i { \left( \sum_{j=1}^{d_1} {W_{i,j} U_{j} V_{j}^\top } \right)_+},
\end{align}
with $U_j = {(\gone)}_{j}$, $V_j = {(\gtwo)}_{j}$.

When the model uses additive features as in $\decoder{(\eone + \etwo)}$, the prediction is
\begin{align}
\tilde{y}(q) &= \alpha^\top \left[W ({(\gone)}_k + {(\gtwo)}_l) \right]_{+}\\
     &= \sum_{i=1}^{h} {\alpha_i \left[ \sum_{j=1}^{d_1} {W_{i,j} ({(\gone)}_{k,j} + {(\gtwo)}_{l,j})} \right]_{+}} .
\end{align}
The resulting matrix decomposition can then be written as
\begin{align}
    \hat{M} &= \sum_{i=1}^{h}  {\alpha_i \left( \sum_{j=1}^{d_1} {W_{i,j} ({(\gone)}_{j}  \mathds{1}_{r_1}^\top + \mathds{1}_{r_1}  {(\gtwo)}_{j}^\top )} \right)_+}  =  \sum_{i=1}^{h}  {\alpha_i \left( \sum_{j=1}^{d_1} {W_{i,j} (U_{j}  \mathds{1}_{r_1}^\top + \mathds{1}_{r_1}  V_{j}^\top )} \right)_+},
\end{align}
again with $U_j = {(\gone)}_{j}$, $V_j = {(\gtwo)}_{j}$.

Finally, we show that concatenation of features results in a very similar expression.

When the model uses concatenated features as in $\decoder{(\eone \odot \etwo)}$ and $\decoder{}$ is a standard nonconvex MLP (with trainable weights $W \in \mathbb{R}^{h \times 2 d_1}$ and  $\alpha \in \mathbb{R}^{h \times 1}$), the prediction at a point $q$ is
\begin{align}
    \tilde{y}(q) &= \alpha^\top \left[ \left(W ({(\gone)}_k \odot {(\gtwo)}_l) \right)_+  \right] .
\end{align}
Denoting the hidden layer weights as a concatenation of 2 matrices, $W = (W_1 \odot W_2)$, where $W_1, W_2 \in \mathbb{R}^{h \times d_1}$, we have the following expression:
\begin{align}
    \tilde{y}(q) &= \sum_{i=1}^{h}  { \alpha_i \left( \sum_{j=1}^{d_1} { {W_1}_{i,j} {(\gone)}_{k,j} +{{W_2}_{i,j}(\gtwo)}_{l,j}} \right)_+} .
\end{align}

Following similar steps, we express the matrix decomposition as
\begin{align}
    \hat{M} &= \sum_{i=1}^{h}  {\alpha_i \left( \sum_{j=1}^{d_1} {{W_1}_{i,j} {(\gone)}_{j}  \mathds{1}_{r_1}^\top + {W_2}_{i,j} \mathds{1}_{r_1}  {(\gtwo)}_{j}^\top } \right)_+}  \\
    &=  \sum_{i=1}^{h}  {\alpha_i \left( \sum_{j=1}^{d_1} {{W_1}_{i,j} U_j  \mathds{1}_{r_1}^\top + {W_2}_{i,j} \mathds{1}_{r_1}  V_{j}^\top } \right)_+},
\end{align}
where $U_j = {(\gone)}_{j}$, $V_j = {(\gtwo)}_{j}$.

By a similar argument to \Cref{sec:prooftriplaneconvexmlp}, the maximum attainable rank of all three representations derived here is limited by $r_1$.

\subsection{Interpolation Comparison}
\label{sec:interpolationcomparison}

\begin{figure}
    \centering
    \includegraphics[width=0.48\linewidth]{figures/image_fitting/image_fitting_bilinear.jpg}
    \includegraphics[width=0.48\linewidth]{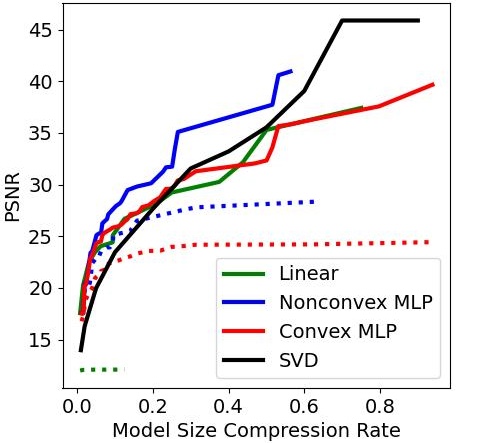}
    \caption{2D image fitting experiments matching the setting of our theoretical results, with a \modelname{} version using only vector (line) features and a decoder as specified in the legend. Left: linear interpolation of features; Right: nearest neighbor interpolation of features. For the SVD baseline we use low-rank factors whose resolution matches the target image, so no interpolation is needed (this, and the use of a nonlinear decoder, is why in some cases 2D \modelname{} can outperform SVD).}
    \label{fig:interpolationcomparison}
\end{figure}

\Cref{fig:2dvalidation} shows experiments fitting the SciPy \emph{astronaut} image using the various models considered in our 2D theoretical results. In \Cref{fig:interpolationcomparison} we compare the same experiment when we use linear interpolation into the vector (line) features (left, same as \Cref{fig:2dvalidation}) versus nearest neighbor interpolation (right, same as theorems). In this experiment we find qualitatively similar results regardless of the type of interpolation, with slightly better performance using linear interpolation; in our 3D experiments we use (bi/tri)linear interpolation.

\subsection{Benefits of Convexity}
\label{sec:benefitsofconvexity}

\begin{figure}
    \centering
    \includegraphics[width=0.65\linewidth]{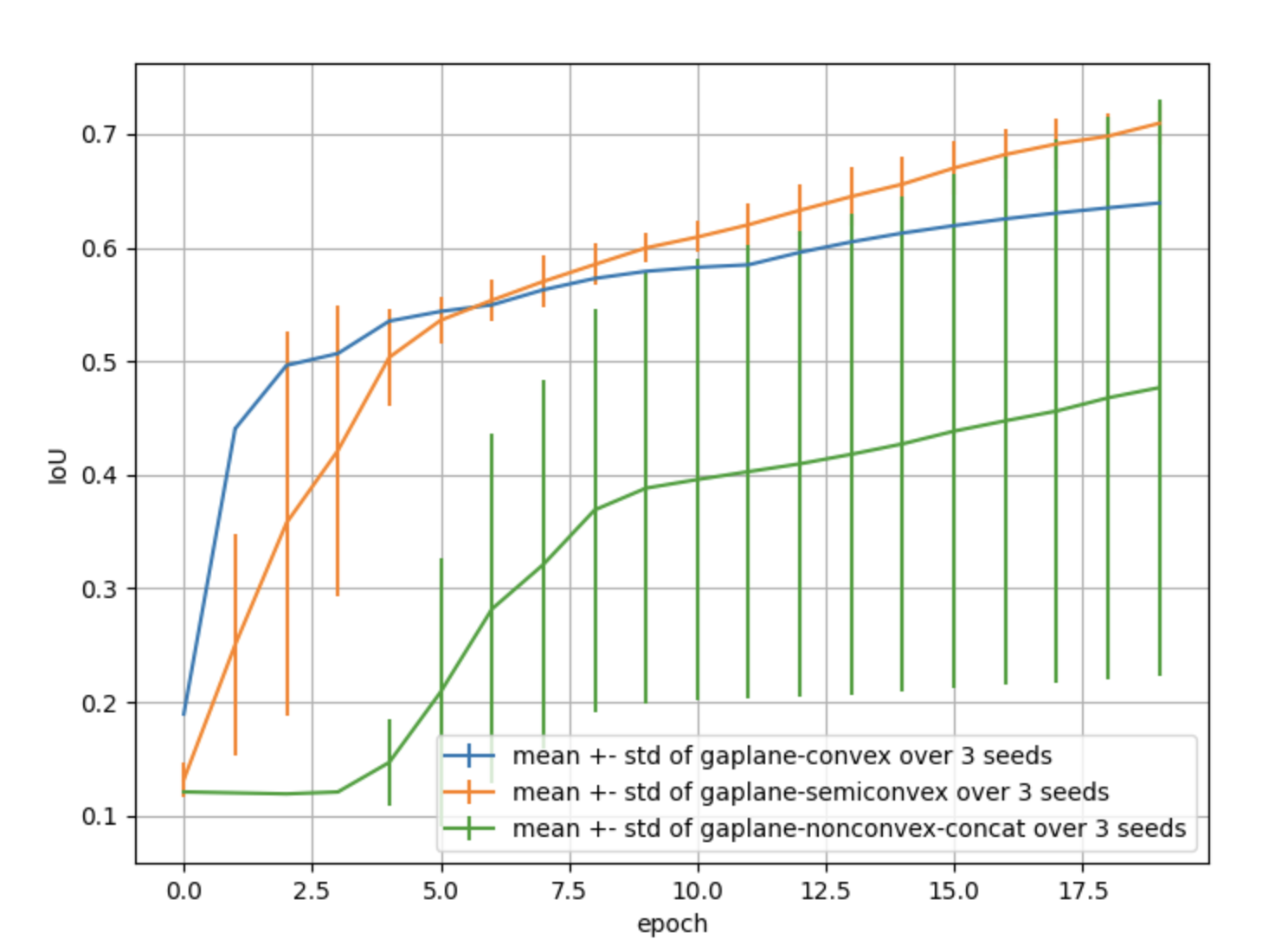}
    \caption{Test performance throughout training on our video fitting task, for a very small \modelname{} model across 3 random seeds. We find that the favorable optimization landscape of our convex and semiconvex models enables reliable training across seeds, whereas the nonconvex model fails to fit any test frame with one of the seeds.}
    \label{fig:convexitystability}
\end{figure}

In most of our experiments, all models (convex, semiconvex, and nonconvex) are large enough that they are able to optimize well. However, we highlight a benefit of our convex and semiconvex models that they enjoy more stable optimization even with very small model sizes. In \Cref{fig:convexitystability} we compare test intersection-over-union (IoU) curves for very small models for our video fitting task (hidden dimension 4 in the decoder MLP, and feature dimensions $[d_1,d_2,d_3] = [4,4,2]$ and resolutions $[r_1,r_2,r_3]=[32,32,16]$ for line, plane, and volume features, respectively). We repeat optimization with 3 different random seeds used to initialize the optimizable parameters (gating weights for the convex and semiconvex models are fixed). While the convex and semiconvex models enjoy stable training curves across random seeds, we find that the nonconvex model experiences much more volatile training behavior (completely failing to optimize with one of the 3 random seeds). 

\subsection{Results for all Nerfstudio-Blender Scenes}
\label{sec:appendixnerf}

\begin{figure}[h]
    \centering
    \plotcrop{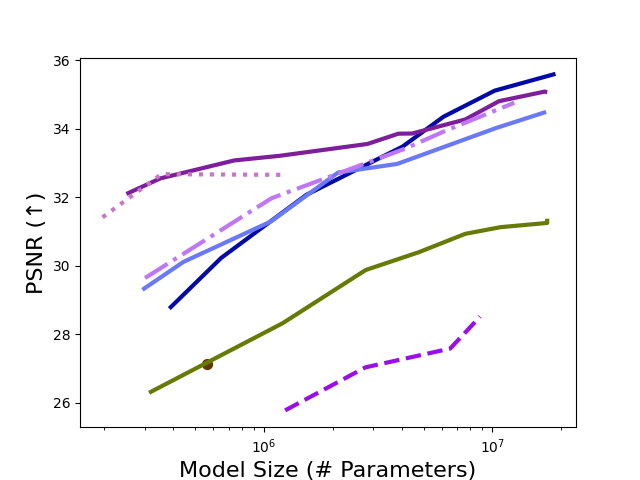}
    \plotcrop{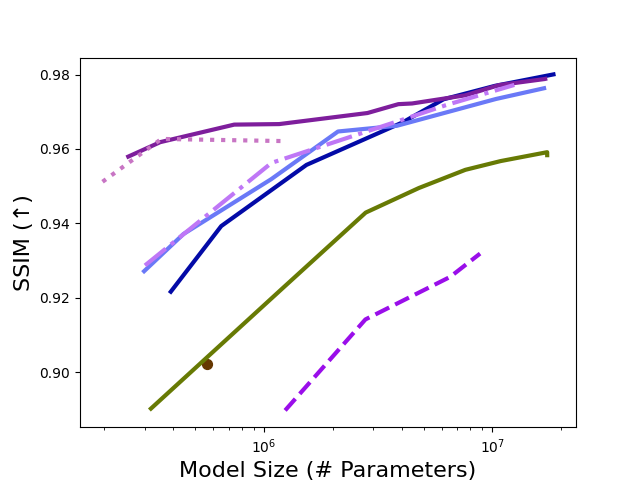}
    \plotcrop{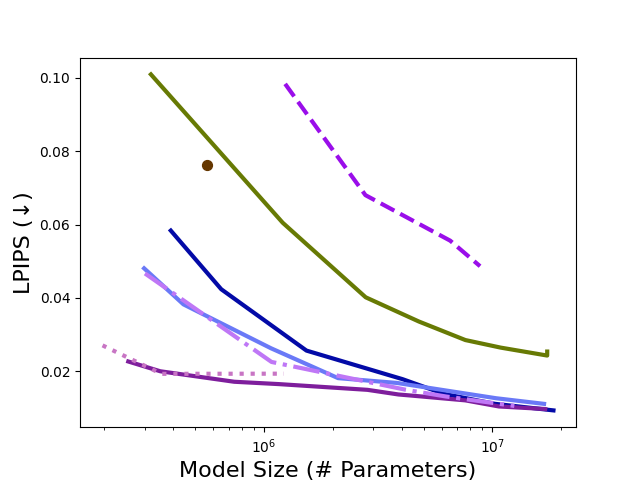}  
    \plotcroplegend{figures/nerfstudio/legend_plot_all.png}
    \caption{Results on radiance field reconstruction. Nonconvex \modelname{} (with feature multiplication) offers the most efficient representation: when the model is large it performs comparably to the state of the art models, but when model size is reduced it retains higher performance than other models. Here all models are trained for the same number of epochs on the \emph{lego} scene.}
    \label{fig:nerfstudio-lego}
\end{figure}

\begin{figure}[h]
    \centering
    \plotcrop{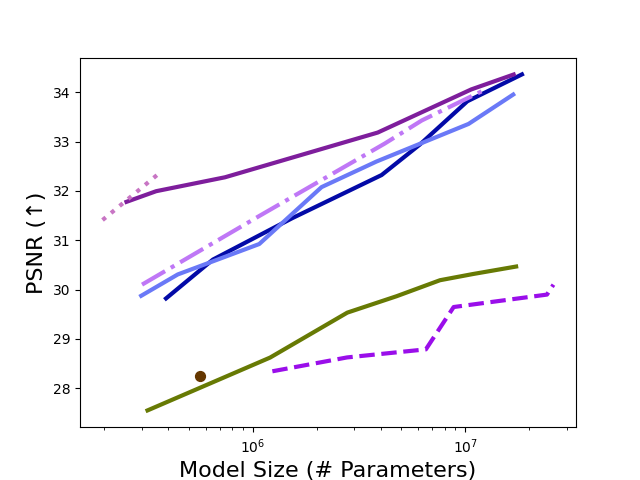}
    \plotcrop{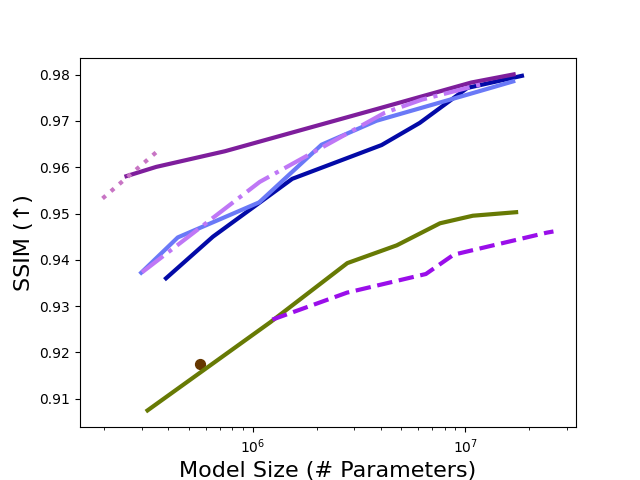}
    \plotcrop{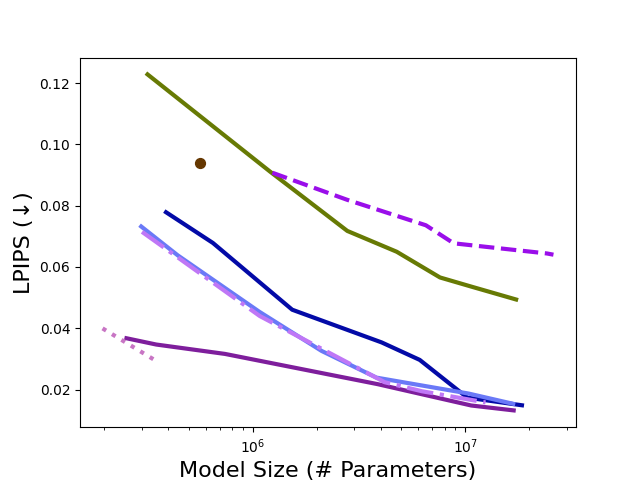}  
    \plotcroplegend{figures/nerfstudio/legend_plot_all.png}
    \caption{Results on radiance field reconstruction. Nonconvex \modelname{} (with feature multiplication) offers the most efficient representation: when the model is large it performs comparably to the state of the art models, but when model size is reduced it retains higher performance than other models. Here all models are trained for the same number of epochs on the \emph{chair} scene.}
    \label{fig:nerfstudio-chair}
\end{figure}

\begin{figure}[h]
    \centering
    \plotcrop{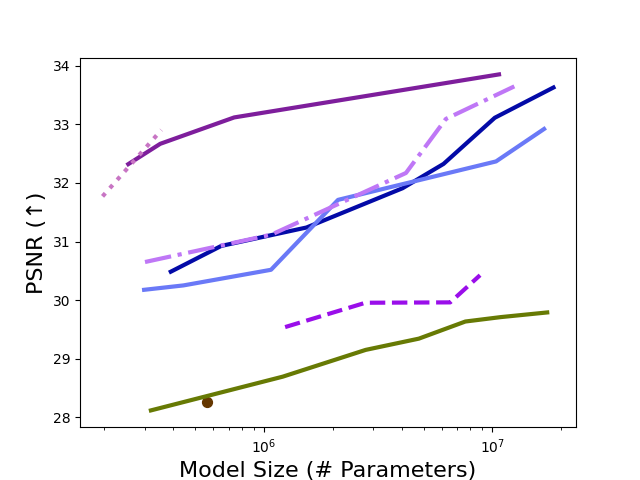}
    \plotcrop{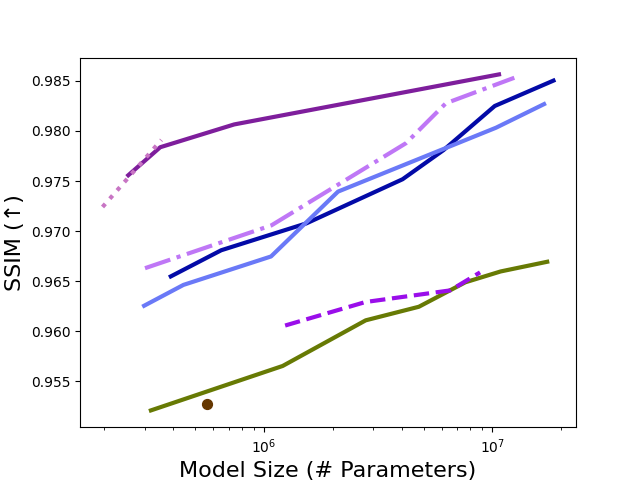}
    \plotcrop{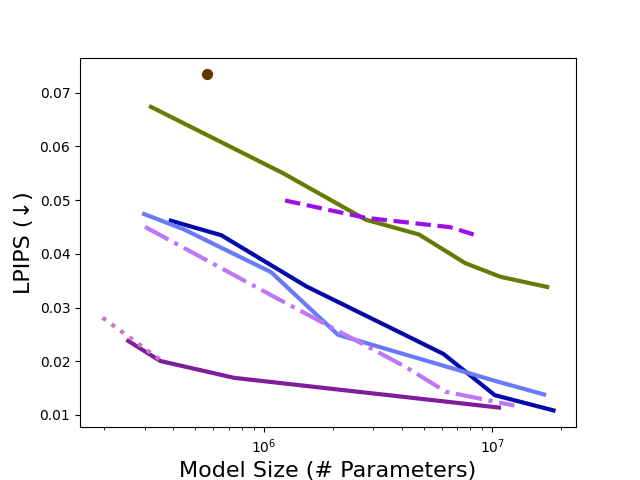}  
    \plotcroplegend{figures/nerfstudio/legend_plot_all.png}
    \caption{Results on radiance field reconstruction. Nonconvex \modelname{} (with feature multiplication) offers the most efficient representation: when the model is large it performs comparably to the state of the art models, but when model size is reduced it retains higher performance than other models. Here all models are trained for the same number of epochs on the \emph{mic} scene.}
    \label{fig:nerfstudio-mic}
\end{figure}

\begin{figure}[h]
    \centering
    \plotcrop{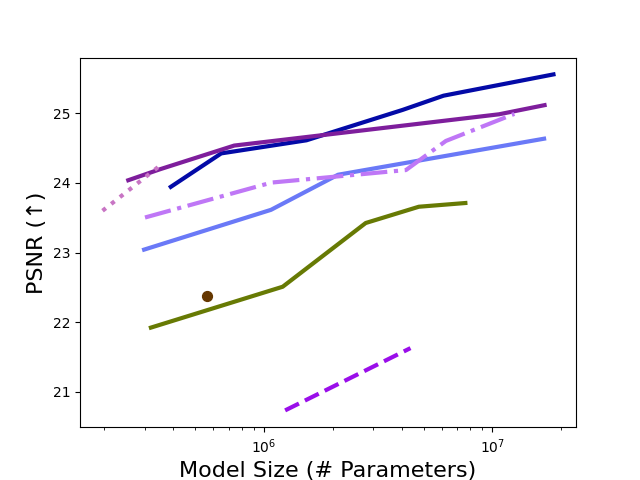}
    \plotcrop{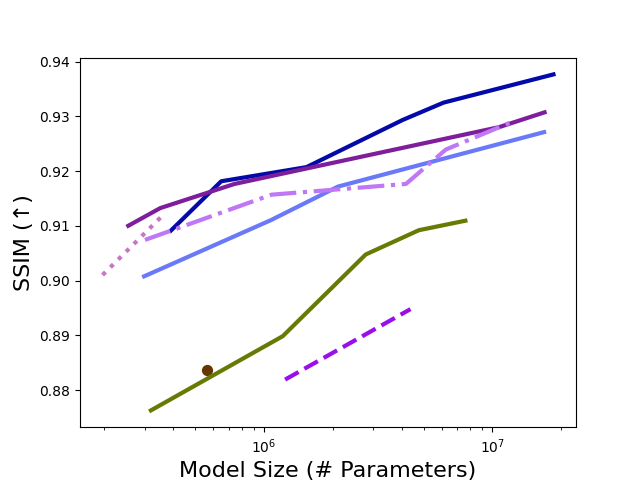}
    \plotcrop{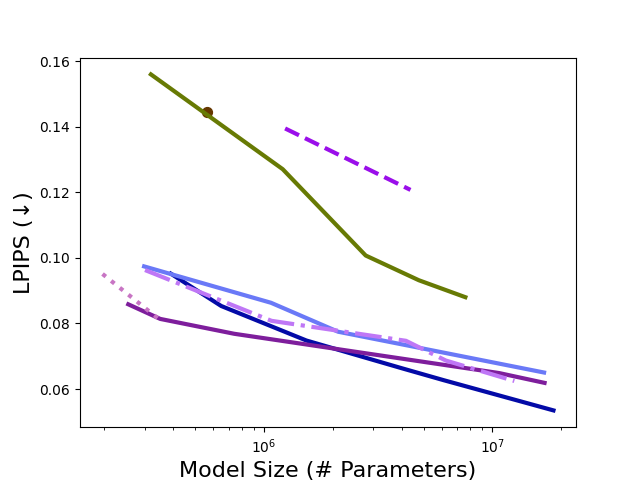}  
    \plotcroplegend{figures/nerfstudio/legend_plot_all.png}
    \caption{Results on radiance field reconstruction. Nonconvex \modelname{} (with feature multiplication) offers the most efficient representation: when the model is large it performs comparably to the state of the art models, but when model size is reduced it retains higher performance than other models. Here all models are trained for the same number of epochs on the \emph{drums} scene.}
    \label{fig:nerfstudio-drums}
\end{figure}

\begin{figure}[h]
    \centering
    \plotcrop{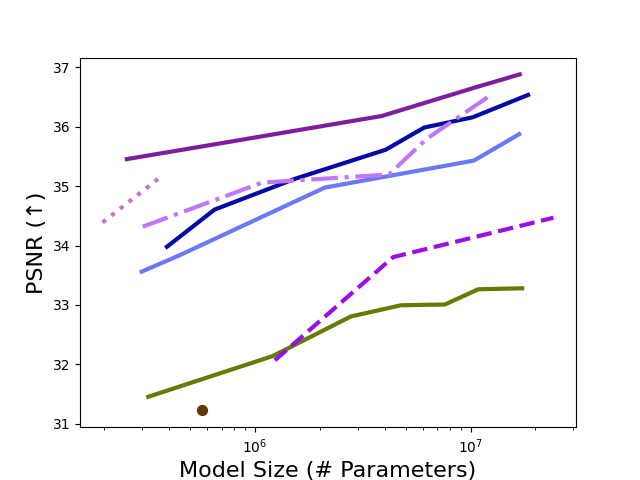}
    \plotcrop{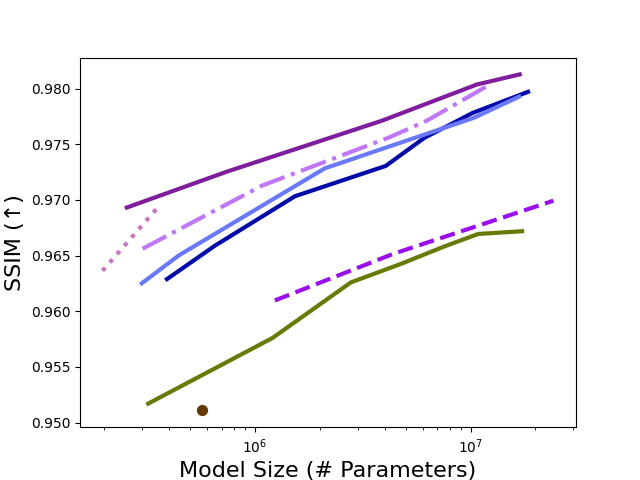}
    \plotcrop{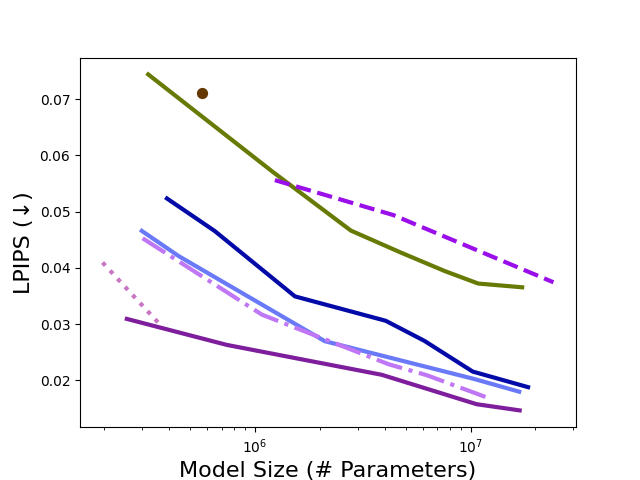}  
    \plotcroplegend{figures/nerfstudio/legend_plot_all.png}
    \caption{Results on radiance field reconstruction. Nonconvex \modelname{} (with feature multiplication) offers the most efficient representation: when the model is large it performs comparably to the state of the art models, but when model size is reduced it retains higher performance than other models. Here all models are trained for the same number of epochs on the \emph{hotdog} scene.}
    \label{fig:nerfstudio-hotdog}
\end{figure}

\begin{figure}[h]
    \centering
    \plotcrop{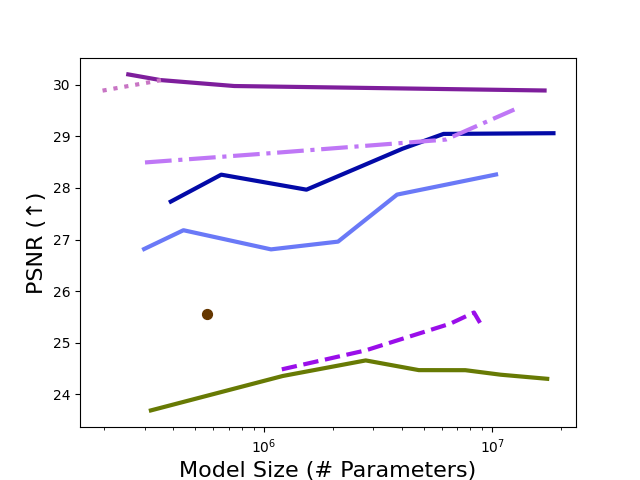}
    \plotcrop{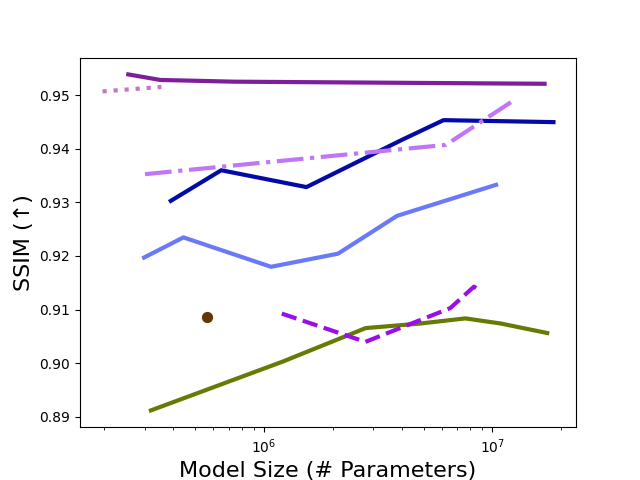}
    \plotcrop{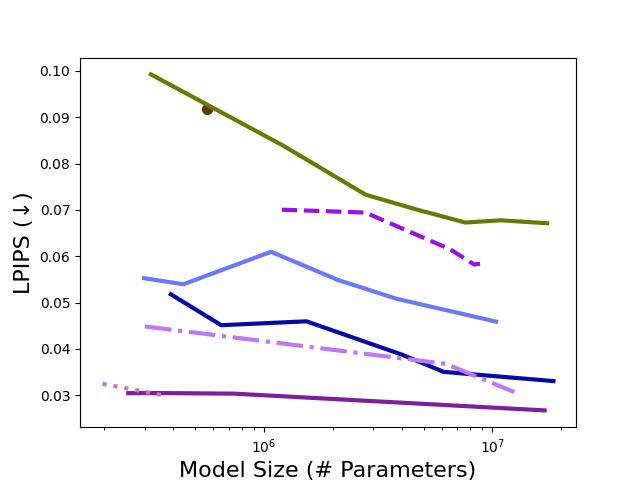}  
    \plotcroplegend{figures/nerfstudio/legend_plot_all.png}
    \caption{Results on radiance field reconstruction. Nonconvex \modelname{} (with feature multiplication) offers the most efficient representation: when the model is large it performs comparably to the state of the art models, but when model size is reduced it retains higher performance than other models. Here all models are trained for the same number of epochs on the \emph{materials} scene.}
    \label{fig:nerfstudio-materials}
\end{figure}

\begin{figure}[h]
    \centering
    \plotcrop{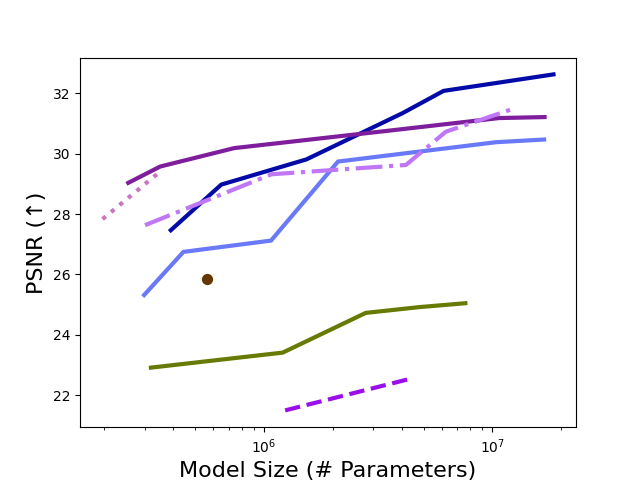}
    \plotcrop{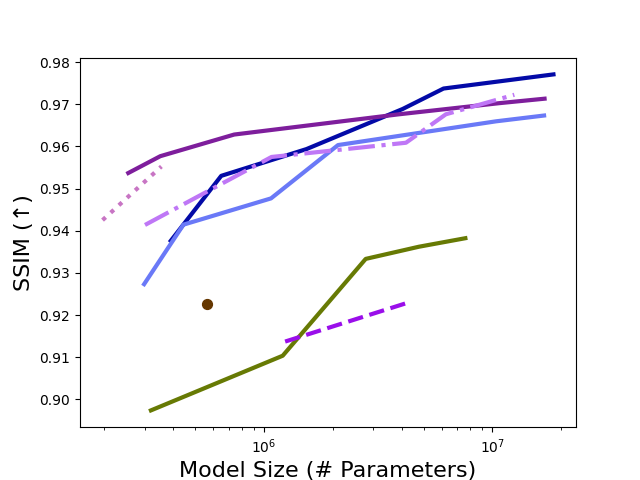}
    \plotcrop{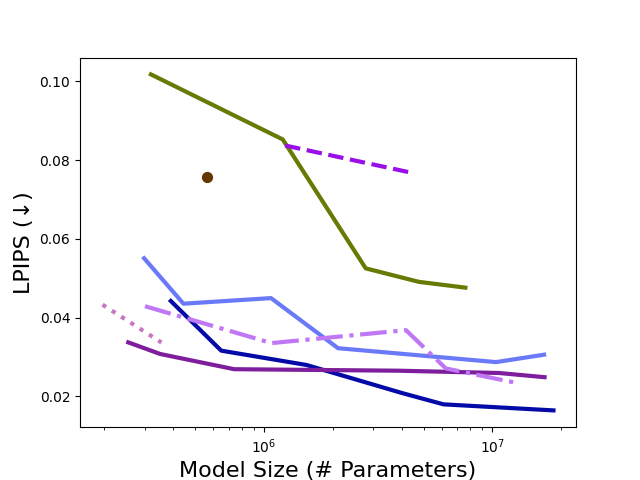}  
    \plotcroplegend{figures/nerfstudio/legend_plot_all.png}
    \caption{Results on radiance field reconstruction. Nonconvex \modelname{} (with feature multiplication) offers the most efficient representation: when the model is large it performs comparably to the state of the art models, but when model size is reduced it retains higher performance than other models. Here all models are trained for the same number of epochs on the \emph{ficus} scene.}
    \label{fig:nerfstudio-ficus}
\end{figure}

\begin{figure}[h]
    \centering
    \plotcrop{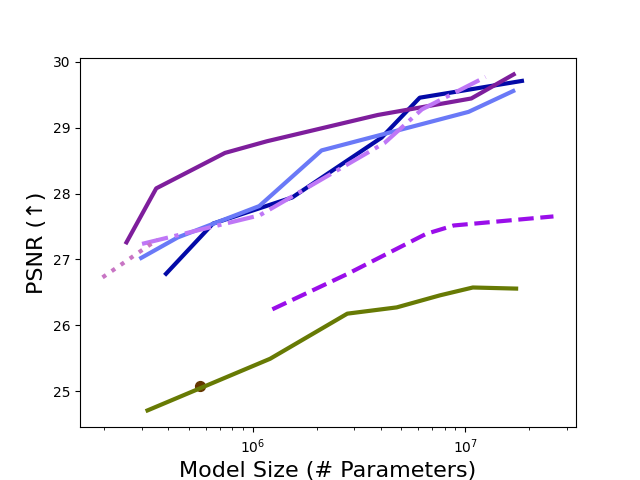}
    \plotcrop{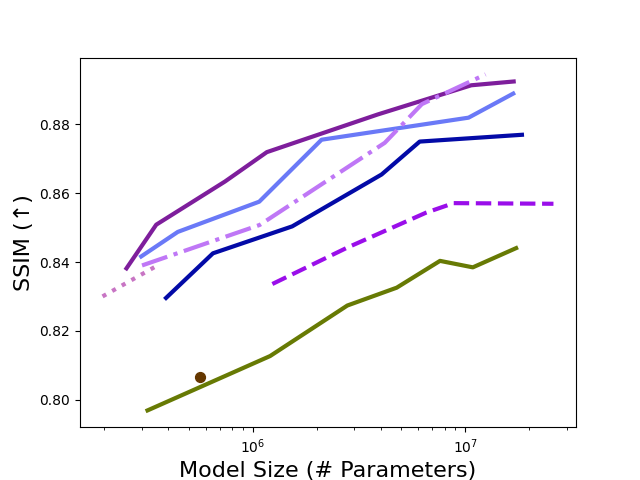}
    \plotcrop{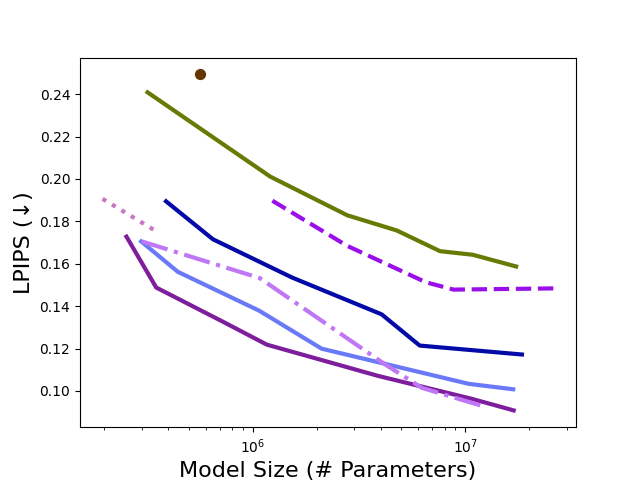}  
    \plotcroplegend{figures/nerfstudio/legend_plot_all.png}
    \caption{Results on radiance field reconstruction. Nonconvex \modelname{} (with feature multiplication) offers the most efficient representation: when the model is large it performs comparably to the state of the art models, but when model size is reduced it retains higher performance than other models. Here all models are trained for the same number of epochs on the \emph{ship} scene.}
    \label{fig:nerfstudio-ship}
\end{figure}

\subsection{Example renderings}
\label{sec:appendix-render}
In this section, we provide qualitative rendering comparisons on various scenes from the Blender dataset. We highlight the superior performance of \modelname{} with limited number of parameters by comparing the smallest K-Planes, TensoRF and \modelname{} models.

\begin{figure}[h]
  \centering
  \begin{minipage}[b]{0.3\textwidth}
    \includegraphics[width=\textwidth]{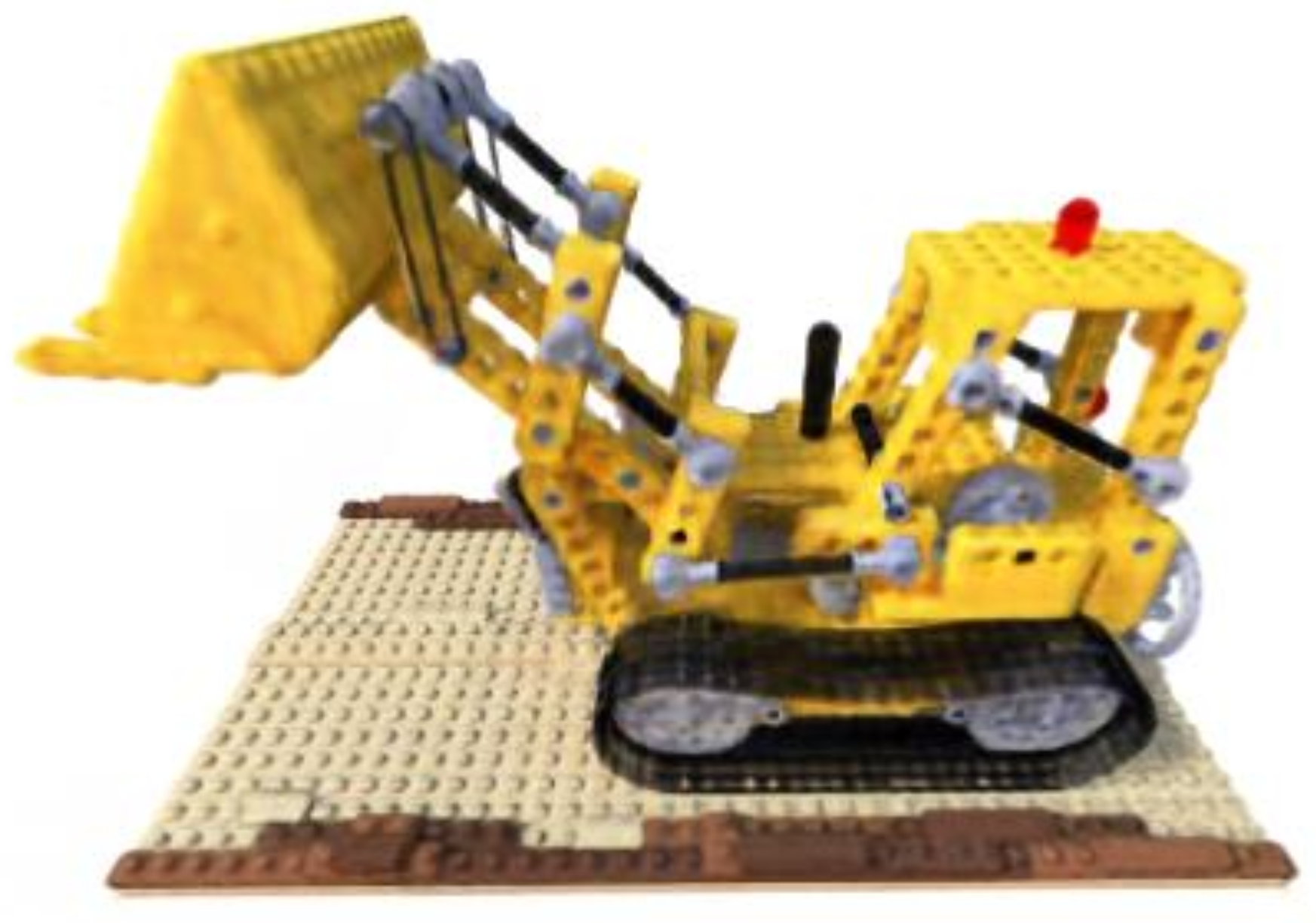}
  \end{minipage}
  \hfill
  \begin{minipage}[b]{0.3\textwidth}
    \includegraphics[width=\textwidth]{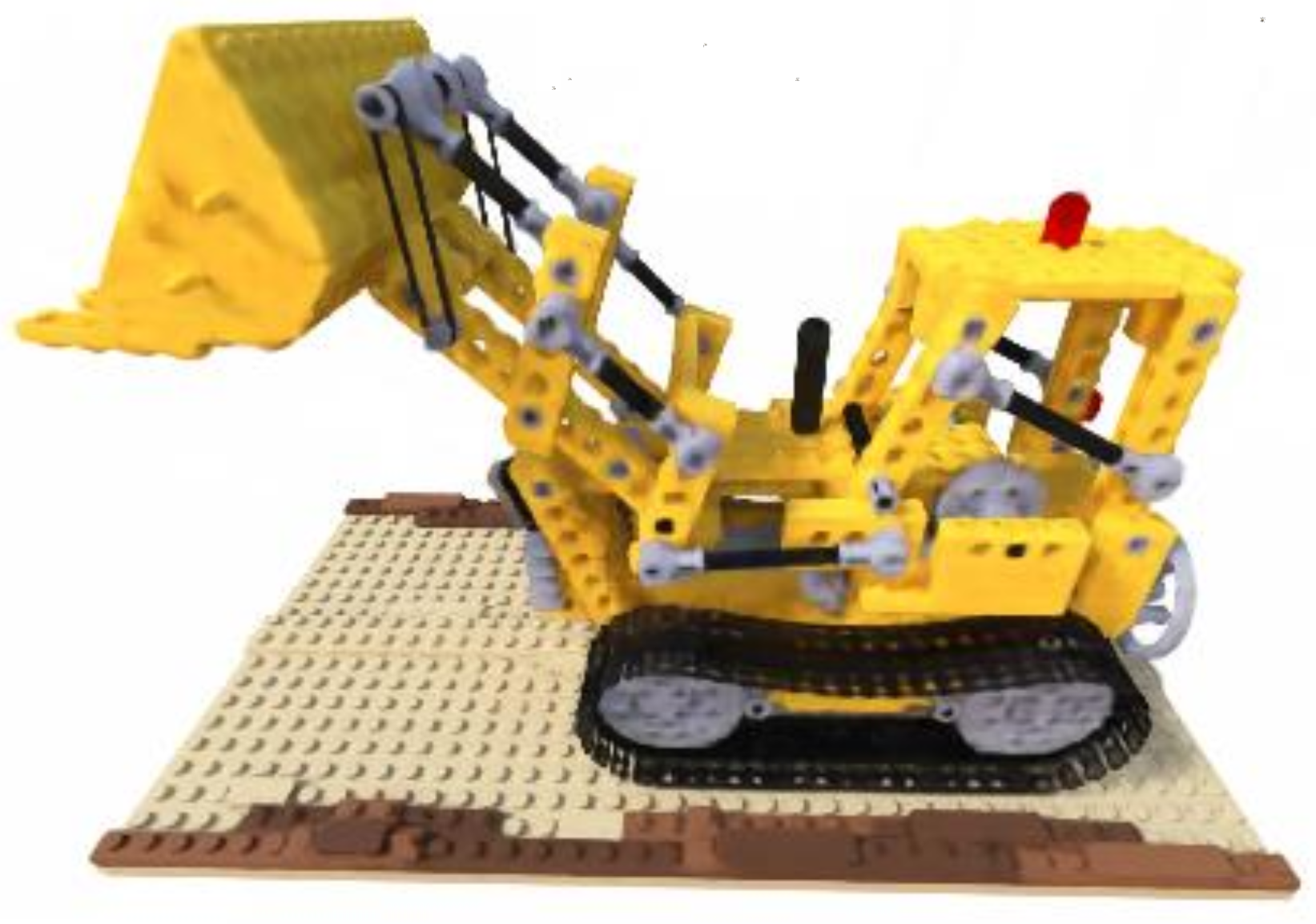}
  \end{minipage}
  \hfill
  \begin{minipage}[b]{0.3\textwidth}
    \includegraphics[width=\textwidth]{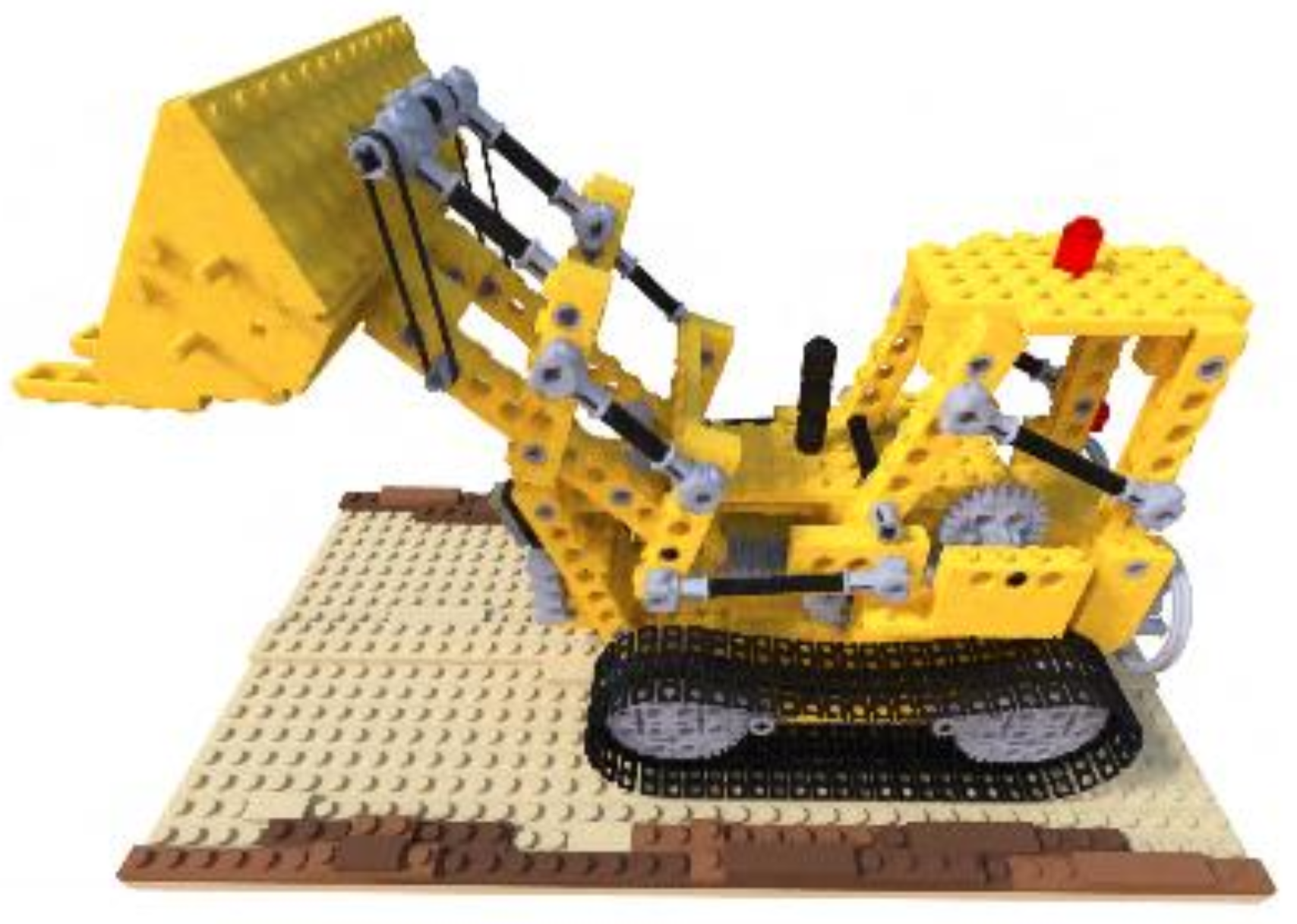}
  \end{minipage}
  \vfill
  \centering
  \begin{minipage}[b]{0.3\textwidth}
    \includegraphics[width=\textwidth]{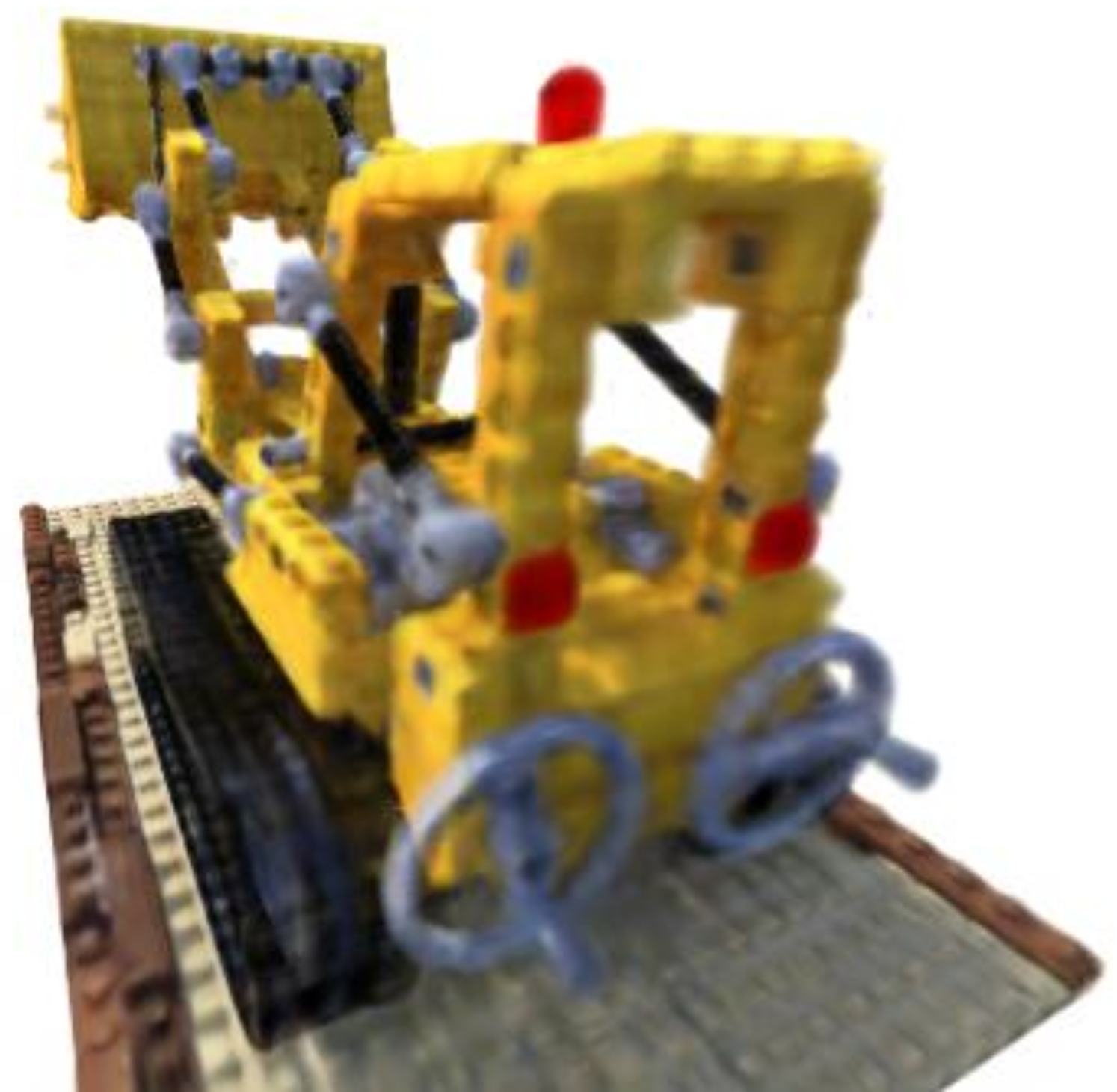}
  \end{minipage}
  \hfill
  \begin{minipage}[b]{0.3\textwidth}
    \includegraphics[width=\textwidth]{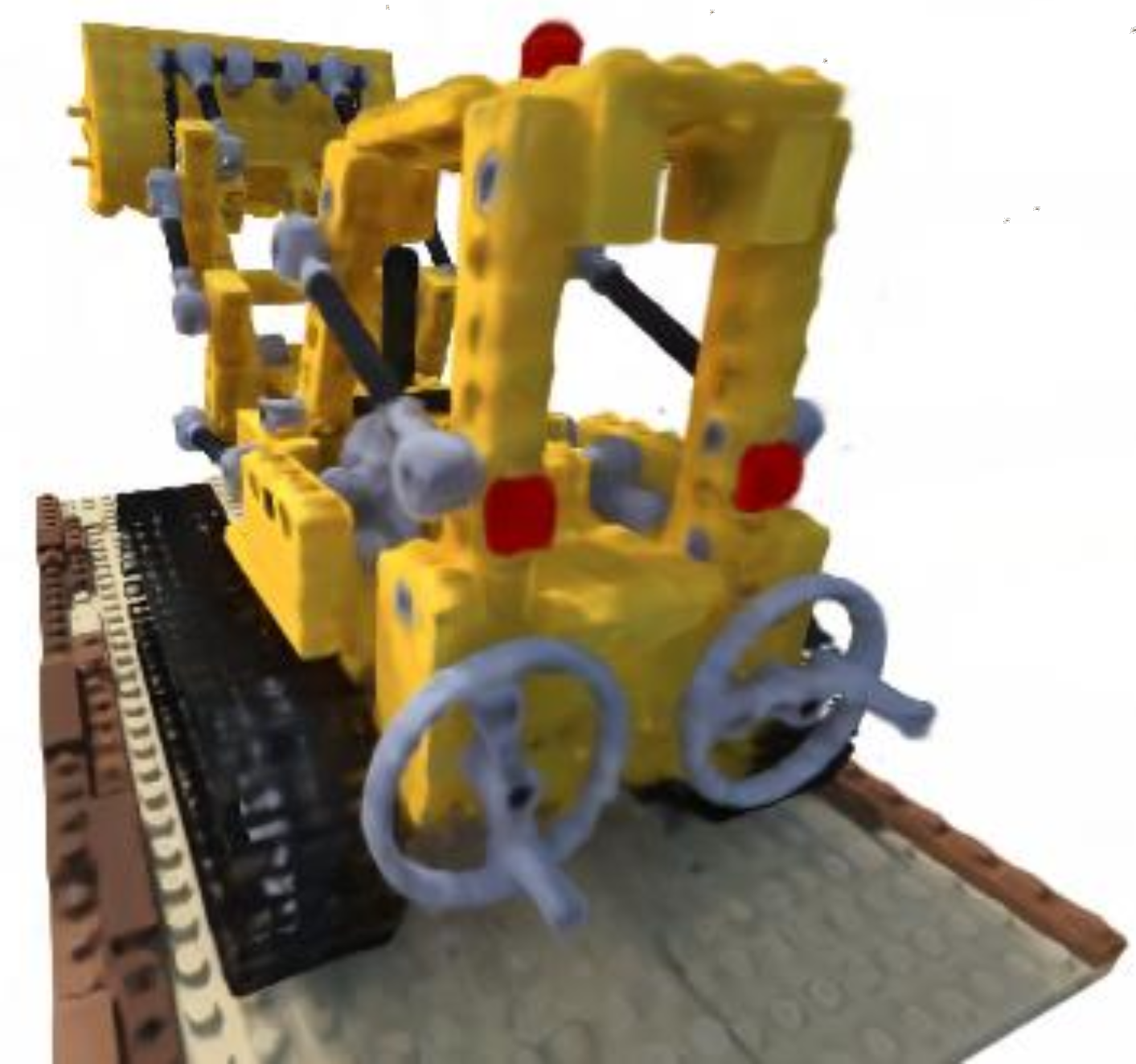}
  \end{minipage}
  \hfill
  \begin{minipage}[b]{0.3\textwidth}
    \includegraphics[width=\textwidth]{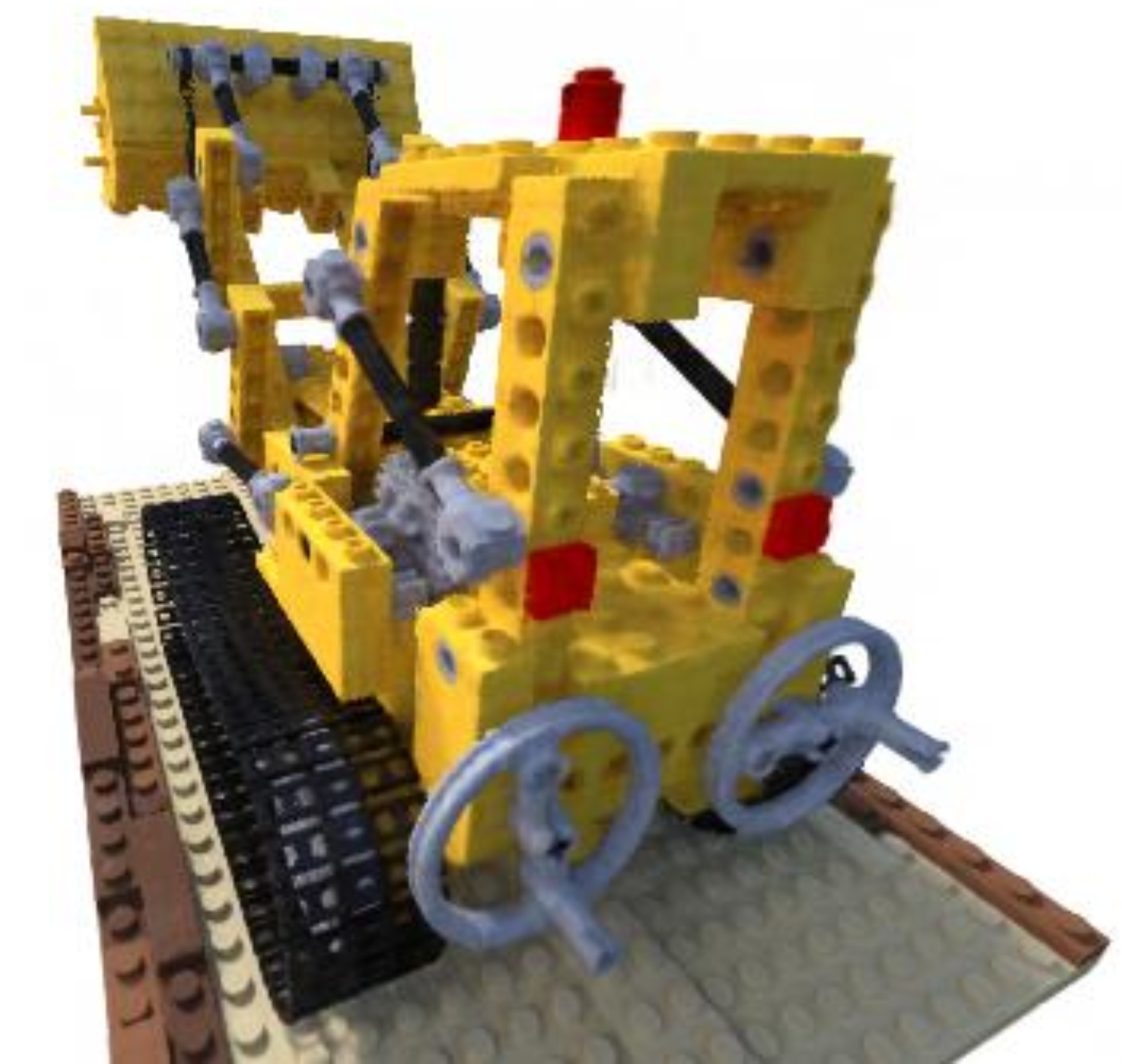}
  \end{minipage}
\caption{Rendering comparison for the \textit{lego} scene: TensoRF on the left (0.32 M parameters), K-Planes in the middle (0.39 M parameters), \modelname{} on the right (0.25 M parameters).}
\end{figure}

\begin{figure}[h]
  \centering
  \begin{minipage}[b]{ 0.3\textwidth}
    \includegraphics[width=\textwidth]{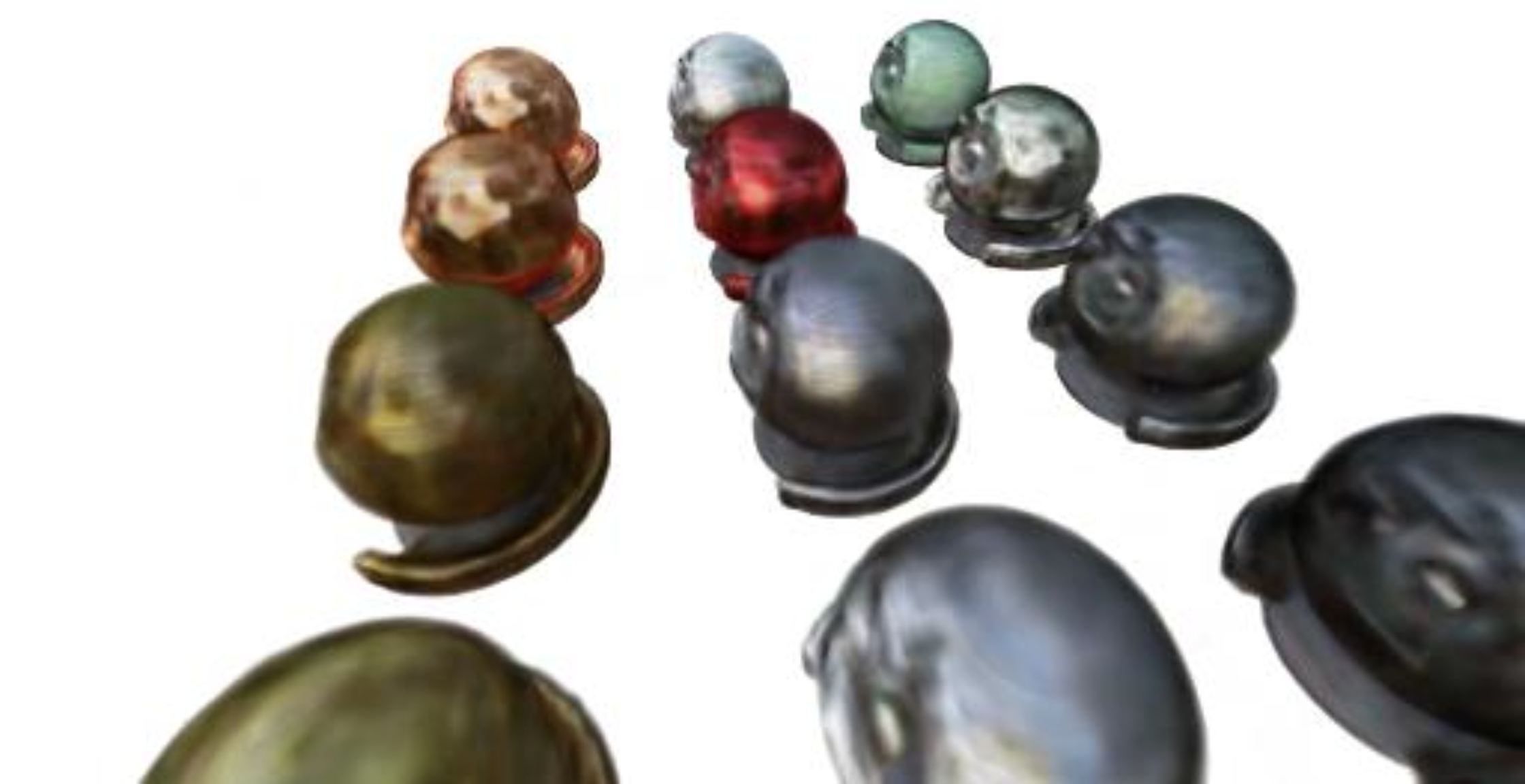}
  \end{minipage}
  \hfill
  \begin{minipage}[b]{ 0.3\textwidth}
    \includegraphics[width=\textwidth]{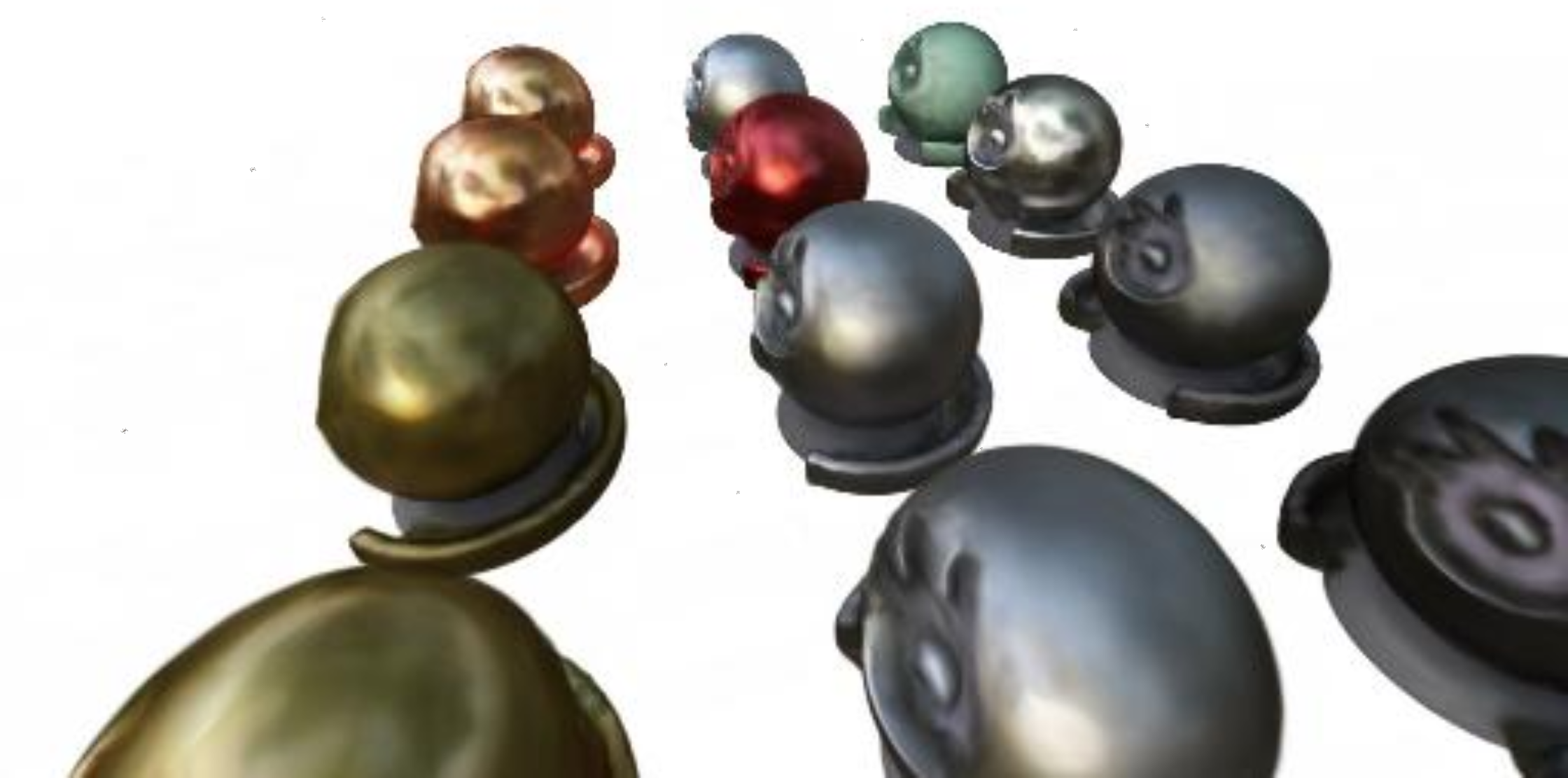}
  \end{minipage}
  \hfill
  \begin{minipage}[b]{ 0.3\textwidth}
    \includegraphics[width=\textwidth]{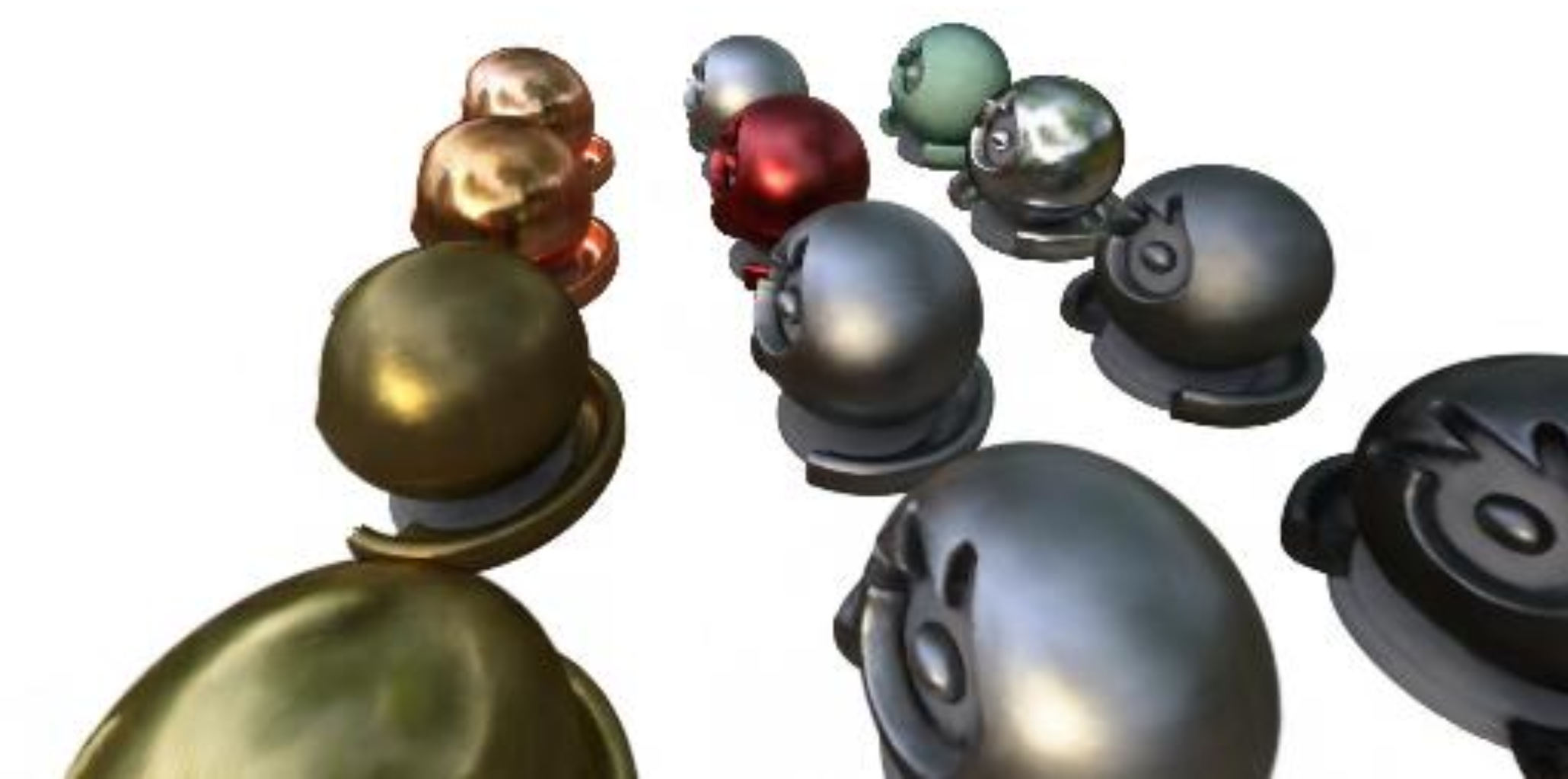}
  \end{minipage}
\caption{Rendering comparison for the \textit{materials} scene: TensoRF on the left (0.32 M parameters), K-Planes in the middle (0.39 M parameters), \modelname{} on the right (0.25 M parameters).}
\end{figure}

\subsection{Model configurations used for experiments}
\label{sec:configs}

\subsubsection{Radiance Field Modeling}
\begin{table}[h]
\centering
\resizebox{\textwidth}{!}{  
\begin{tabular}{|c|p{2cm}|p{2.5cm}|p{2.2cm}|p{3cm}|p{3cm}|} 
\hline
\textbf{Model} & \textbf{Resolutions} & \textbf{Channel Dimensions} & \textbf{Multiresolution} & \textbf{Proposal Network Resolutions} & \textbf{Number of model parameters (M)} \\ 
\hline
\multirow{7}{*}{\textbf{K-plane}} & 32 & 4 & [1, 2, 4] & [32, 64] & 0.390   \\
 & 32 & 8 & [1, 2, 4] & [32, 64] & 0.649   \\
 & 64 & 4 & [1, 2, 4] & [64, 128] & 1.533   \\
 & 64 & 8 & [1, 2, 4] & [128, 256] & 4.041   \\
 & 64 & 16 & [1, 2, 4] & [128, 256] & 6.107   \\
 & 128 & 8 & [1, 2, 4] & [128, 256] & 10.234   \\
 & 128 & 16 & [1, 2, 4] & [128, 256] & 18.493   \\
\hline
\multirow{7}{*}{\textbf{K-plane without proposal sampling}} & 32 & 4 & [1, 2, 4] & - & 0.298  \\
 & 40 & 4 & [1, 2, 4] & - & 0.444   \\
 & 64 & 4 & [1, 2, 4] & - & 1.073   \\
 & 64 & 8 & [1, 2, 4] & - & 2.108  \\
 & 100 & 6 & [1, 2, 4] & - & 3.822  \\
 & 128 & 10 & [1, 2, 4] & - & 10.367   \\
 & 129 & 16 & [1, 2, 4] & - & 16.824   \\
\hline
\multirow{8}{*}{\textbf{TensoRF}} & 128 & [2, 4] & - & - & 0.320  \\
 & 256 & [2, 4] & - & - &  1.207 \\
 & 256 & [6, 8] & - & - &   2.786\\
 & 256 & [12, 12] & - & - &   4.760\\
 & 300 & [12, 16] & - & - &  7.609 \\
 & 300 & [16, 24] & - & - &  10.860 \\
 & 300 & [32, 32] & - & - &   17.362\\
 & 300 & [16, 48] & - & - &  17.364 \\
\hline
\multirow{7}{*}{\textbf{GA-plane}} & [200, 4, 4] & [32, 32, 4] & [1, 2, 4] & - & 0.254   \\
 & [200, 8, 4] & [32, 32, 4] & [1, 2, 4] & - & 0.351   \\
 & [200, 16, 8] & [32, 32, 4] & [1, 2, 4] & - & 0.740   \\
 & [200, 32, 8] & [16, 16, 4] & [1, 2, 4] & - & 1.164   \\
 & [100, 100, 16] & [6, 6, 8] & [1, 2, 4] & - & 3.874   \\
 & [200, 128, 32] & [10, 10, 8] & [1, 2, 4] & - & 10.681   \\
 & [200, 128, 32] & [16, 16, 8] & [1, 2, 4] & - & 16.908   \\
\hline
\multirow{5}{*}{\textbf{GA-plane ablation-VM}} & 32 & 4 & [1, 2, 4] & - & 0.301   \\
 & 64 & 4 & [1, 2, 4] & - & 1.078  \\
 & 128 & 4 & [1, 2, 4] & - & 4.180  \\
 & 128 & 6 & [1, 2, 4] & - & 6.251   \\
 & 128 & 12 & [1, 2, 4] & - & 12.465   \\
\hline
\multirow{2}{*}{\textbf{GA-plane ablation-CP}} & 200 & 32 & [1, 2, 4] & - & 0.196   \\
 & 200 & 64 & [1, 2, 4] & - & 0.355   \\
\hline
 \multirow{4}{*}{\textbf{GA-plane ablation-volume}} & 18 & [3, 5, 6] & [1, 2, 3] & - & 1.236   \\
 & 24 & [4, 4, 6] & [1, 2, 3] & - & 2.778   \\
 & 32 & [4, 4, 6] & [1, 2, 3] & - & 6.529   \\
 & 32 & [4, 6, 8] & [1, 2, 3] & - & 8.824  \\
\hline
\end{tabular}
}
\caption{Model configurations used for the radiance field modeling task on the Blender dataset.}
\label{table:models}
\end{table}

The original K-planes model uses 2 proposal networks with different resolutions (as noted in \Cref{table:models}) and a fixed channel dimension of 8 for both. The resolutions and channel dimensions for either K-planes model (with vs. without proposal sampling) refer to $r_2$ and $d_2$, respectively. TensoRF model resolutions and channel dimensions can be interpreted in a similar way, since their feature combination dictates that $d_1 = d_2$ and they initialize the line and plane grids with the same resolution. The only nuance is that TensoRF constructs separate features for color and density decoding. Hence, the channel dimensions for density and color features are listed. Instead of a multiresolution scheme, TensoRF starts from the base resolution of $r_1=r_2=128$ and upsamples the grids to reach the final resolutions on \Cref{table:models}. Resolutions listed under \modelname{} should be interpreted as $[r_1, r_2, r_3]$; channel dimensions as $[d_1, d_2, d_3]$. For all models that use the multiresolution scheme, the base resolutions (i.e. $[r_1, r_2, r_3]$) are multiplied with the upsampling factors. For instance, a base resolution $[r_1, r_2, r_3]$ with channel dimensions $[d_1, d_2, d_3]$ and multiresolution copies $[m_1,m_2,m_3]$ will generate the grids of \modelname{} as follows: Linear feature grids $\gone, \gtwo, \gthree$ will have the shapes $\{ [m_1r_1, d_1], [m_2r_1, d_1], [m_3r_1, d_1]\}$, plane grids $\gonetwo, \gtwothree, \gonethree$ will have the shapes $\{ [m_1r_2, m_1r_2, d_2], [m_2r_2, m_2r_2, d_2], [m_3r_2, m_3r_2, d_2]\}$, and the volume grid $\gonetwothree $ will have the shape $[r_3,r_3,r_3,d_3]$. Although we don't use multiresolution copies for the volume grid in \modelname{}, we do use multiresolution for the volume-only \modelname{} ablation. The resolution for that model refers to $r_3$, and the channel dimensions are also allowed to vary for each resolution (unlike other variants with multiresolution, where the feature dimension is fixed across resolutions). If we denote these varying feature dimensions as $[d_{3a}, d_{3b}, d_{3c}]$, the multiresolution copies of the volume grids will have the shapes $\{ [m_1r_3, m_1r_3, m_1r_3, d_{3a}], [m_2r_3, m_2r_3, m_2r_3, d_{3b}], [m_3r_3, m_3r_3, m_3r_3, d_{3c}]  \}$.

\subsubsection{3D Segmentation}
\modelname{} model uses feature dimensions $[d_1, d_2, d_3] = [36, 24, 8]$ (with $\odot$) or $[d_1, d_2, d_3] = [25, 25, 8]$ (with $\circ$) and resolutions $[r_1, r_2, r_3] = [128, 32, 24]$. Multiresolution grids are not used for this task since density prediction can be achieved by a simpler architecture. The model size is 0.22 M. Tri-Planes model has the feature dimension $d_2 = 4$, and resolution $r_2 = 128$ resulting in a total number of parameters of 0.2 M. Note that we fix these sizes across (non/semi)convex formulations, which causes slight variations in the size of the decoder, however, the grids constitute the most number of parameters, making this effect negligible.

\subsubsection{Video Segmentation}
\modelname{} model uses feature dimensions $[d_1, d_2, d_3] = [32, 16, 8]$ and resolutions $[r_1, r_2, r_3] = [128, 128, 64]$. When the features are combined by multiplication in the nonconvex model, $d_1 = d_2 = 16$. Multiresolution grids are not used for this task. The model size is 2.9 M. Tri-Planes model has the feature dimension $d_2 = 59$, and resolution $r_2 = 128$ resulting in a total number of parameters of 2.9 M.

\end{document}